\documentclass[10pt,twocolumn,letterpaper]{article}

\usepackage{iccv}
\usepackage{times}
\usepackage{epsfig}
\usepackage{graphicx}
\usepackage{amsmath}
\usepackage{amssymb}

\usepackage{booktabs}
\usepackage{subcaption}
\usepackage{caption}
\usepackage{color}
\usepackage{colortbl}
\usepackage[dvipsnames]{xcolor}
\usepackage[pagebackref=true,breaklinks=true,letterpaper=true,colorlinks,bookmarks=false]{hyperref}

\iccvfinalcopy % *** Uncomment this line for the final submission

\ificcvfinal\fi

\begin{document}

%%%%%%%%% TITLE
\title{Inverting and Understanding Object Detectors}

\author{Ang Cao, \; Justin Johnson\\
University of Michigan\\
{\tt\small \{ancao, justincj\}@umich.edu}}

\maketitle
\ificcvfinal\thispagestyle{empty}\fi

%%%%%%%%% ABSTRACT
\begin{abstract}
    As a core problem in computer vision, the performance of object detection has improved drastically in the past few years. 
    Despite their impressive performance, object detectors suffer from a lack of interpretability.
    Visualization techniques have been developed and widely applied to introspect the decisions made by other kinds of deep learning models; however visualizing object detectors has been underexplored.
    In this paper, we propose using inversion as a primary tool to understand modern object detectors and develop an optimization-based approach to layout inversion, allowing us to generate synthetic images recognized by trained detectors as containing a desired configuration of objects.
    We reveal intriguing properties of detectors by applying our layout inversion technique to a variety of modern object detectors, and further investigate them via validation experiments:
    they rely on qualitatively different features for classification and regression; they learn canonical motifs of commonly co-occurring objects; they use different visual cues to recognize objects of varying sizes.
    We hope our insights can help practitioners improve object detectors.
    \footnote{The code is \href{https://github.com/Caoang327/vis_det}{\url{https://github.com/Caoang327/vis_det}}}
    
% 22 lines
% However, visualizations of object detectors have been underexplored, even though there is a significant interest in using visualization to understand the decisions made by deep learning models.
% This paper generalizes visualizations to object detectors by solving the layout inversion problem via an alternating optimization approach. 
% Our visualizations reveal qualitative differences between visual cues used by a wide variety of detectors, which vary in meta-architecture~(single-stage vs. two-stage, anchor-based vs. anchor-free) and are trained for various tasks (detection vs. instance segmentation). We use inversion transfer to quantify the performance of our inversions further. 
% By visualization, we reveal intriguing properties of object detectors: relying on qualitatively different features for classification and regression; 
% learning canonical motifs of commonly co-occurring objects;
% using different features for recognizing objects with varying scales.  
% We design quantitative experiments to validate these observations. 
% We hope that these insights can help practitioners diagnose and improve object detectors.
\end{abstract}

%%%%%%%%% BODY TEXT
\section{Introduction}

Object detection, which requires jointly classifying and localizing objects, is a core computer vision task widely used in applications such as autonomous vehicles, biomedical imaging, and content recommendation.
Deep learning has enabled swift progress on this task, making detectors based on convolutional networks accurate~\cite{girshick2014rich,ren2015faster,he2017mask} and efficient~\cite{liu2016ssd,redmon2016you,lin2017focal,bochkovskiy2020yolov4} on challenging datasets~\cite{lin2014microsoft,gupta2019lvis}.

Despite their impressive performance, object detectors (like other deep learning models) suffer from a lack of interpretability.
Their internal representations and decisions are opaque, and the visual cues used to make these decisions are nonobvious.
An improved understanding of object detectors may help researchers diagnose errors and ultimately improve their object detection models.

\begin{figure}[ht!]
    \centering
    \includegraphics[width=0.45\textwidth]{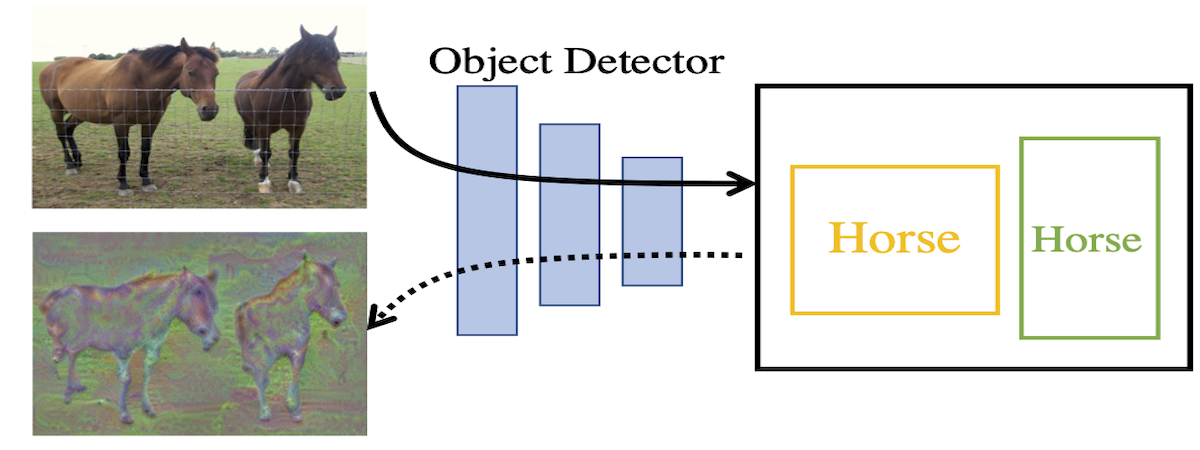}
    \vspace{-2mm}
    \caption{\textbf{Layout Inversion}:
    An object detector predicts a \emph{scene layout} from an image (solid arrow).
    We develop methods for \emph{layout inversion} (dotted arrow) that invert trained detectors, and generate images recognized as containing the target layout.}
    \vspace{-5mm}
    \label{fig:teaser}
\end{figure}

Prior work has developed techniques for visualizing and understanding other types of deep models, including convolutional~\cite{zeiler2014visualizing}, recurrent~\cite{karpathy2015visualizing,li2015visualizing} and adversarial~\cite{bau2018gan} networks.
Generalizing these techniques to object detectors is nontrivial due to their highly structured outputs: while an image classification model gives a single category label per image, an object detector emits a variably-sized set of detections, each involving classification, localization, and (optionally) segmentation~\cite{hariharan2014simultaneous,he2017mask}.
Object detectors also rely on non-differentiable operations like non-max suppression (NMS) which break the gradient pathway between predicted objects and image pixels, preventing the use of na\"{i}ve gradient ascent used to visualize other model types.

This paper extends network visualization techniques to modern object detectors to gain insight into these models and better understand them.
As our primary tool for investigation, we propose an optimization-based approach to \emph{layout inversion} (Figure~\ref{fig:teaser}) that generates synthetic images recognized by trained detectors as containing a desired configuration of objects.
We circumvent non-differentiabilities with an alternating optimization algorithm inspired by the alternating direction method of multipliers (ADMM)~\cite{boyd2011distributed}.

In Section~\ref{sec:invert} we show that our layout inversion method can be applied to a wide variety of modern object detectors that vary in meta-architecture (single-stage vs. two-stage, anchor-based vs. anchor-free) and that are trained for different tasks (object detection vs. instance segmentation).
These visualizations reveal qualitative differences between the visual cues used by different detectors for recognition.

In Section~\ref{sec:inver_trans} we use \emph{inversion transfer} to quantify the performance of our layout inversions by checking whether an image generated from one detector can be recognized by another detector.
This verifies the effectiveness of our layout inversion approach, and also allows us to measure whether different detectors rely on differing visual cues.

Object detectors predict both \emph{what} objects are present and \emph{where} they are located.
In Section~\ref{sec:dis_losses} we disentangle the effects of these subtasks by performing layout inversion targeting different combinations of the classification, regression, and segmentation heads of detectors.
These experiments show that detectors rely on qualitatively different visual features for classification and localization.
We investigate further by using \emph{attribution methods} to show that classification and regression rely on different image regions.

Full layout inversion requires finding an image where a target set of regions are recognized by the detector as objects, and where all other regions are dismissed as background.
In Section~\ref{sec:vis_obj} we probe the priors learned by object detectors by performing layout inversions requiring only a single object to be recognized as foreground, and omitting the requirement that other regions be dismissed.

In Section~\ref{sec:sur_con} we investigate the role of \emph{context}, and demonstrate that detectors learn canonical motifs of commonly co-occurring objects.
We thus hypothesize that detectors rely heavily on context to perform recognition.

In Section~\ref{sec:var_scale} we observe that varying \emph{object scale} causes detectors to rely on qualitatively different features for recognition.
Small objects are recognized primarily by shape and context, while large objects are recognized as amalgamations of part and texture.
We quantify this hypothesis by eliminating object texture via masked style transfer.

Our contributions are two-folded. 
First, our experiments show that the proposed layout inversion is a versatile tool for introspecting object detectors.
Second, we gain novel insights into these models' inner workings via inversion and design experiments to validate these hypotheses. 
We hope these insights can help practitioners better understand and improve these models' performance in the future.

\section{Related Work}

% 11 lines
\textbf{Object Detection.}
Classical approaches use a sliding-window approach, densely applying a classifier to image patches~\cite{lecun1989backpropagation,vaillant1994original,dalal2005histograms,felzenszwalb2008discriminatively}.
More recently the two-stage R-CNN framework~\cite{girshick2014rich,he2015spatial,girshick2015fast,ren2015faster} has become a dominant paradigm, where one neural network applied to images predicts regions of interest (RoIs) and a second per-region network classifies objects.
This versatile framework has been extended to tasks including instance segmentation~\cite{he2017mask}, keypoint estimation~\cite{he2017mask}, and 3D shape prediction~\cite{gkioxari2019mesh}.
Single-stage detectors form a second category of recent methods, in which a single fully-convolutional network is applied per-image to jointly propose and classify regions~\cite{liu2016ssd,redmon2016you,redmon2017yolo9000,lin2017focal,redmon2018yolov3}.
Single and two-stage methods typically regress boxes from a set of convolutional anchors; recently a number of anchor-free approaches have been proposed that model object corners~\cite{law2018cornernet} or centers~\cite{zhou2019objects,duan2019centernet,tian2019fcos} or use transformers~\cite{carion2020end}.
In this paper we focus primarily on RetinaNet~\cite{lin2017focal}, Faster R-CNN~\cite{ren2015faster} and Mask R-CNN~\cite{he2017mask} as representatives of the common anchor-based single and two-stage approaches; we also work on FCOS~\cite{tian2019fcos} as a representative of anchor-free methods.

\textbf{Neural Network Visualization.}
Techniques have been developed for introspecting the decisions made by convolutional~\cite{simonyan2013deep,zeiler2014visualizing}, recurrent~\cite{karpathy2015visualizing,li2015visualizing,strobelt2017lstmvis} and adversarial~\cite{bau2018gan} networks.
A typical approach is synthesizing images which maximize predicted category scores or internal neuron values~\cite{simonyan2013deep,zeiler2014visualizing,yosinski2015understanding,nguyen2016multifaceted}; other approaches invert feature vectors~\cite{mahendran2015understanding,dosovitskiy2016inverting}.
Images are typically generated via gradient descent on pixels, though some approaches use evolutionary search~\cite{nguyen2015deep}.
Image regularizers are used to encourage generated images to appear natural~\cite{simonyan2013deep,yosinski2015understanding,mordvintsev2015inceptionism,nguyen2016multifaceted, nguyen2016synthesizing,nguyen2017plug}
% Image regularizers are used to encourage generated images to appear natural including norm constraints~\cite{simonyan2013deep}, Gaussian blur~\cite{yosinski2015understanding}, jitter~\cite{mordvintsev2015inceptionism}, multifaceted initialization~\cite{nguyen2016multifaceted}, and GAN priors~\cite{nguyen2016synthesizing,nguyen2017plug}.
Alternatives to synthesizing images include measuring agreement between neuron activation maps and semantic maps~\cite{zhou2014object,zhou2018interpreting,bau2017network}, or finding patches that maximize hidden activations~\cite{zeiler2014visualizing}.
To date these techniques have been mainly applied to image classification models. 
Object detectors are only be visualized before the deep learning era, with HOG features and SVM~\cite{vondrick2013hoggles}; we generalize them to modern object detectors.

\textbf{Attribution Methods.}
Some methods introspect deep models by determining which part of an input image was responsible for the network's decision by computing gradients of the network output with respect to the input image~\cite{simonyan2013deep,zeiler2014visualizing,springenberg2014striving,smilkov2017smoothgrad} or using saliency maps~\cite{zhou2016learning,rebuffi2020there,fong2019understanding,selvaraju2017grad}.
Other methods fuse information from gradients, activations, and weights using classification activation mapping (CAM)~\cite{zhou2016learning,selvaraju2017grad,rebuffi2020there}, or measure how occlusions to input images affect network outputs~\cite{zeiler2014visualizing,petsiuk2018rise,fong2019understanding}.
These methods have been applied to image-level tasks like classification~\cite{simonyan2013deep,zeiler2014visualizing}, question answering~\cite{zhou2016learning}, and captioning~\cite{selvaraju2017grad}; we extend them to the region-based task of object detection.

% 10 lines
% \textbf{Neural Network Understanding.}
% A popular approach to understanding neural networks is via attribution methods, which aim to investigate how each part of the image is responsible for determining the outputs, where a saliency map is often involved in this manner\cite{zhou2016learning,rebuffi2020there,fong2019understanding,selvaraju2017grad}. \cite{simonyan2013deep} calculates the derivate of the network's output w.r.t the input image to visualize the contribution of each pixel, and the following works (e.g., DeCovNet \cite{zeiler2014visualizing}, Guided Backprop \cite{springenberg2014striving}, SmoothGrad \cite{smilkov2017smoothgrad}) reduce the noise in saliency map by tweaking the gradients. Furthermore, the localization approaches like CAM \cite{zhou2016learning}, Grad-Cam \cite{selvaraju2017grad}, and Norm-Cam \cite{rebuffi2020there} can be category discriminative by fusing the gradients, weights, and activation map.  Another strategy to understand the contributions of the individual input component is perturbing the input and observing the output changes. Occlusion \cite{zeiler2014visualizing}, RISE \cite{petsiuk2018rise}, Extremal Perturbations \cite{fong2019understanding} are proposed in this family. \JJ{Shorten}

% 4 lines
\textbf{Image Generation from Layouts.}
Some conditional generative models synthesize images from different kinds of scene layouts such as keypoints~\cite{reed2016learning,reed2017parallel}, scene graphs~\cite{johnson2018image,ashual2019specifying}, bounding boxes~\cite{zhao2019image,sun2019image}, or segmentation maps~\cite{isola2017image,zhu2017unpaired,chen2017photographic,wang2018high,park2019semantic}.
We also generate images from layouts, but rather than training conditional generative models we invert pretrained object detectors as an introspection tool.

% 9 lines
%\textbf{Image Generation with Layout Information.}
%Our work is also closely related to the research about synthesizing images with multiple objects. Generating realistic multi-object scenes is a difficult task since it is very complicated for the neural network to generate many components and reasoning for their relationships. Some approaches take the scene graphs as the inputs and transform them into the coarse layout information as the intermediate representation, where the scene is depicted by the bounding boxes and annotations\cite{johnson2018image,ashual2019specifying}. Another strategy is generating scene images directly from the coarse layouts input \cite{zhao2019image,sun2019image}. In our paper, rather than taking the layout information as the input of the neural network, we accomplish the image generation by fully utilizing the pre-trained object detection model and back-propagation operation.

\section{Inverting Object Detectors}
\begin{figure}[ht!]
\centering
\includegraphics[width=0.48\textwidth]{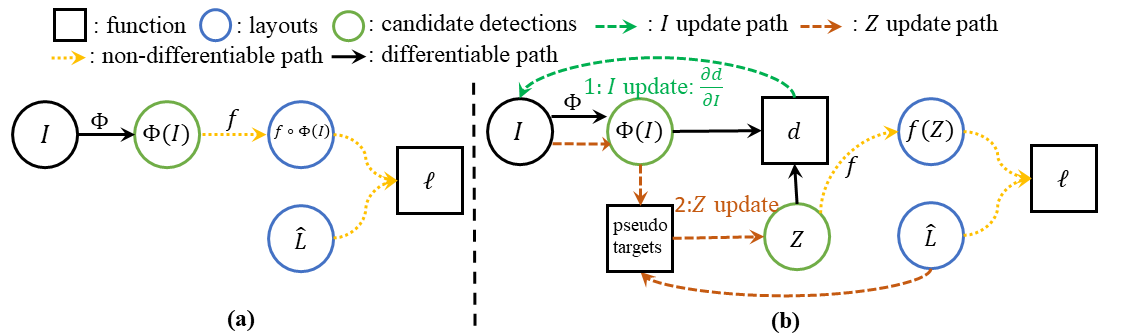}
\vspace{-7mm}
\caption{
(a) Gradient descent cannot be used to invert single-stage detectors since non-differentiable post-processing $f$ blocks gradient flow between the image $I$ and loss $\ell$.
(b) We overcome this problem by introducing auxiliary \emph{pseudo-targets} $Z$. We alternate between updating the image $I$ so the network output $\Phi(I)$ match the pseudo-targets $Z$, and updating $Z$ to be a minimal perturbation of $\Phi(I)$ that would give the target layout $\hat L$ after applying $f$.
% (a): the layout inversion problem of single-stage method, in which non-differentiabilities block the gradient flow.
% (b): our proposed approach with auxiliary variable $Z$ and alternatively updating between $Z$ and $I$.
% $Z$ can be regarded as \emph{pseudo-targets} for the raw detector outputs $\Phi(I)$.
% $I$ is optimized to minimize the distance between raw detector outputs $\Phi(I)$ and pseudo-targets $Z$;
% $Z$ updating is to find suitable pseudo-targets which is the smallest change to $\Phi(I)$ and also able to result in the target layout.
}
\vspace{-5mm}
\label{fig:system}
\end{figure}

Our goal in this section is to develop a general method for inverting universal object detectors.
This is challenging because modern detectors comprise both differentiable neural network layers and non-differentiable operators which preclude the use of na\"{i}ve  gradient-based optimization commonly used to invert image classification models.

We overcome this challenge with an alternating optimization procedure inspired by the alternating direction method of multipliers (ADMM)~\cite{boyd2011distributed}.
Introducing additional auxiliary variables splits the intractable objective to differentiable terms that can be optimized via gradient ascent, and non-differentiable terms that can be optimized heuristically.

\subsection{Problem Setup}
Considered as a black box, an object detector is a function $\mathcal{F}$ which inputs an image $I$ and returns a scene layout $L$ as a collection of detected objects: $\mathcal{F}(I) = L = \{(c_i, b_i, m_i)\}$.
Each object has category label $c_i\in\mathcal{C}$ from a fixed set of categories $\mathcal{C}$, a bounding box $b_i=(x_i, y_i, h_i, w_i)$, and (optionally, if the model performs instance segmentation) a binary segmentation mask $m_i$. 

Given a trained detector $\mathcal{F}$, we wish to introspect its decisions by inverting it: starting from a target layout $\hat{L}$, find an image $I^*$ such that $\mathcal{F}(I^*) = \hat{L}$.
By analogy with methods commonly used to visualize image classification models~\cite{simonyan2013deep,mahendran2016visualizing},
one path forward might be defining a loss function $\ell$ between the detector's output and the desired layout, an image regularizer $R$ encouraging natural images, and optimize
via gradient descent:
\begin{equation}
    I^* = \arg\min_I \ell(\mathcal{F}(I), \hat{L}) + R(I)
    \label{eq:naive-inversion} 
  \end{equation}
  
However, this method cannot work when the detector $\mathcal{F}$ involves non-differentiable components such as eliminating detections below a confidence threshold, and non-maximum suppression (NMS) to prevent duplicate detections.
As shown in Figure~\ref{fig:system}, these non-differentiabilities block the gradient flow between the loss and the input image, precluding the use of na\"{i}ve gradient ascent.

We proceed by writing the detector $\mathcal{F}$
as a composition of differentiable neural networks $\Phi_i$ and nondifferentiable steps $f_i$.
Single-stage detectors can generally be written in the form $\mathcal{F}(I) = f_1(\Phi_1(I))$, while two-stage detectors can be written as $\mathcal{F}(I)=f_2(\Phi_2( f_1(\Phi_1(I)), I))$.
Like ADMM, we introduce auxiliary variables $Z_i$ and constraints encouraging $Z_i$ to be equal to the output of $\Phi_i$.
We then minimize via alternating optimization, updating $I$ and each $Z_i$ in turn.
We will describe the setup for single-stage detectors in detail, then summarize its extension to two-stage methods; see the supplementary material for a more explicit discussion.

\begin{figure}
    \includegraphics[width=0.48\textwidth]{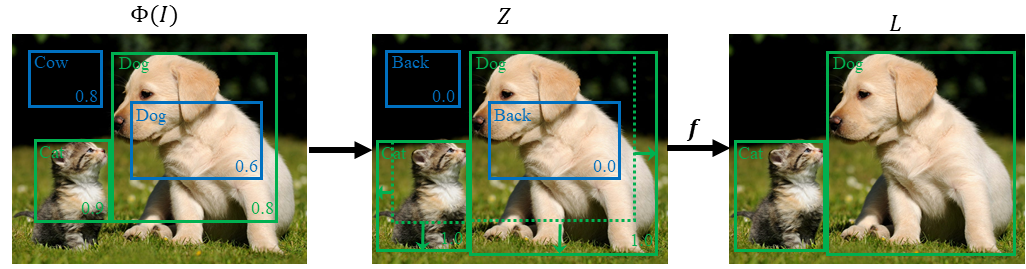}
    \vspace{-5mm}

    \caption{
    % \Ang{
    % % We show the projection process with several candidate detection examples. 
    % % The predicted probabilities and categories are annotated at the right-bottom and left-top of boxes.
    % % Green boxes represent matched detections~(with the highest IoU with ground truth objects),
    % % blue boxes represent unmatched detections. 
    % % $Z$ is a modified version of $\Phi(I)$, where matched detections~(blue boxes) are assigned to gt category labels and boxes,
    % % and unmatched detections~(green boxes) are modified as backgrounds.  
    % % After projection, $L = f(Z) = \hat{L}$.
    % % We optimize $I$ to minimize difference between $\Phi(I)$ and $Z$ so that $f \circ \Phi(I) = \hat{L}$ eventually.
    % % }
    % }
    We show updating the pseudo-targets $Z$ of raw network outputs $\Phi(I)$ with several candidate detection examples.
    The predicted probabilities and categories are annotated at the right-bottom and left-top of boxes.
    Green boxes represent matched detections~(with the highest IoU with ground truth objects), 
    and blue boxes represent unmatched detections~(with low IoU and kept after post-processing).
    The ground truth categories and box positions are assigned to matched detections~(green boxes),
    while unmatched detections~(blue box) are assigned background categories as classification targets.
    $Z$ results in $f(Z) =\hat{L}$.
    }
    \vspace{-5mm}
    \label{fig:projection}
\end{figure}

\subsection{Inverting Single-Stage Detectors}
Let $\mathcal{F}(I)=f(\Phi(I))$ be a single-stage detector.
$\Phi$ is a network receiving the image $I$ and outputing a fixed-size tensor $\Phi(I)$ of shape $N\times(D_{cls}+D_{reg})$ giving raw classification scores and box positions for $N$ candidate detections.
These raw outputs are converted into a set of detected objects by non-differentiable post-processing $f$.

Inspired by ADMM, we introduce auxiliary variables $Z$ with the same shape as $\Phi(I)$.
This allows us to rewrite Equation~\ref{eq:naive-inversion} as a constrained optimization problem:
\begin{equation}
  I^*, Z^* = \arg\min_{I,Z} \ell(f(Z), \hat L) + R(I) \textrm{ with } Z = \Phi(I)
  \label{eq:constrained}
\end{equation}
We can then relax this back to an unconstrained problem:
\begin{align}
  & I^*, Z^* = \arg\min_{I,Z} L_\lambda(I,Z) \label{eq:splitting_single} \\
  & L_{\lambda}(I,Z) = \ell(f(Z), \hat{L}) + \lambda d(Z,\Phi(I)) + R(I).
\end{align}
Here $\lambda>0$ is a hyperparameter and $d$ is a differentiable distance function.
We choose $\ell$ to be an indicator function that is $0$ when $f(Z)=\hat L$ and $+\infty$ otherwise.
We now minimize Equation~\ref{eq:splitting_single} by alternating updates to $I$ and $Z$:
\begin{align}
  I^{k+1} &= \arg\min_I L_{\lambda}(I, Z^{k}) \\
    &= \arg\min_I \lambda d(Z^k, \Phi(I)) + R(I) \label{eq:I-update} \\
  Z^{k+1} &= \arg\min_Z L_{\lambda}(I^{k+1}, Z) \\
    &= \arg\min_Z \ell(f(Z),\hat L)+\lambda d(Z,\Phi(I^{k+1})) \label{eq:Z-update}
\end{align}
We can update $I$ via gradient descent since all terms of Equation~\ref{eq:I-update} are differentiable.
Since $\ell$ is an indicator, Equation~\ref{eq:Z-update} is equivalent to finding $Z$
which minimizes $d(Z,\Phi(I^{k+1}))$ subject to $f(Z)=\hat L$.
In other words, updating $Z$ amounts to finding a \emph{minimal perturbation} (measured by $d$) of the current raw detector outputs $\Phi(I)$ that would produce the target layout $\hat L$ after post-processing.

Intuitively, we can think of $Z$ as \emph{pseudo-targets} for the raw outputs of the network $\Phi$.
We alternate between updating the image $I$ so the network outputs $\Phi(I)$ are closer to the pseudo-targets,
and updating the pseudo-targets $Z$ to be the smallest change to the current network outputs $\Phi(I)$ which would result in the target layout.

\textbf{Updating Pseudo-Targets.}
We update the pseudo-targets $Z$ heuristically by analyzing the post-processing $f$:
it filters out most candidate detections, and few detections are kept in the final layout.
Therefore, the pseudo-targets $Z$ for raw detector outputs $\Phi(I)$  with a minimal perturbation
can be reached by assigning ground-truth categories and positions as targets to several detections in $\Phi(I)$ 
while letting targets equal to outputs for other detections. 

Like Figure~\ref{fig:projection},  we match ground-truth instances in target layouts with candidate detections of the highest IoU and assign the ground-truth categories and box positions as targets to these detections. 
For candidate detections which are not matched to any ground-truth instances and would be kept after post-processing,  we assign background as target predictions so that it would be eliminated by post-processing.

% We empirically find that the minimal distance constraints can be slightly relaxed: 
% multiple candidate detections can be matched to one object to improve the recognition of images.

\textbf{Loss Function.}
We use a differentiable loss $d$ to measure the agreement between the raw network outputs $\Phi(I)$ and the pseudo-targets $Z$.
Inspired by losses commonly used to train object detectors, we define $d$ as a weighted combination of a cross-entropy loss $\ell_{cls}$
% between the per-box classification scores in $\Phi(I)$ and labels in $Z$,
and a GIoU loss~\cite{rezatofighi2019generalized} $\ell_{reg}$:
% between the positions of the non-background boxes as specified by $Z$:
% We specify the differentiable loss function $d$. 
%We use the popular cross entropy loss $\ell_{cls}$ and GIoU loss~\cite{rezatofighi2019generalized} $\ell_{reg}$ to measure the distance 
%of classifications and box positions respectively between $Z$ and $\Phi(I)$:
\begin{equation}
    \scalebox{0.94}{$
    d(\Phi(I), Z) = \lambda_{cls}\ell_{cls}(\Phi(I), Z) + \lambda_{reg}\ell_{reg}(\Phi(I), Z)
    %\ell(f(Z), \hat{L}) + \lambda \ell_{cls}(\Phi(I), Z) + \lambda \ell_{reg}(\Phi(I), Z)+R(I)
    $}
\end{equation}
% Following the convention of classification network visualization, we also maximize the pre-sigmoid value of classification predictions which are kept after post-processing to improve the visualization quality.

\subsection{Inverting Two-Stage Detectors}
We use the same idea to invert two-stage detectors by introducing auxiliary variables $Z_{1}$, $Z_{2}$ with constraints $Z_{1} = \Phi_{1}(I)$ and $Z_{2} = \Phi_{2}(f_{1}(\Phi_{1}(I)),I)$.
$\Phi_{1}(I)$ is the output of RPN network $\Phi_1$, and $\Phi_{2}(f_{1}(\Phi_{1}(I)),I)$ is the prediction result of $\Phi_2$ on every RoI. 
Loss $\ell_{rpn}(f_{1}(\Phi_{1}(I)), \hat{L})$ is added to improve the quality of proposed RoIs:    
\begin{align}
\begin{split}
    \ell(f_{2}(Z_2), \hat{L}) + \ell_{rpn}(f_{1}(Z_1), \hat{L})+\lambda_1 \ell_{1}(Z_1, \Phi_{1}(I))  \\
    +\lambda_2 \ell_{2}(Z_2, \Phi_{2}(f_{1}(\Phi_{1}(I)), I)) + R(I) 
\end{split}
    \label{eq:splitting_two_stage}
\end{align}
This could be solved by alternating optimization between $I, Z_1$ and $Z_2$ following the same manner as single-stage methods. 
Intuitively, $Z_1$ and $Z_2$ are the pseudo-targets for outputs of network $\Phi_1$ and $\Phi_2$.
$\ell_{1}$ and $\ell_2$ are the loss functions between network outputs and pseudo-targets for $\Phi_1$ and $\Phi_2$ respectively.
See more details in supplementary.
% Intuitively, $Z_1$ is the pseudo-target for outputs of RPN network and $Z_2$ is the pseudo-target for the prediction results on every RoI.
% % We optimize the input image $I$ to minimize the loss for target layouts. 
% See more details in supplementary.

\section{Experiments}
% We use layout inversion as our primary tool to visualize object detectors,
% where the inverted images reveal the learned features and representative visual cues for detections. 
% On the other hand, unlike visualizing classification networks, layout inversion is a highly structured task, with multiple objects and different predictions~(category, regression, and mask). 
% To better understand object detectors,  it is essential to disentangle these elements and visualize them individually. 
% To this end, we disentangle different losses in \ref{sec:dis_losses} and visualize individual objects in \ref{sec:vis_obj}.

We apply layout inversion as our primary visualization tool to a wide variety of object detectors, where the inverted images reveal the learned features and representative visual cues for detections. 
In Section~\ref{sec:invert}, we show the qualitative difference between inversions of varying detectors.
We further quantify the inversion performance by \emph{inversion transfer} in Section~\ref{sec:inver_trans}, which also reflects the similarity of visual cues learned by different object detectors.

On the other hand, object detection is a highly structured task: predicting both \emph{what} is present and \emph{where} things are located for multiple objects in a scene.
To this end, we disentangle subtasks and invert individual objects. 

We disentangle subtask losses in Section~\ref{sec:dis_losses} by performing layout inversion targeting with loss combinations,  to reveal the visual cues learned by each head.
We also use \emph{attribution} to find image regions responsible for each subtask.
% We further validate this observation by attribution methods and compare image regions responsible for each head. 

% We visualize individual objects in Section~\ref{sec:vis_obj}, which demonstrates that detectors learn canonical motifs of commonly co-occurring objects.
% Moreover, they rely on different features when detecting objects of different scales. 
% These observations are further investigated by following quantitative experiments.
We visualize individual objects in Section~\ref{sec:vis_obj}.
Inspired by observations, we investigate the role of \emph{contexts} in Section~\ref{sec:sur_con} 
and measure the feature difference used for detecting objects with varying scales in Section~\ref{sec:var_scale}.
\begin{figure*}[ht!]
    \centering
    \begin{tabular}{@{}c@{\hspace{0.5mm}}c@{\hspace{0.5mm}}c@{\hspace{0.5mm}}c@{\hspace{0.5mm}}c@{\hspace{0.5mm}}c@{\hspace{0.5mm}}c@{}}
    
    \rotatebox{90}{\small   \hspace{2mm} Layout}
    \includegraphics[width=0.138\textwidth]{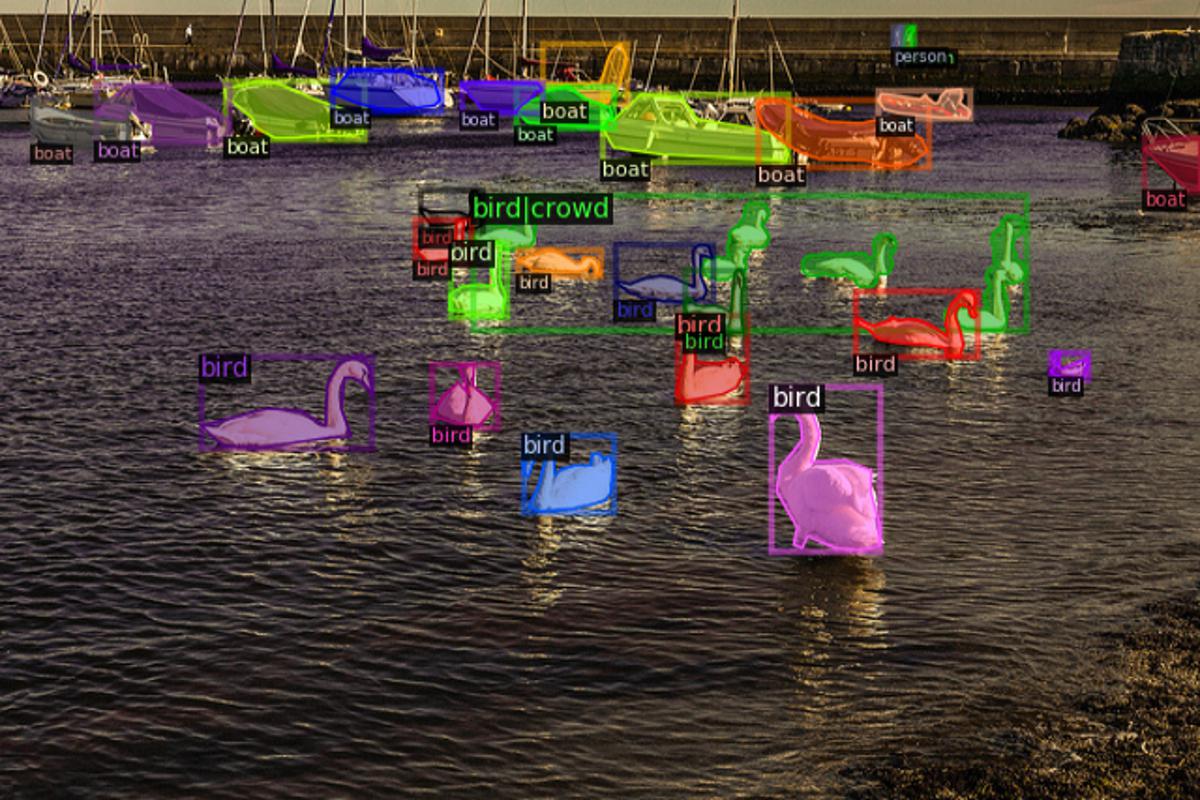}&
    \includegraphics[width=0.138\textwidth]{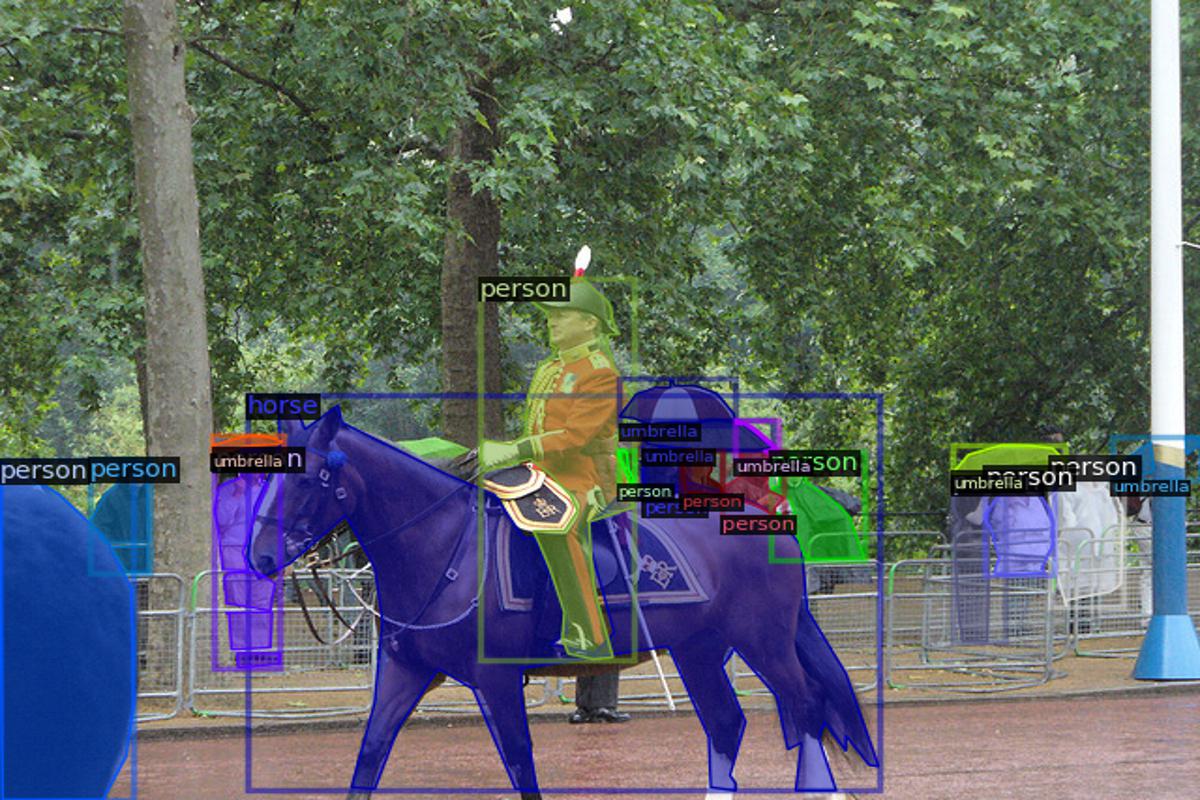}&
    \includegraphics[width=0.138\textwidth]{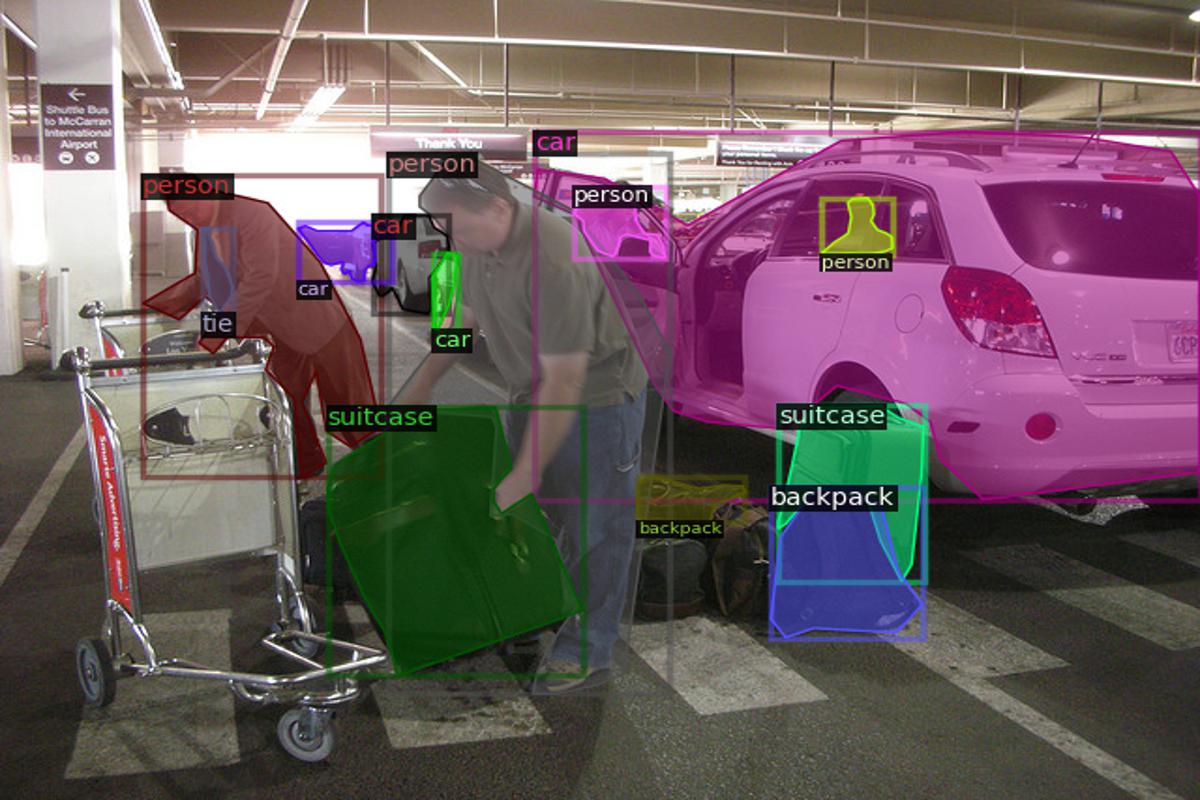}&
    \includegraphics[width=0.138\textwidth]{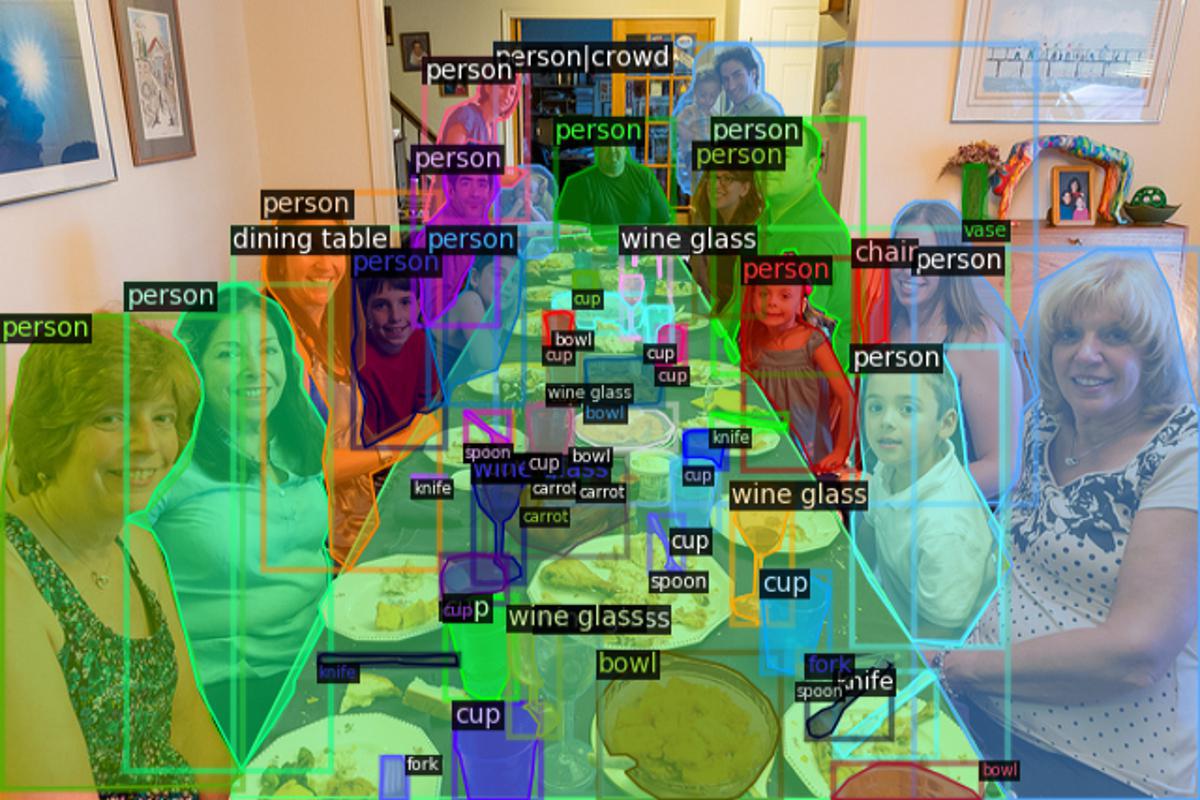}&
    \includegraphics[width=0.138\textwidth]{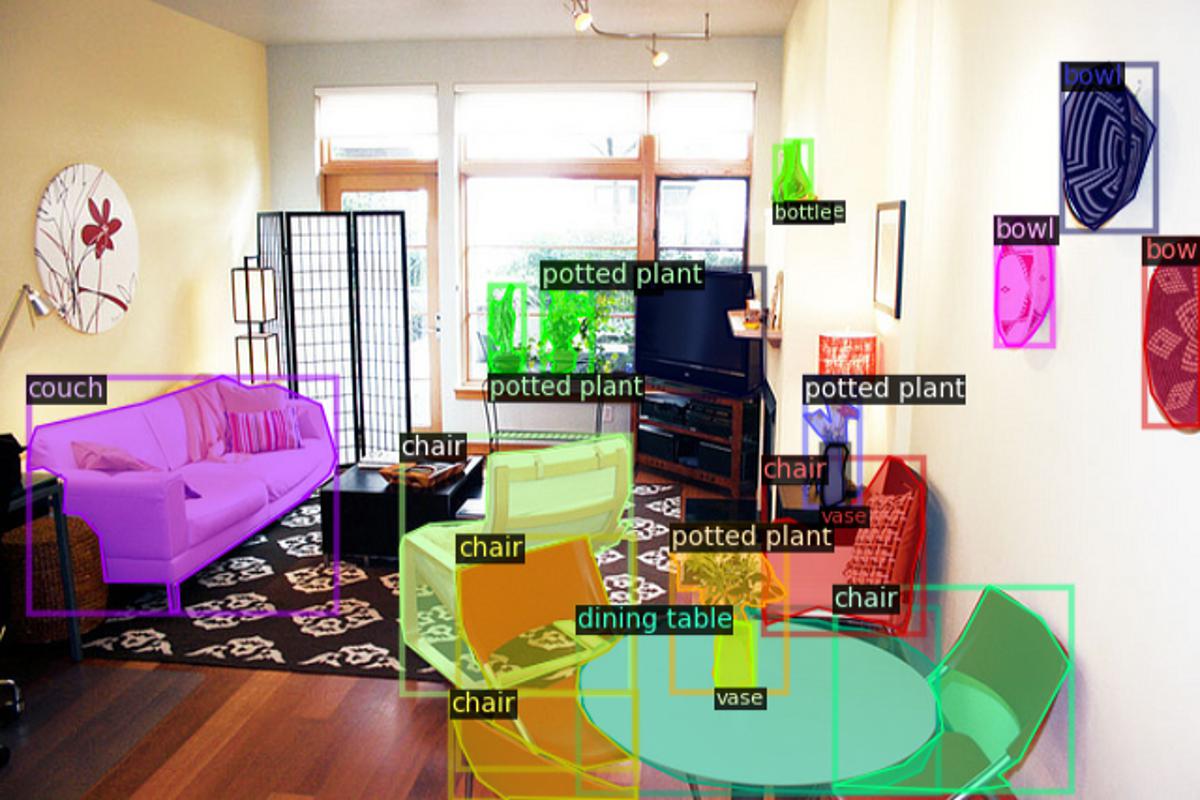}&
    \includegraphics[width=0.138\textwidth]{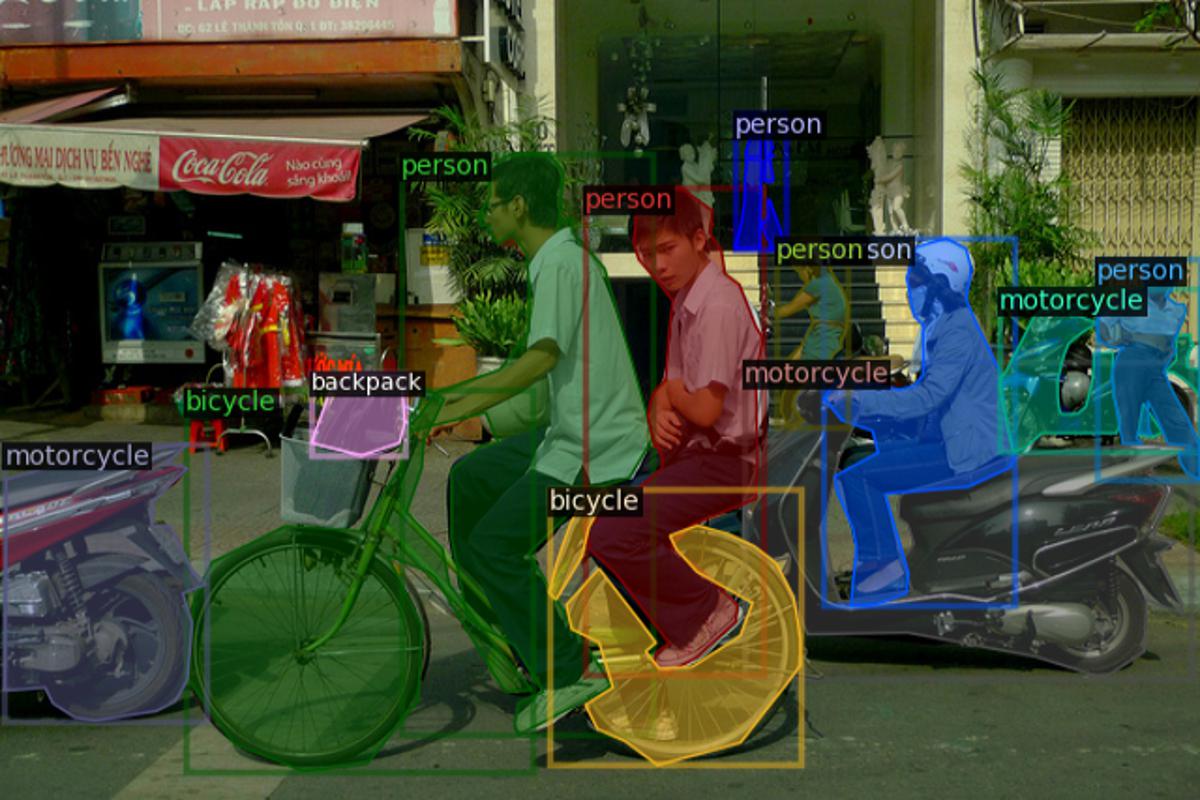}&
    \includegraphics[width=0.138\textwidth]{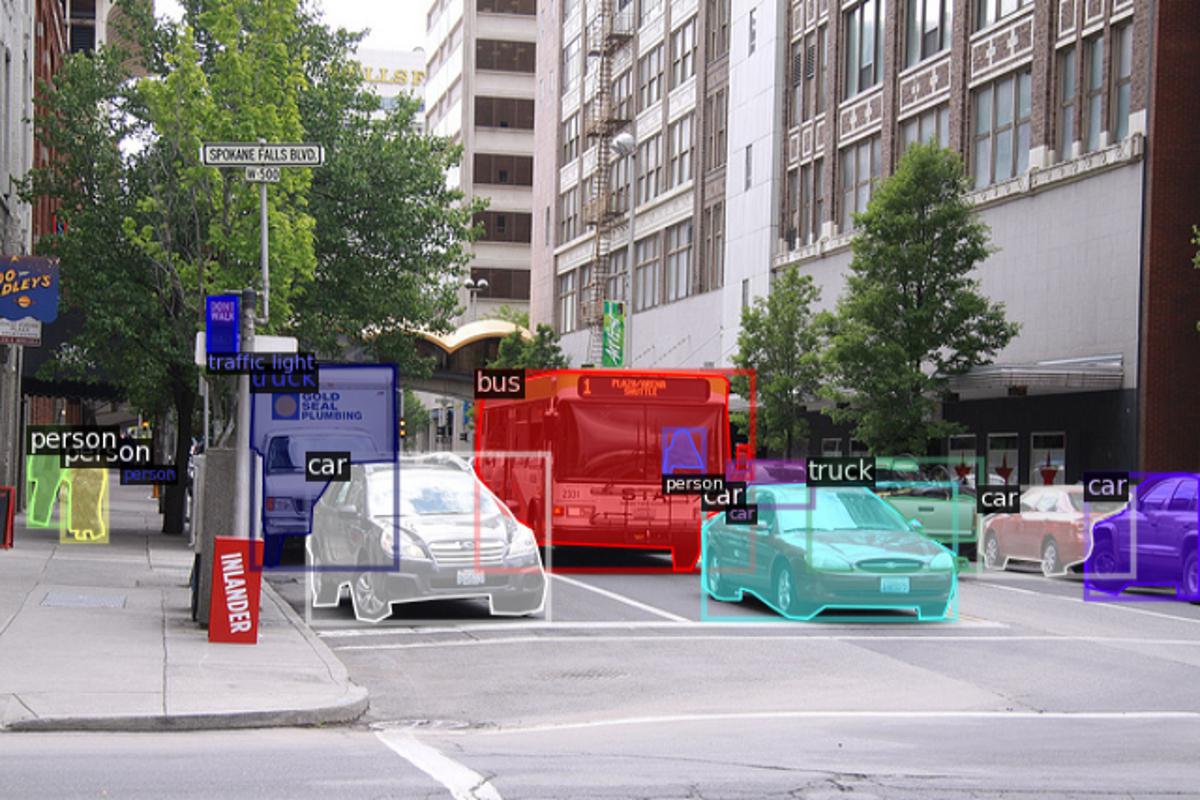}\\*[-0.5mm]
    \rotatebox{90}{\small   \hspace{3mm} Mask}
    \includegraphics[width=0.138\textwidth]{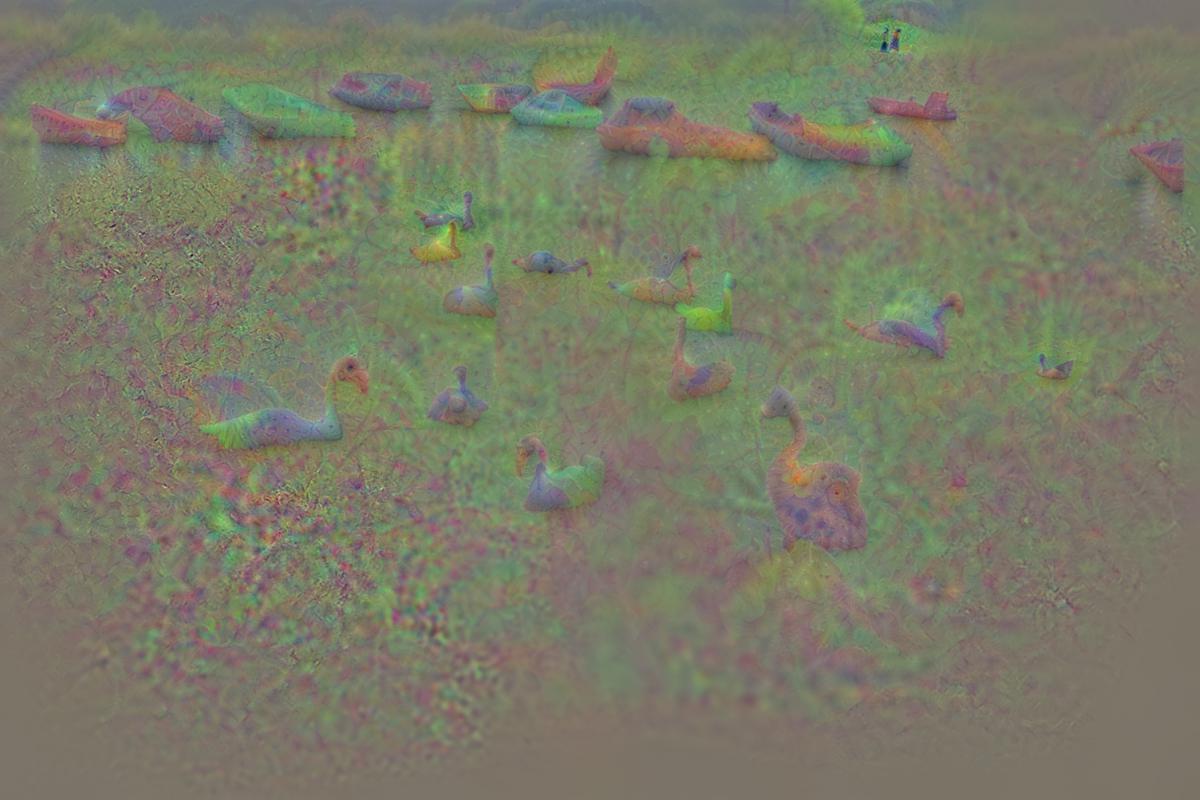}&
    \includegraphics[width=0.138\textwidth]{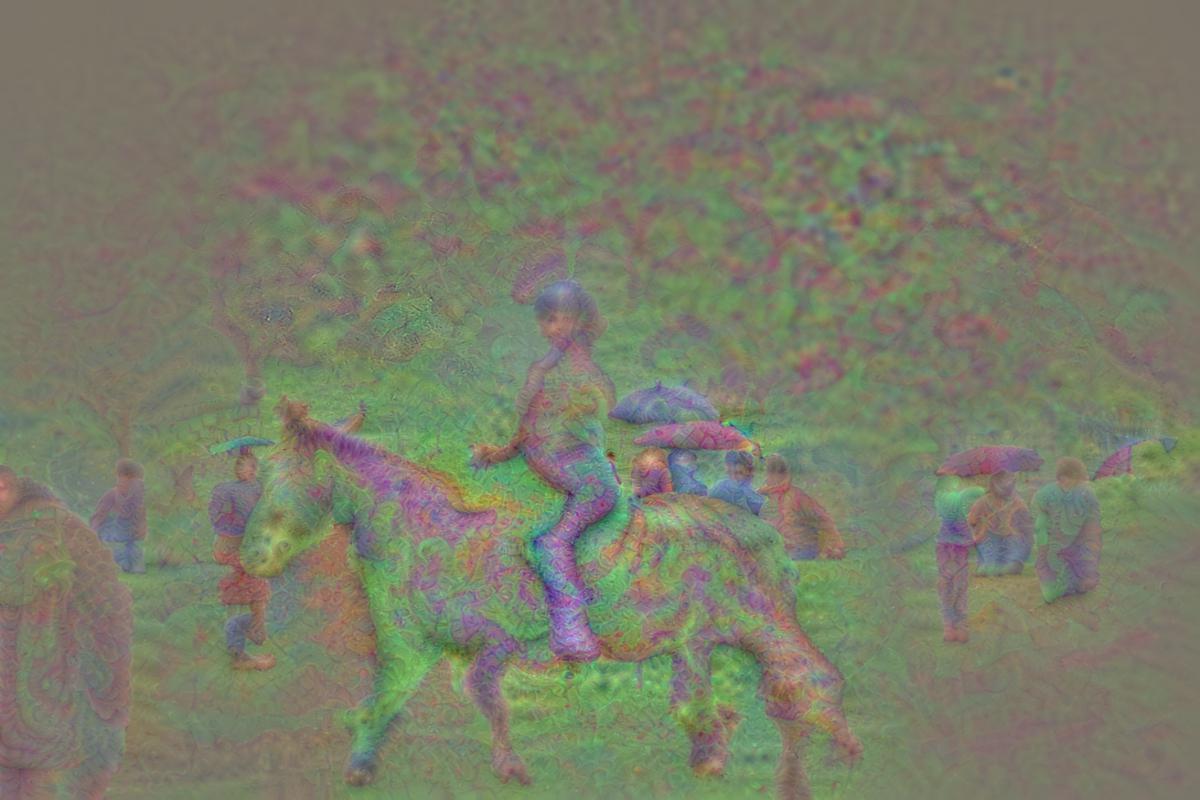}&
    \includegraphics[width=0.138\textwidth]{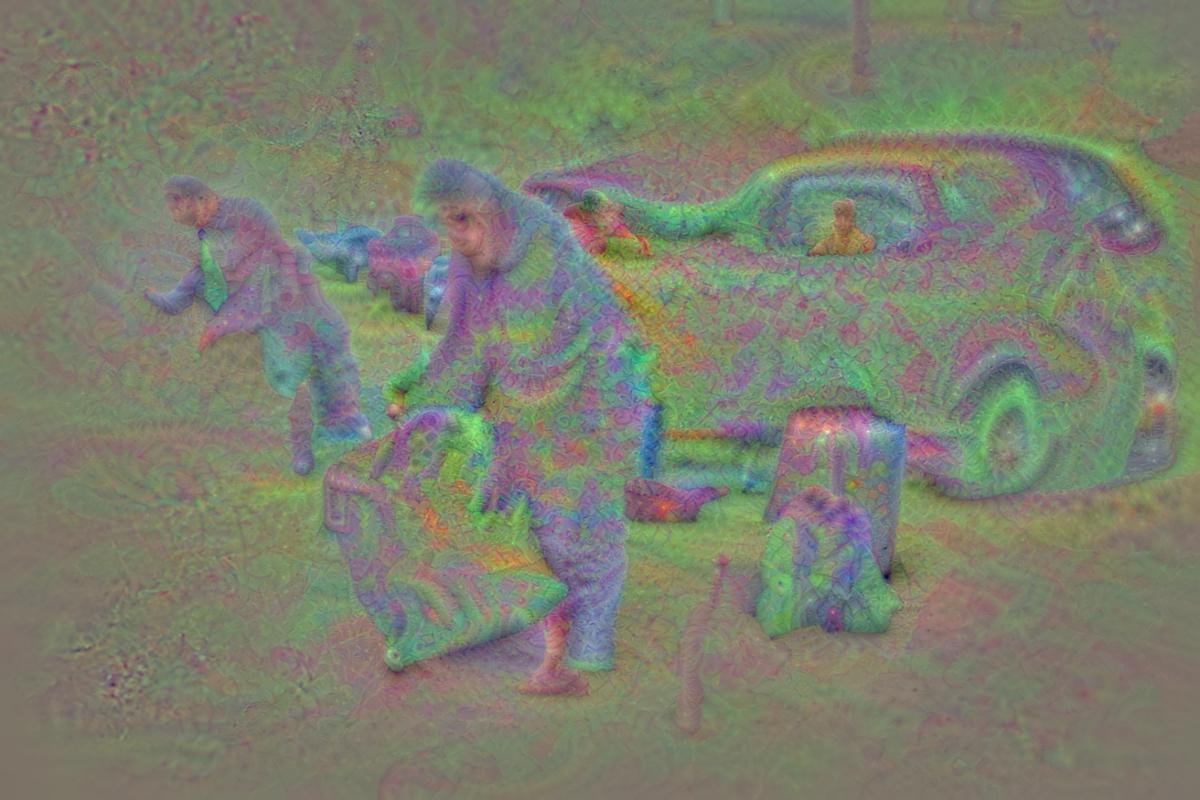}&
    \includegraphics[width=0.138\textwidth]{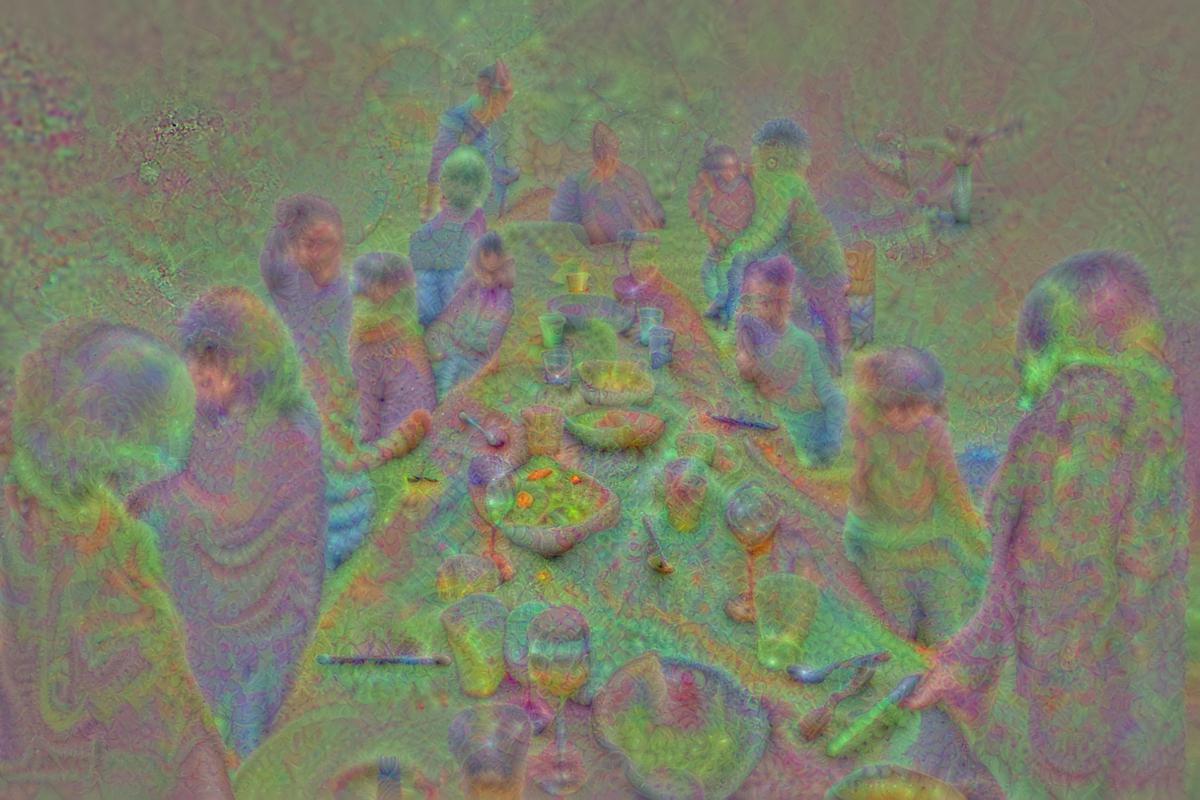}&
    \includegraphics[width=0.138\textwidth]{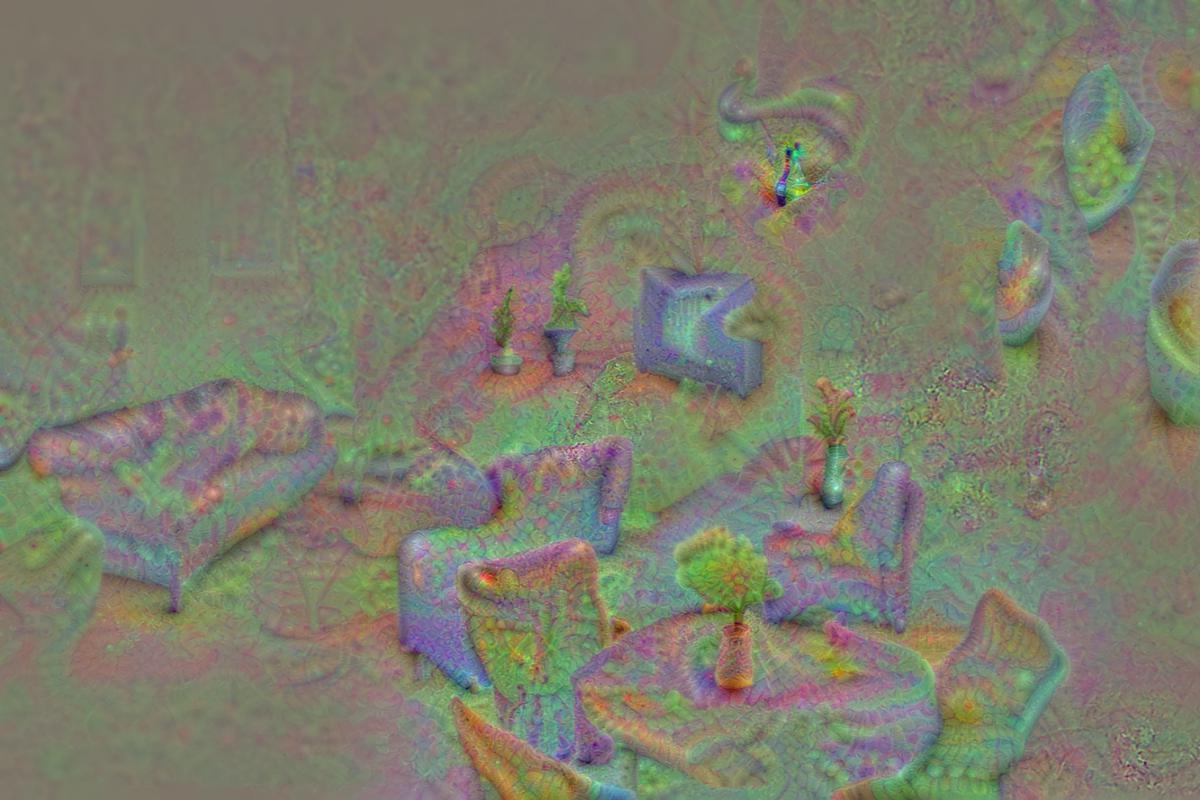}&
    \includegraphics[width=0.138\textwidth]{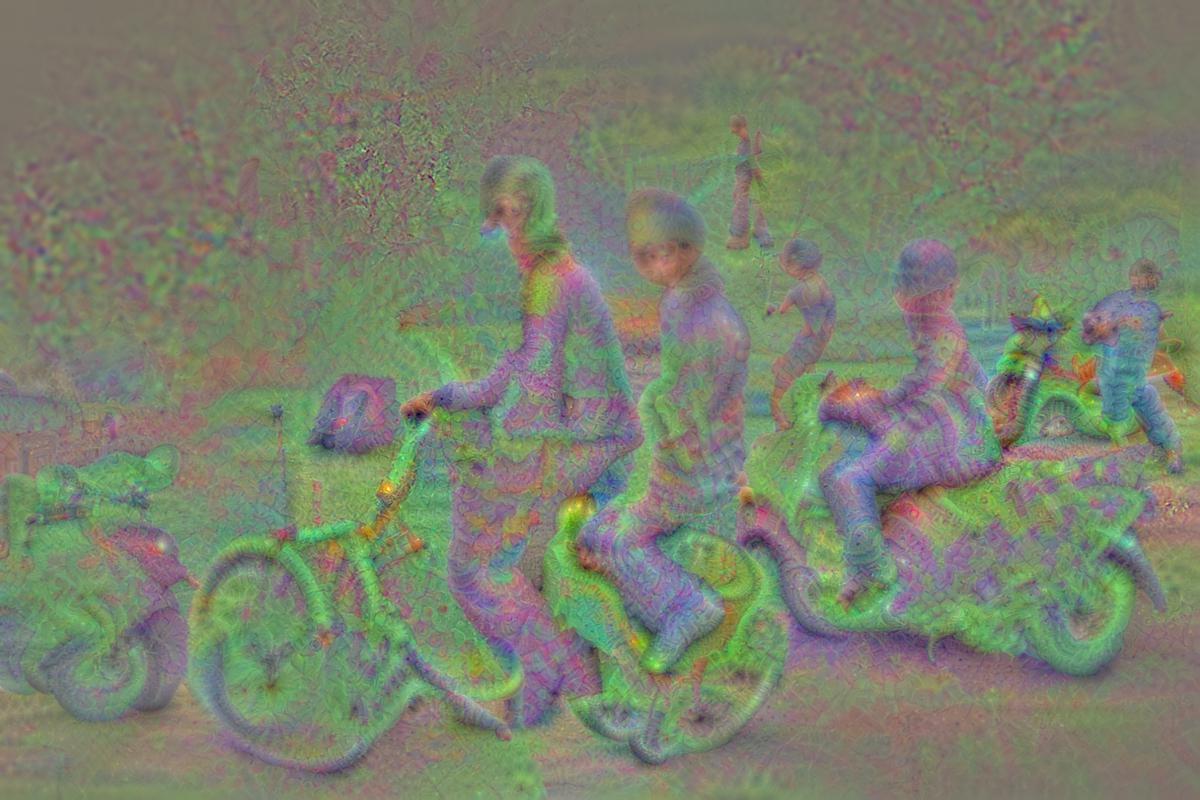}&
    \includegraphics[width=0.138\textwidth]{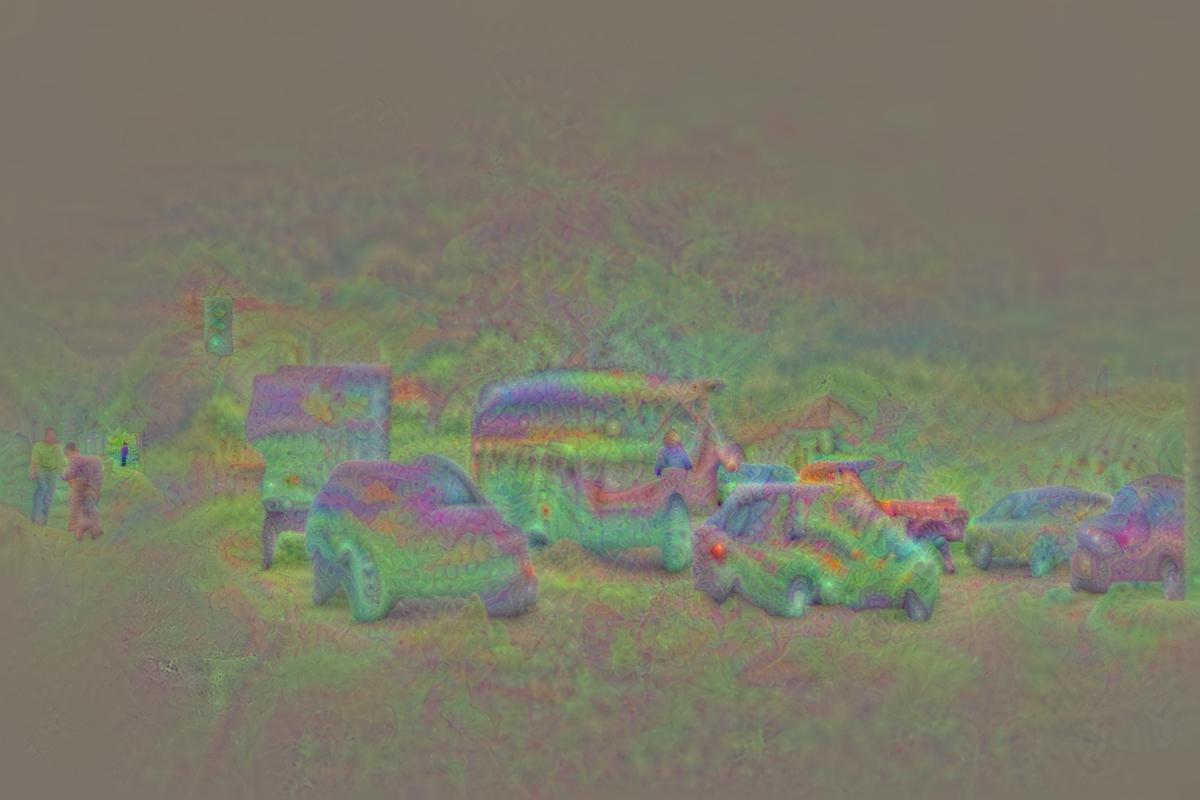}\\*[-0.5mm]
    \rotatebox{90}{\small   \hspace{3mm} Faster}
    \includegraphics[width=0.138\textwidth]{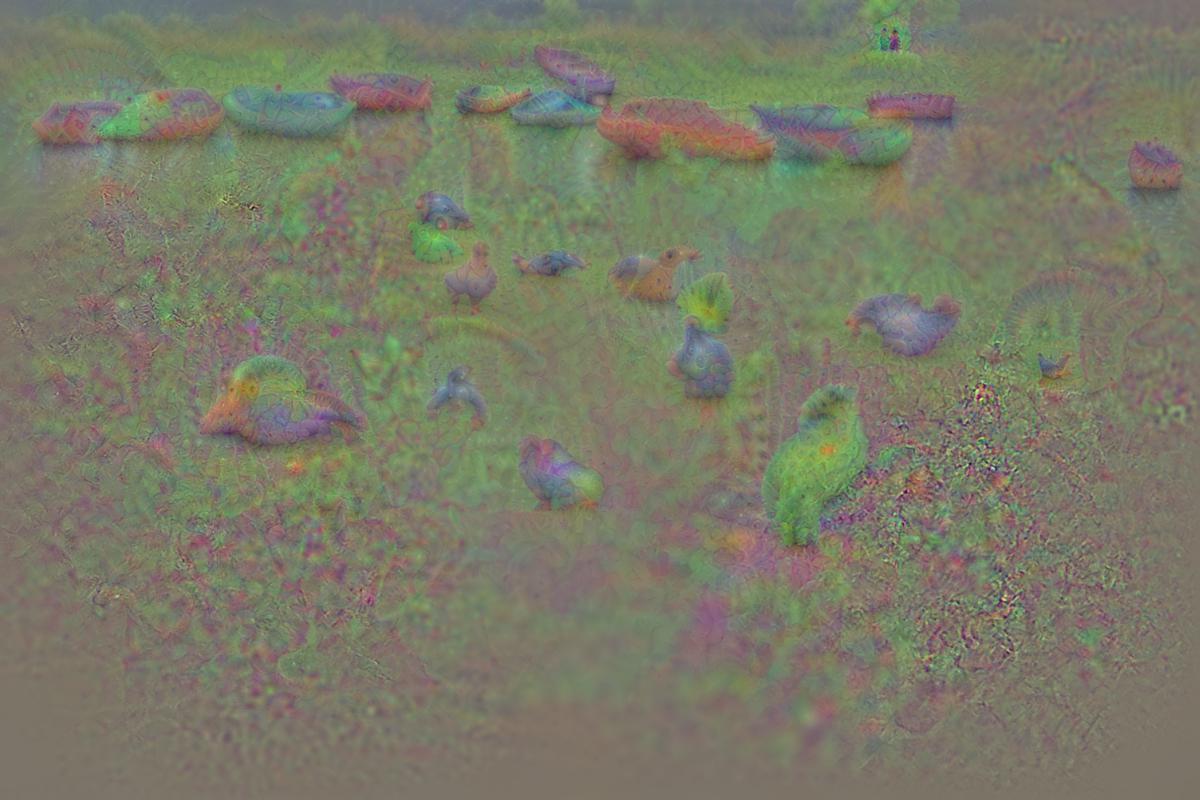}&
    \includegraphics[width=0.138\textwidth]{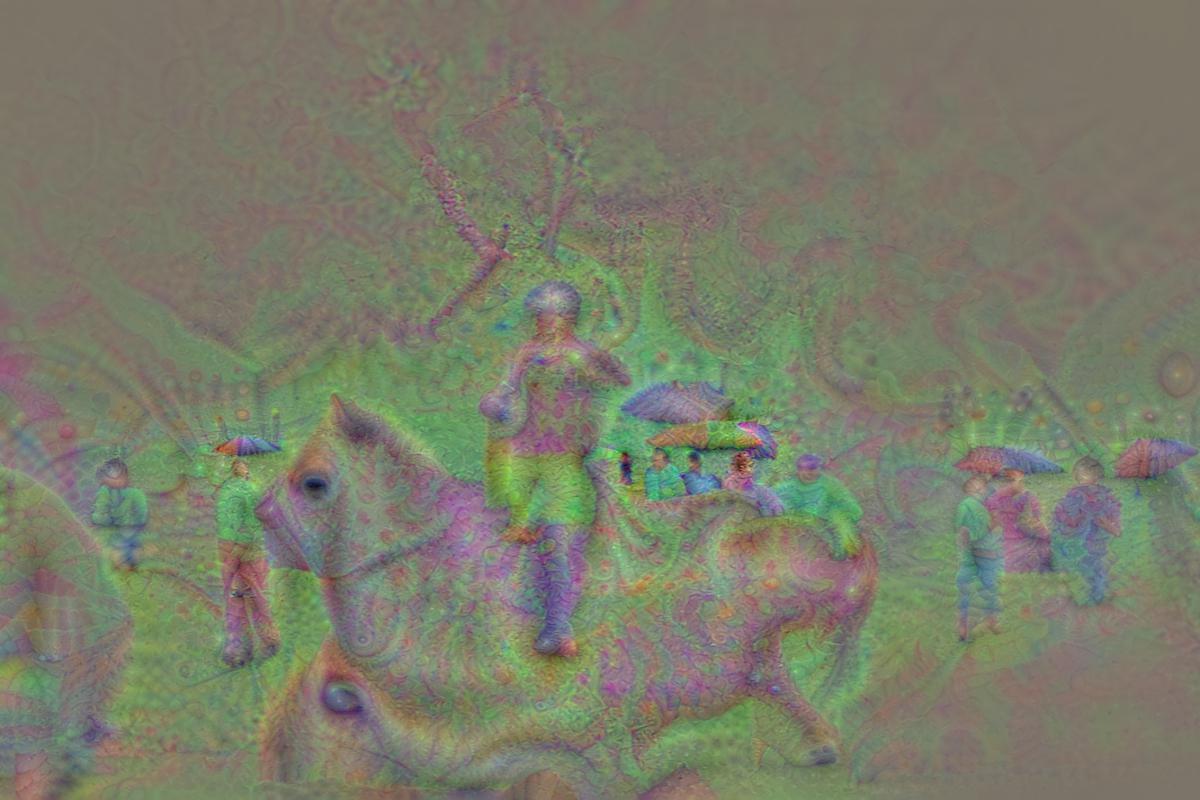}&
    \includegraphics[width=0.138\textwidth]{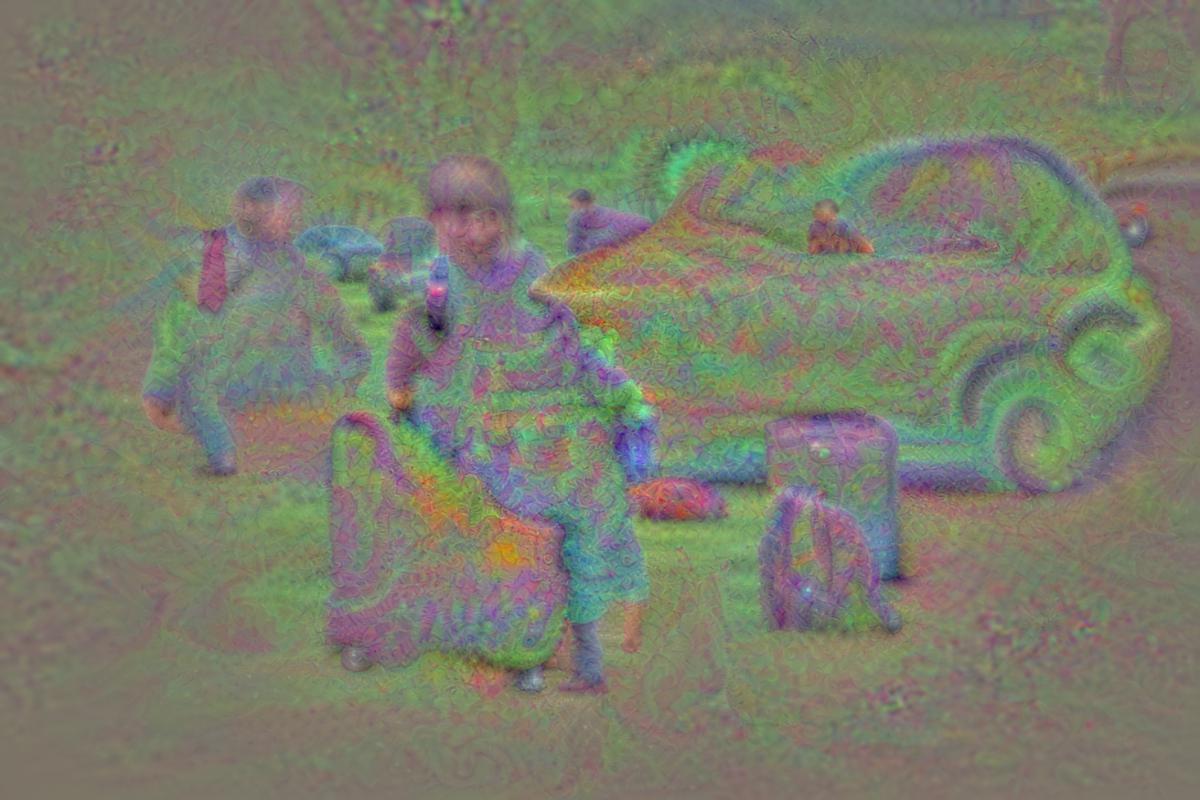}&
    \includegraphics[width=0.138\textwidth]{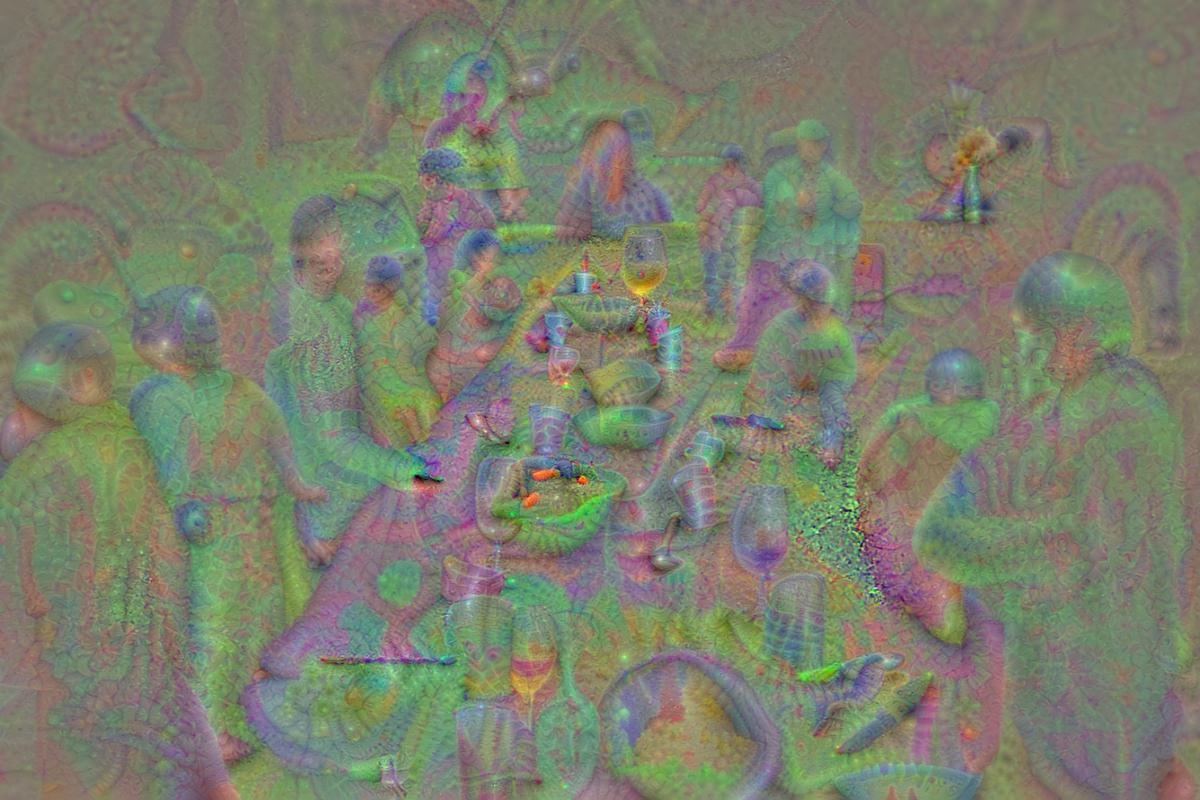}&
    \includegraphics[width=0.138\textwidth]{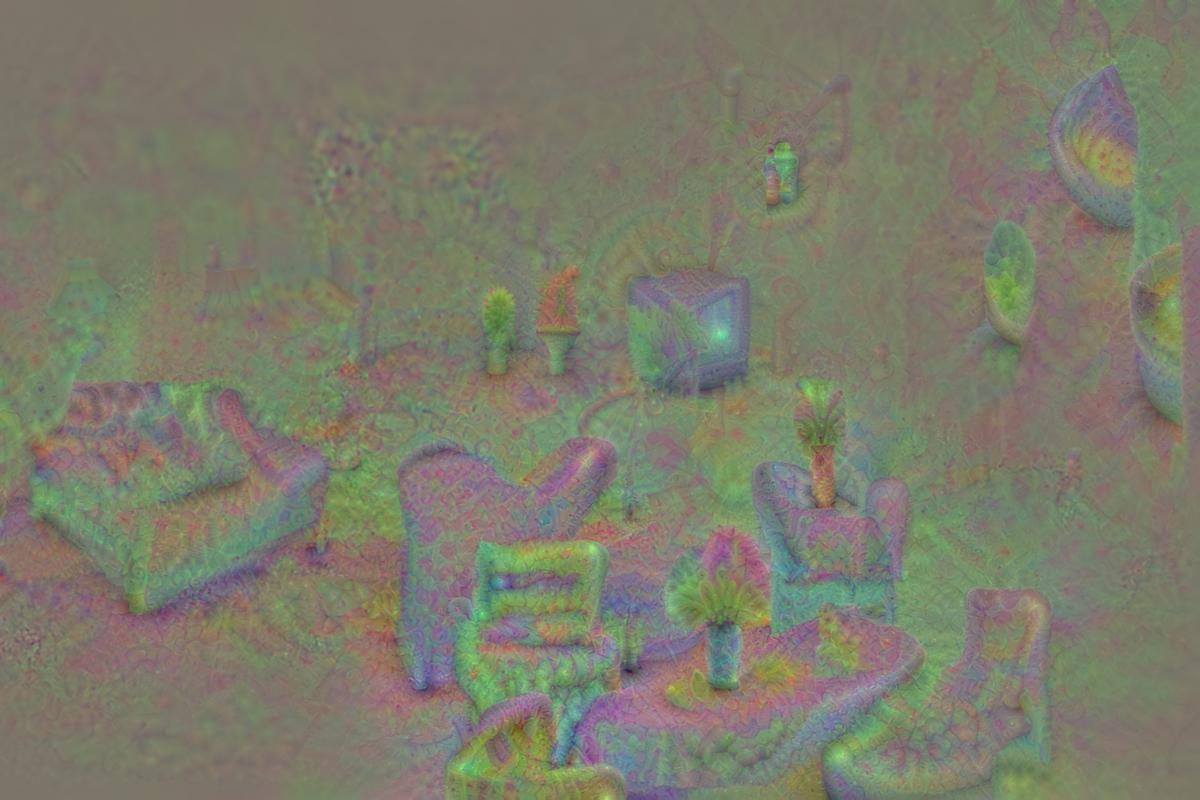}&
    \includegraphics[width=0.138\textwidth]{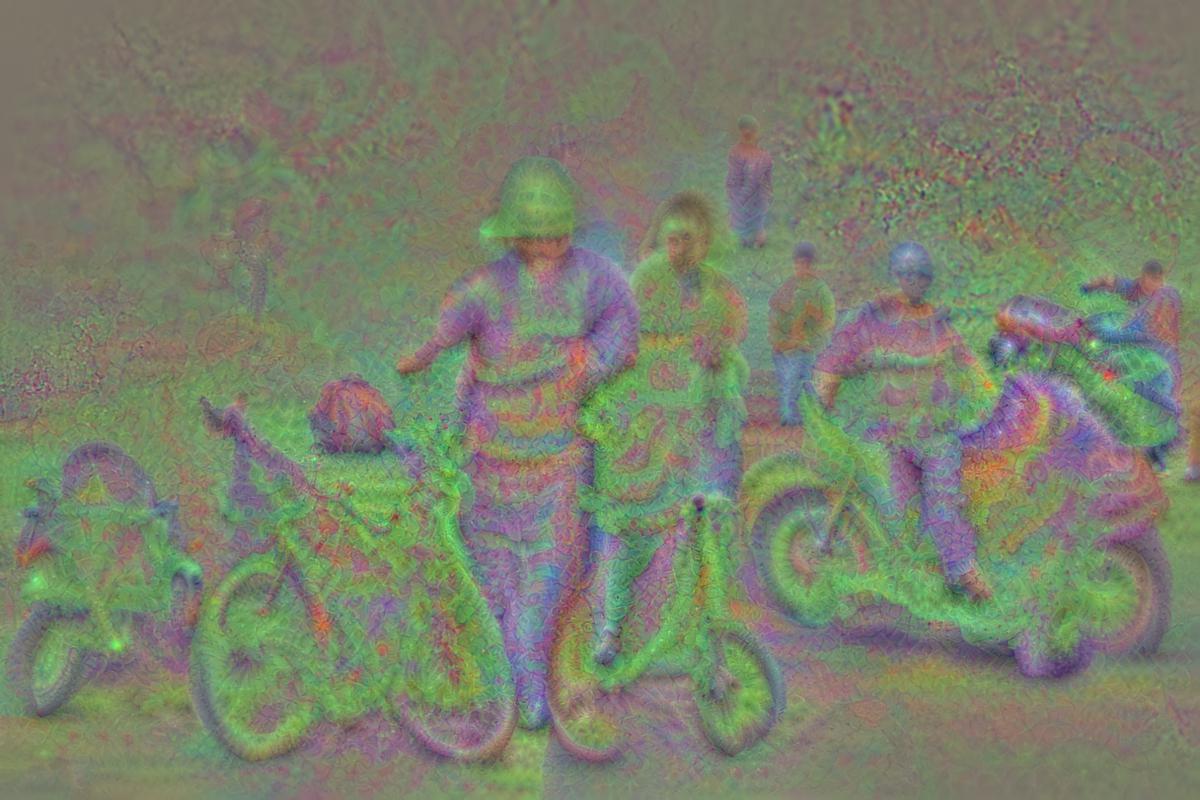}&
    \includegraphics[width=0.138\textwidth]{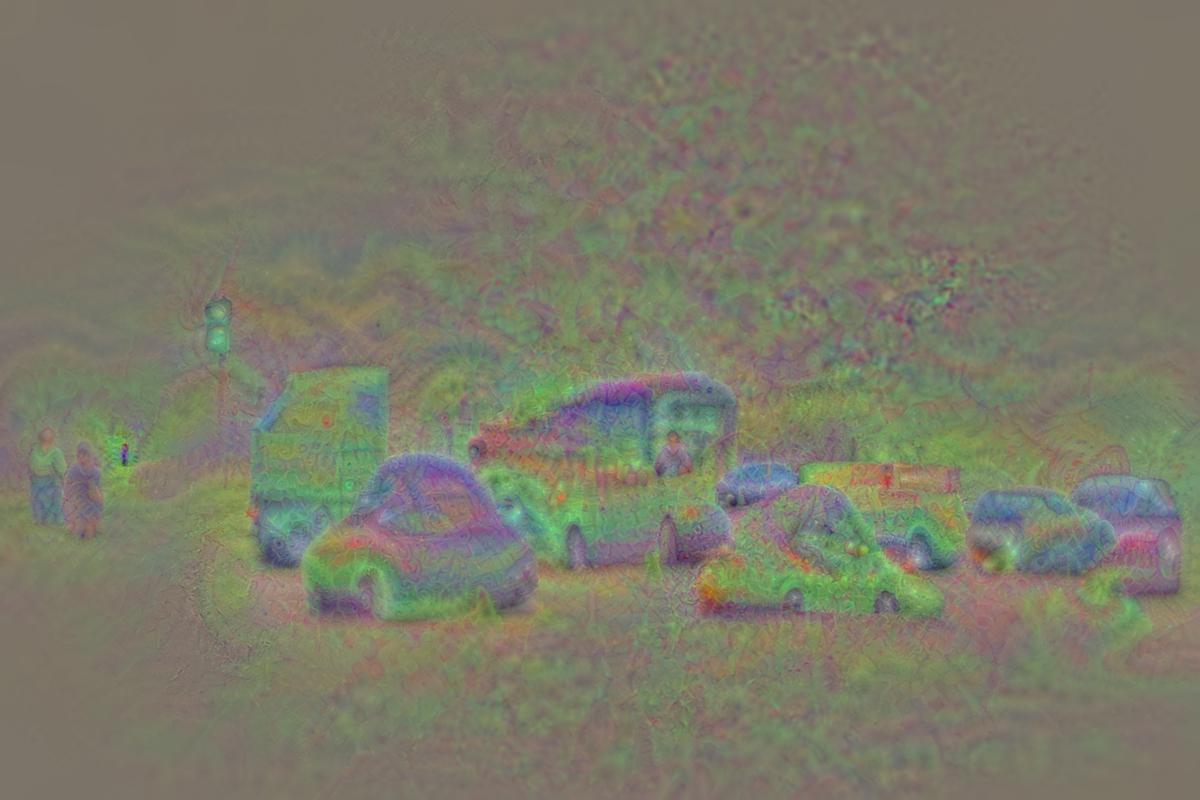}\\*[-0.5mm]
    \rotatebox{90}{\small  \hspace{2.5mm} Retina}
    \includegraphics[width=0.138\textwidth]{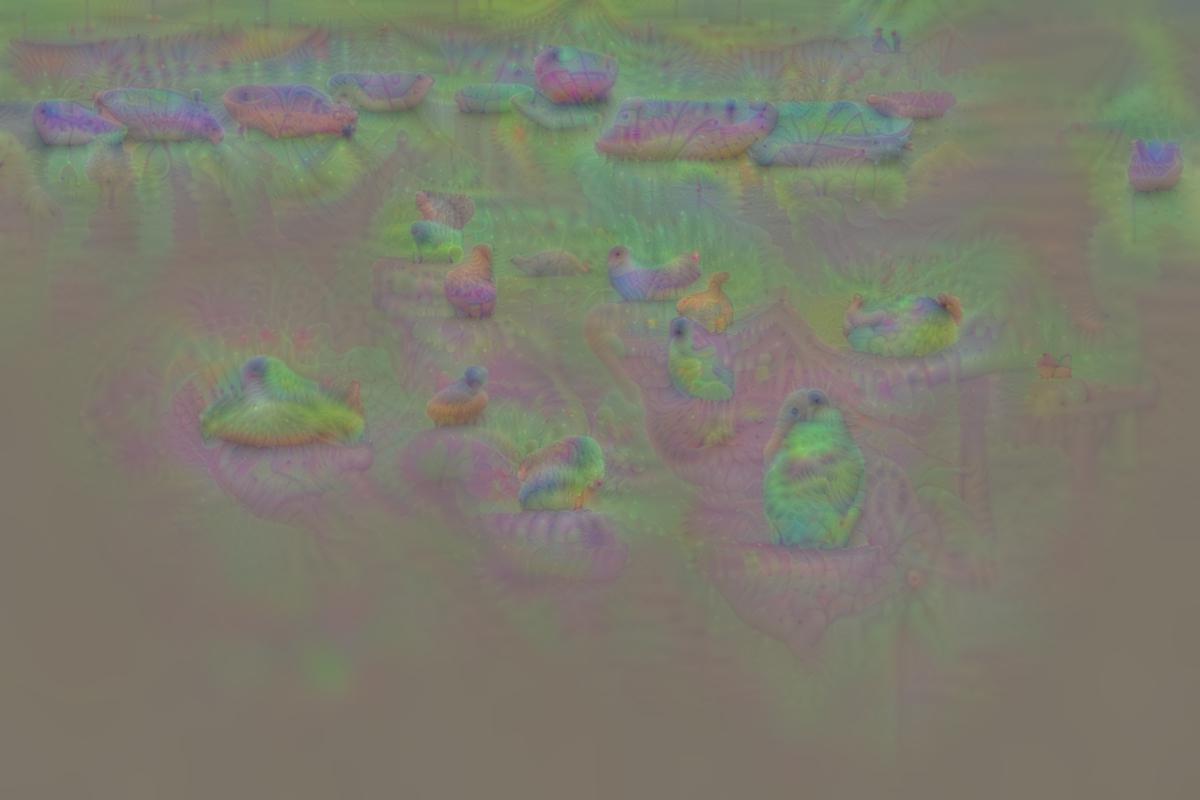}&
    \includegraphics[width=0.138\textwidth]{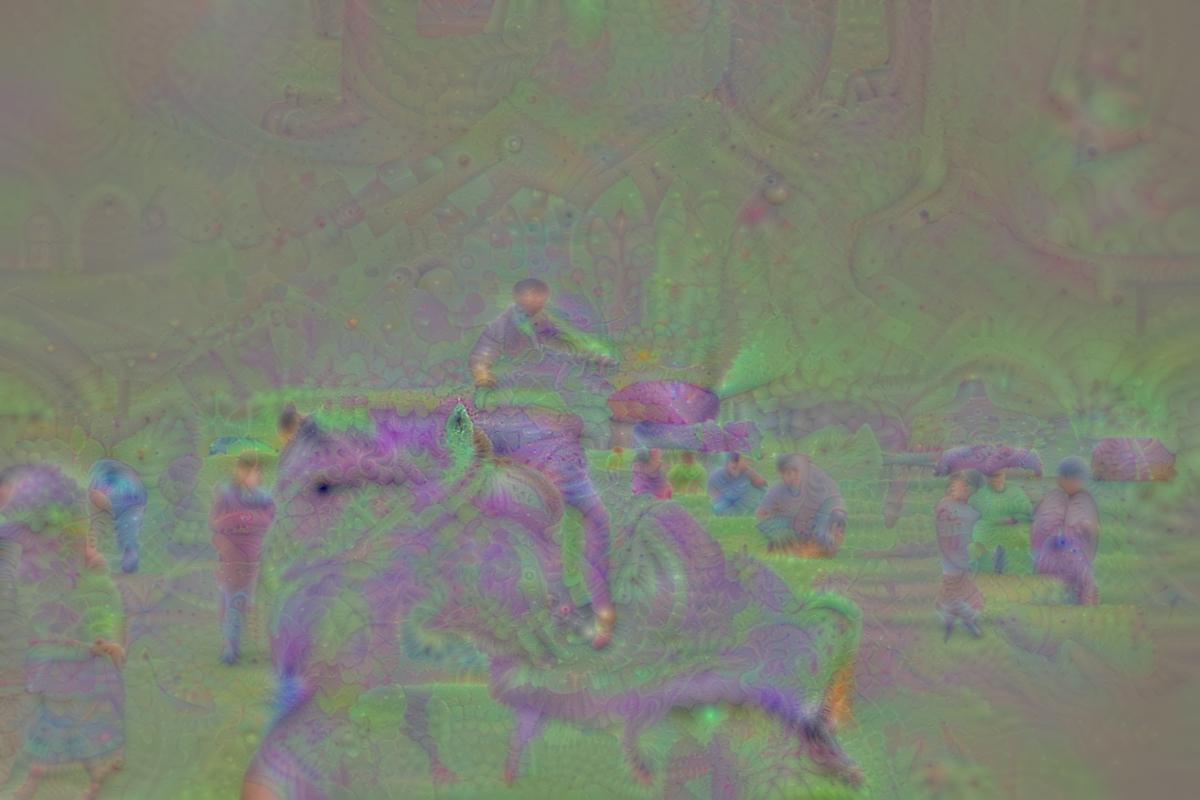}&
    \includegraphics[width=0.138\textwidth]{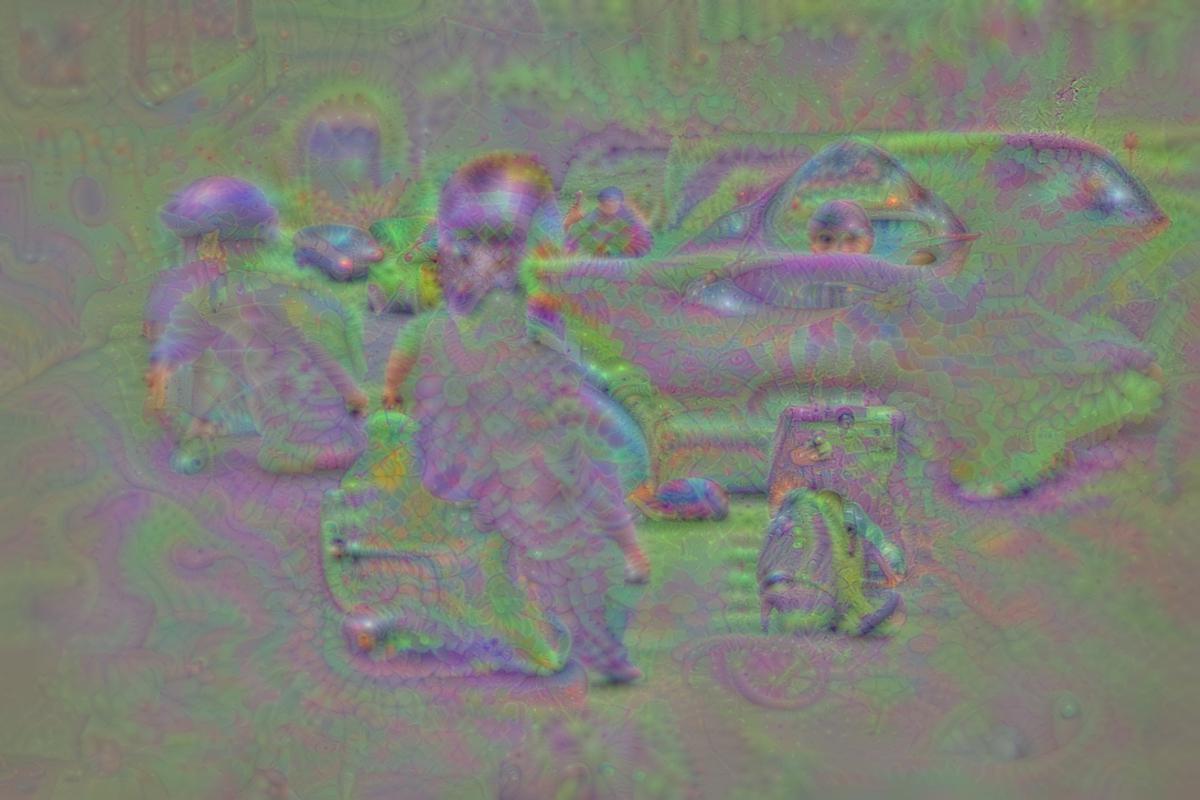}&
    \includegraphics[width=0.138\textwidth]{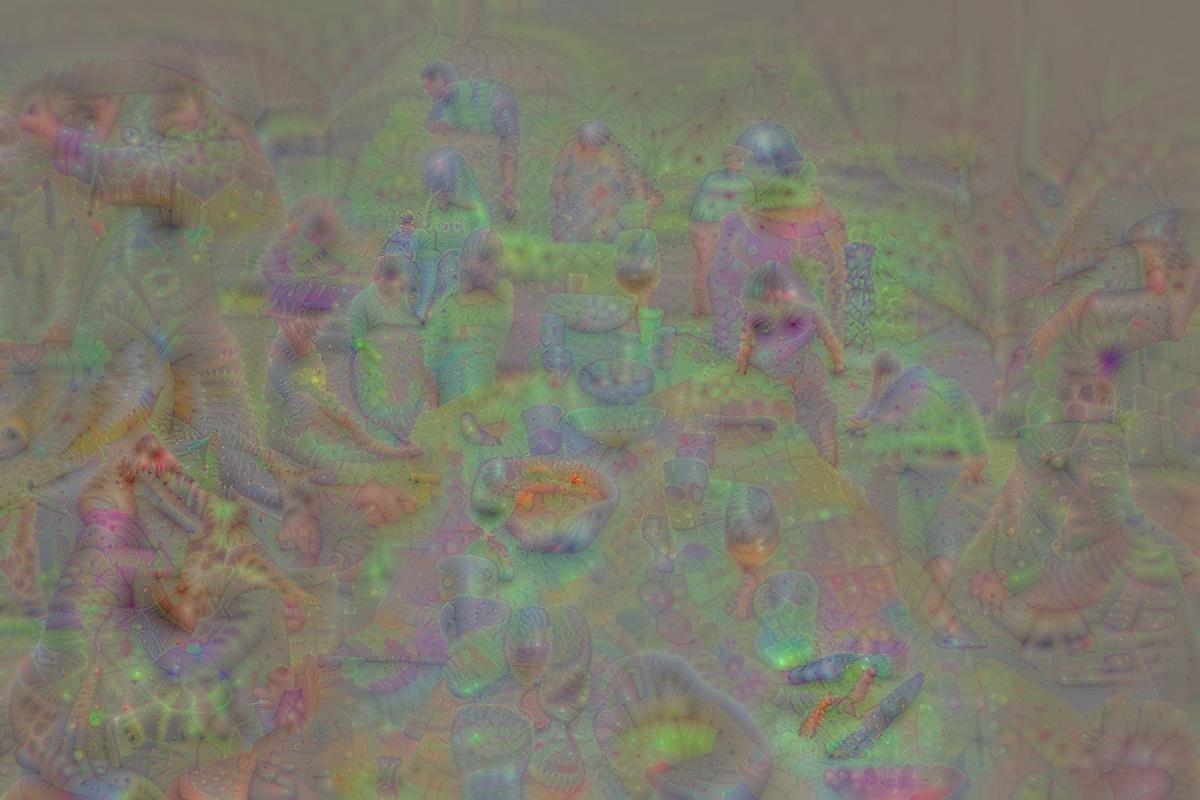}&
    \includegraphics[width=0.138\textwidth]{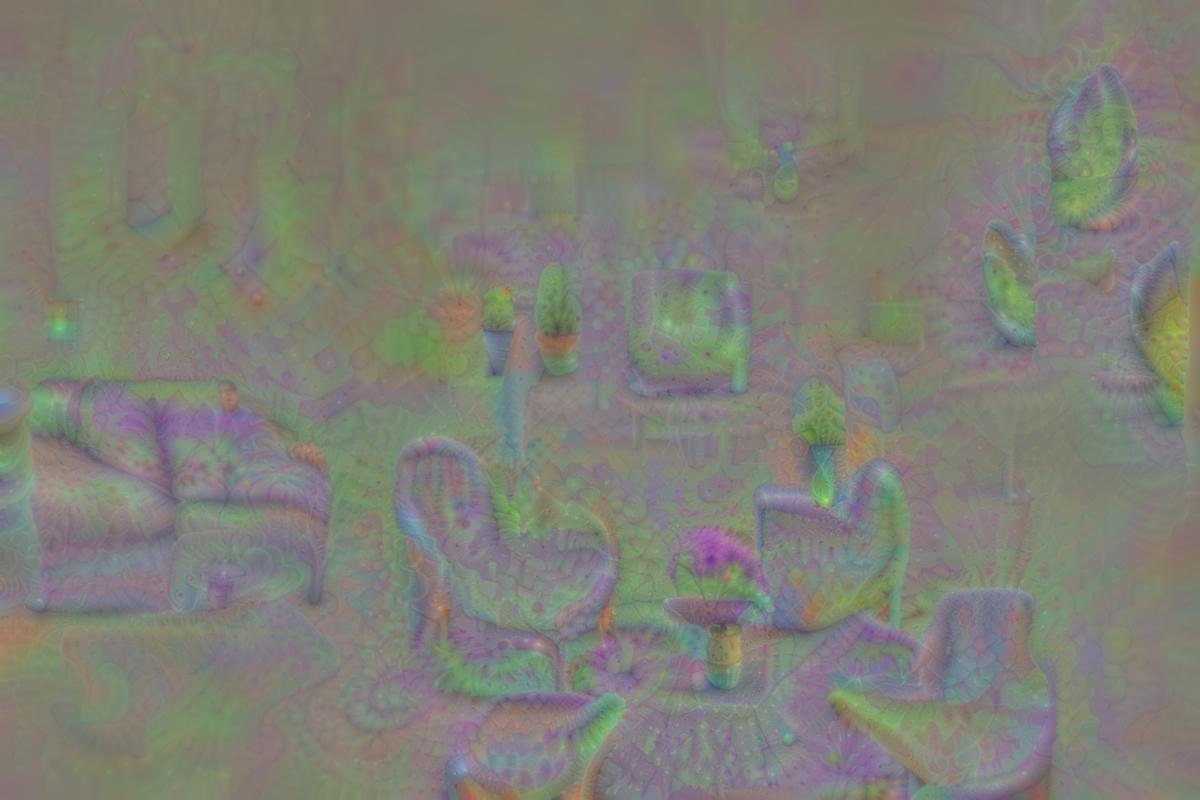}&
    \includegraphics[width=0.138\textwidth]{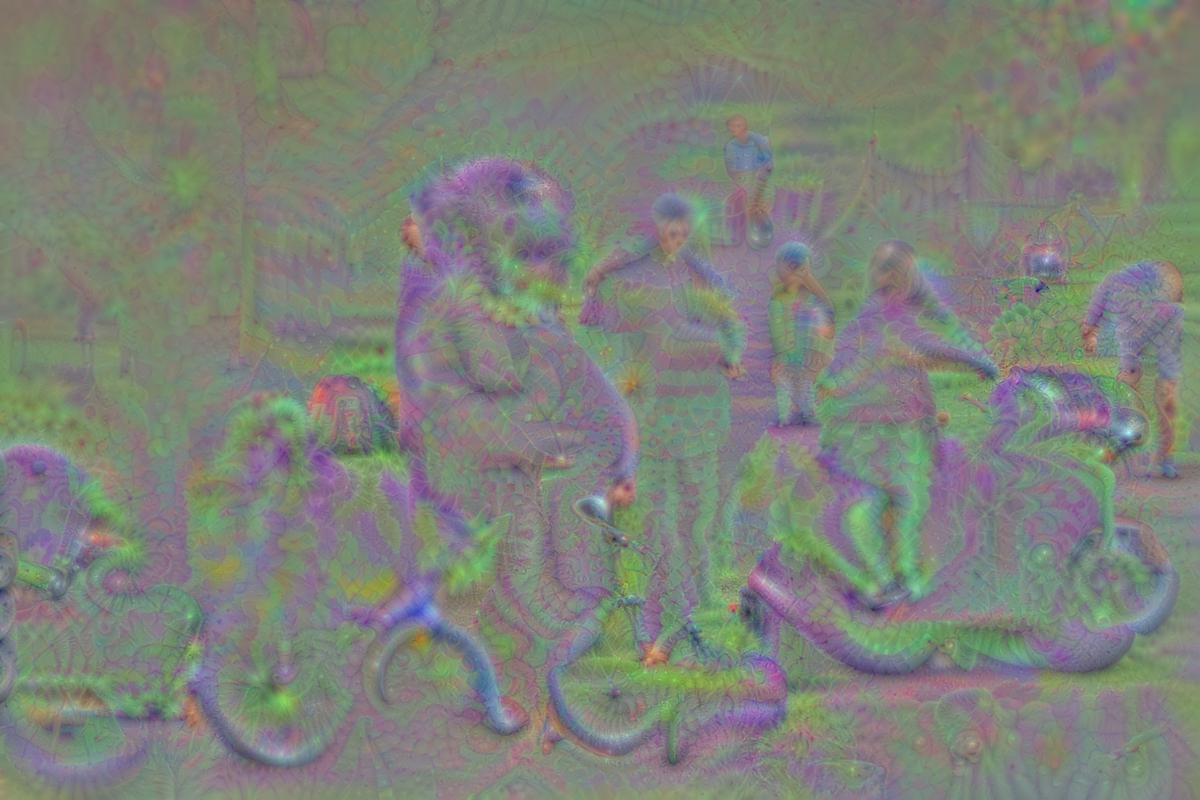}&
    \includegraphics[width=0.138\textwidth]{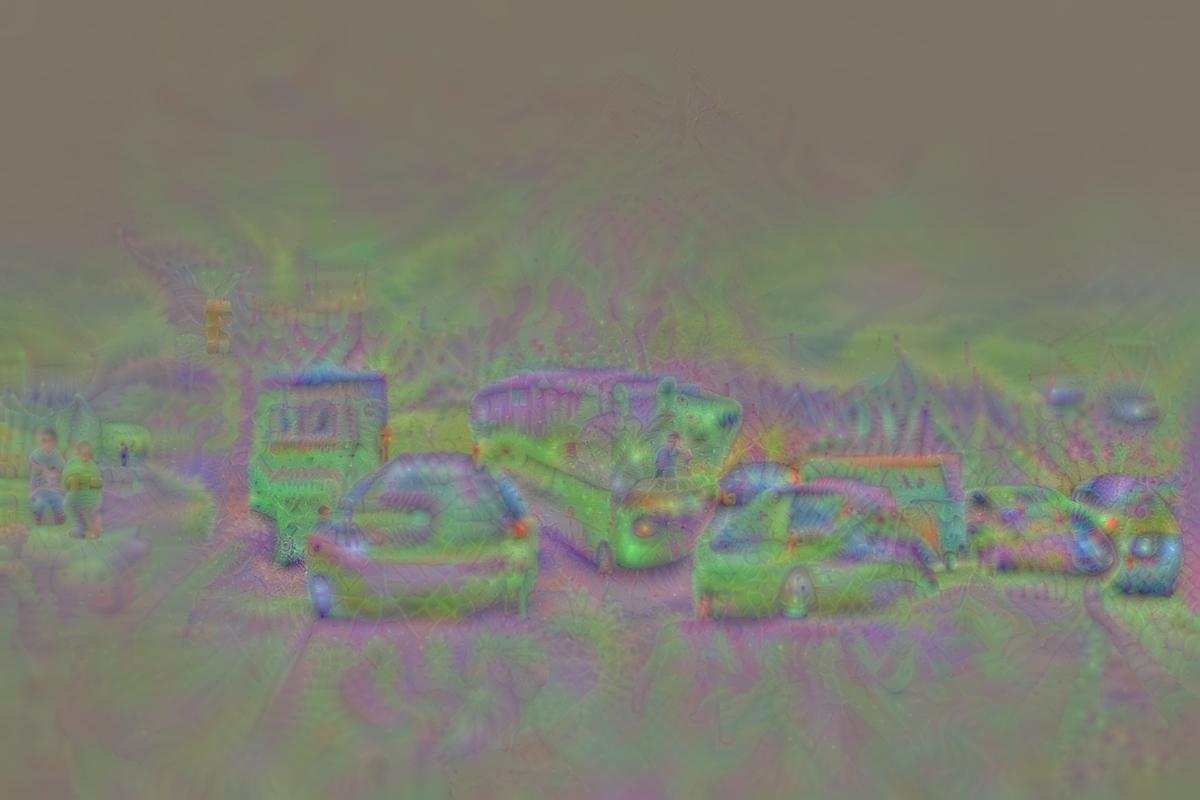}\\*[-0.5mm]
    \rotatebox{90}{\small  \hspace{2.5mm} FCOS}
    \includegraphics[width=0.138\textwidth]{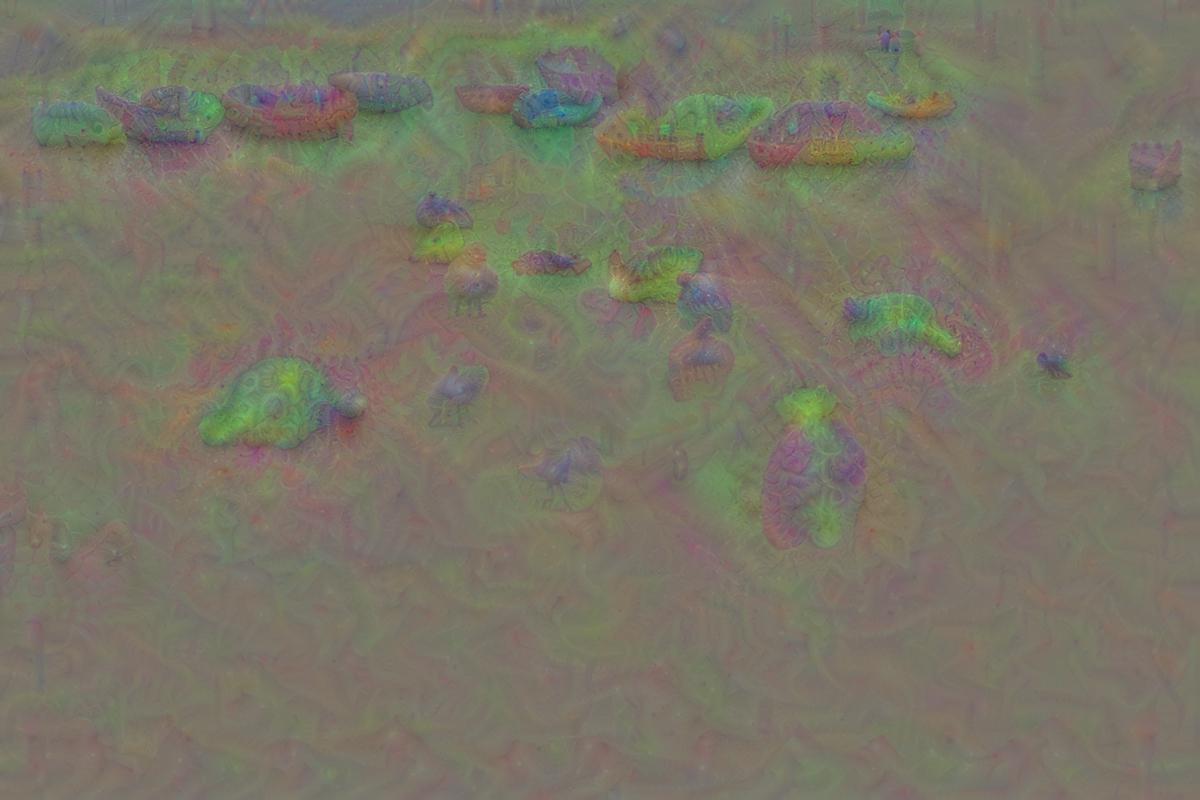}&
    \includegraphics[width=0.138\textwidth]{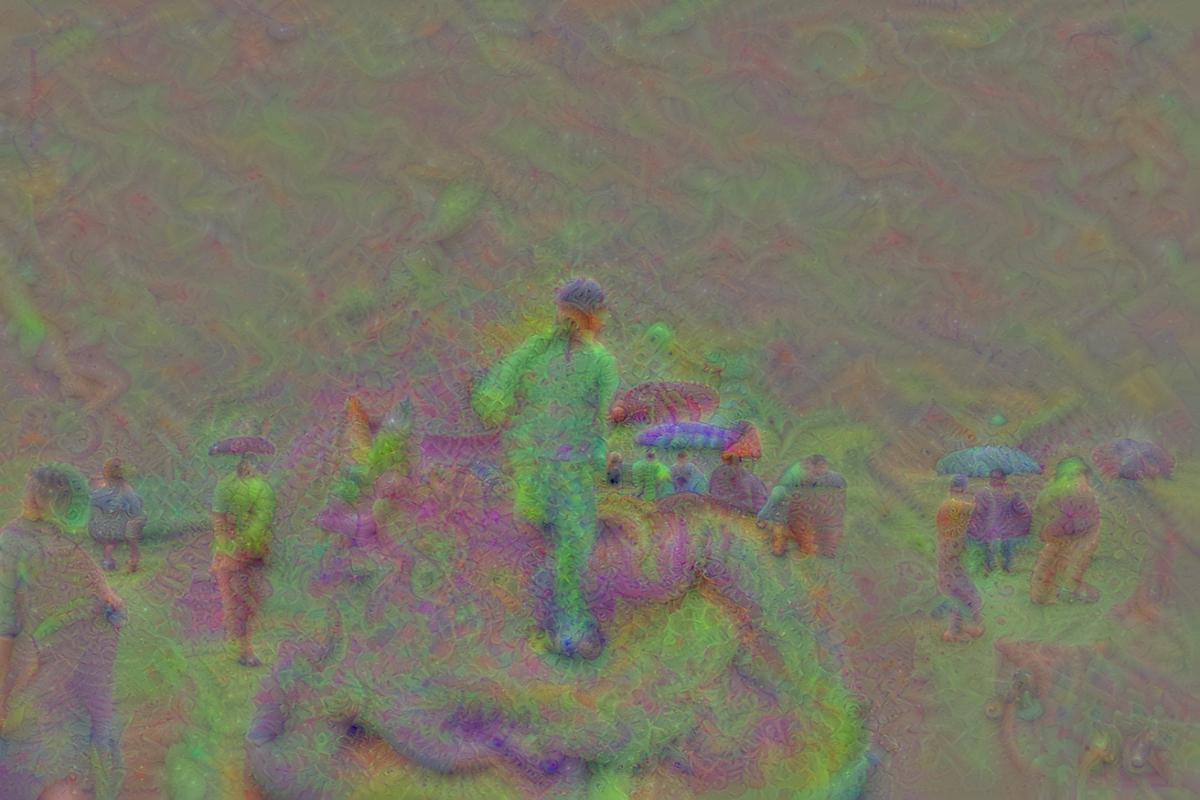}&
    \includegraphics[width=0.138\textwidth]{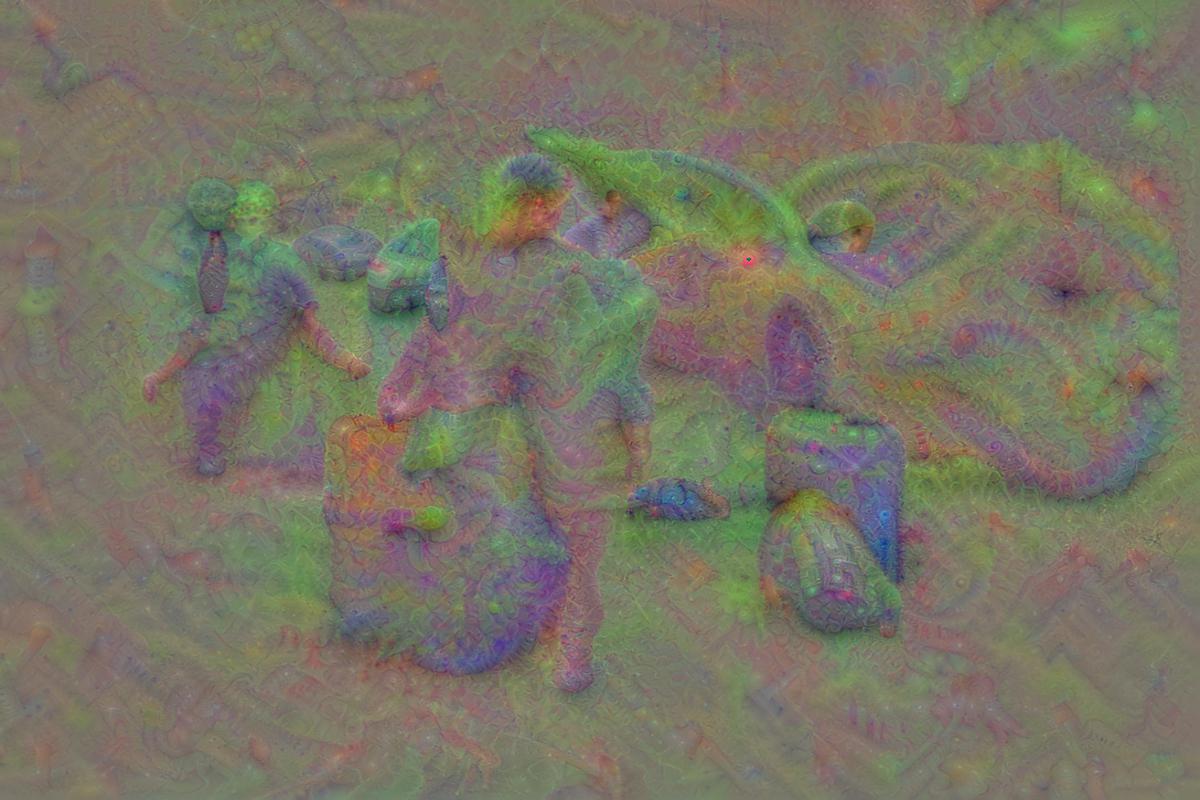}&
    \includegraphics[width=0.138\textwidth]{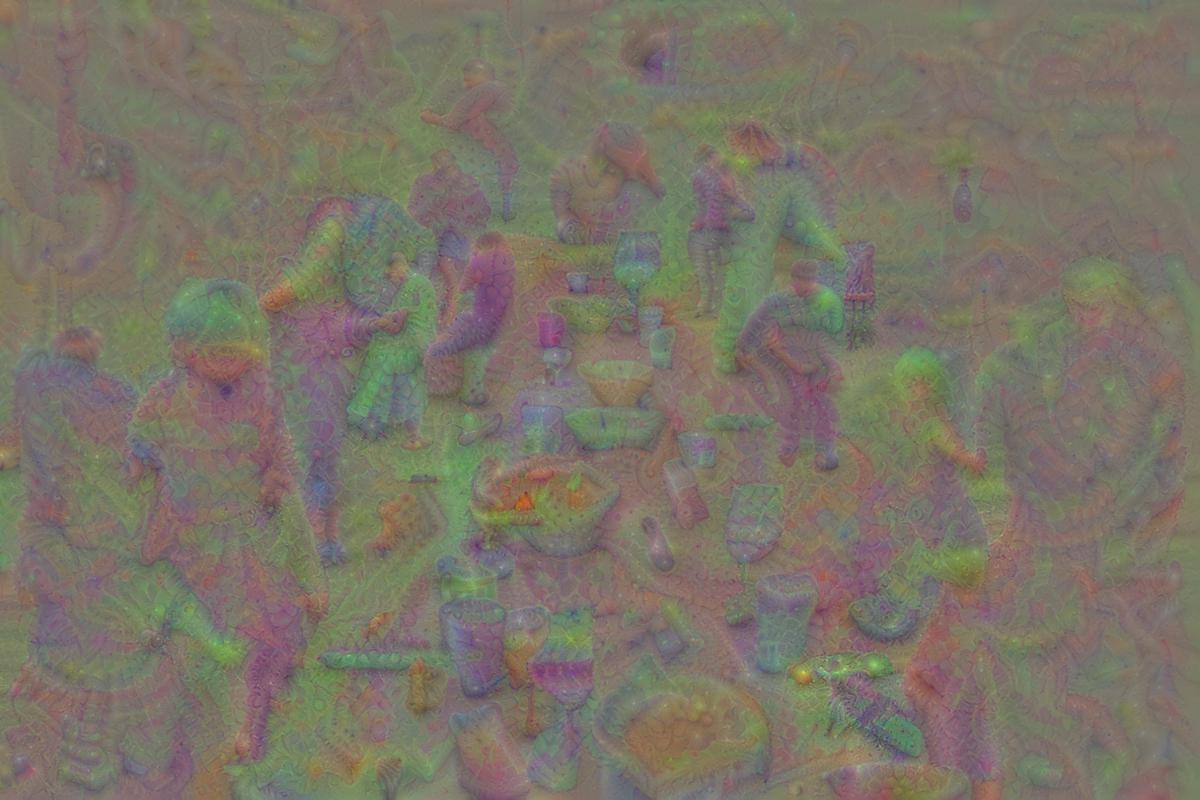}&
    \includegraphics[width=0.138\textwidth]{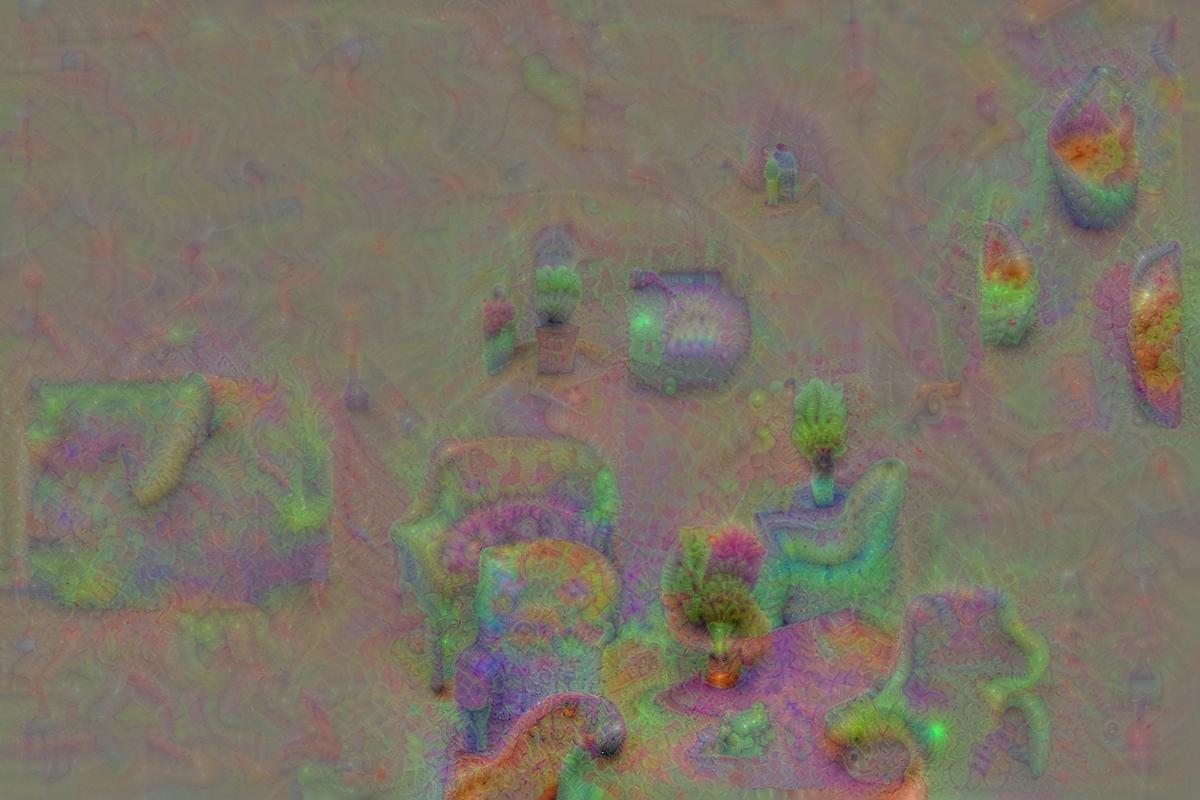}&
    \includegraphics[width=0.138\textwidth]{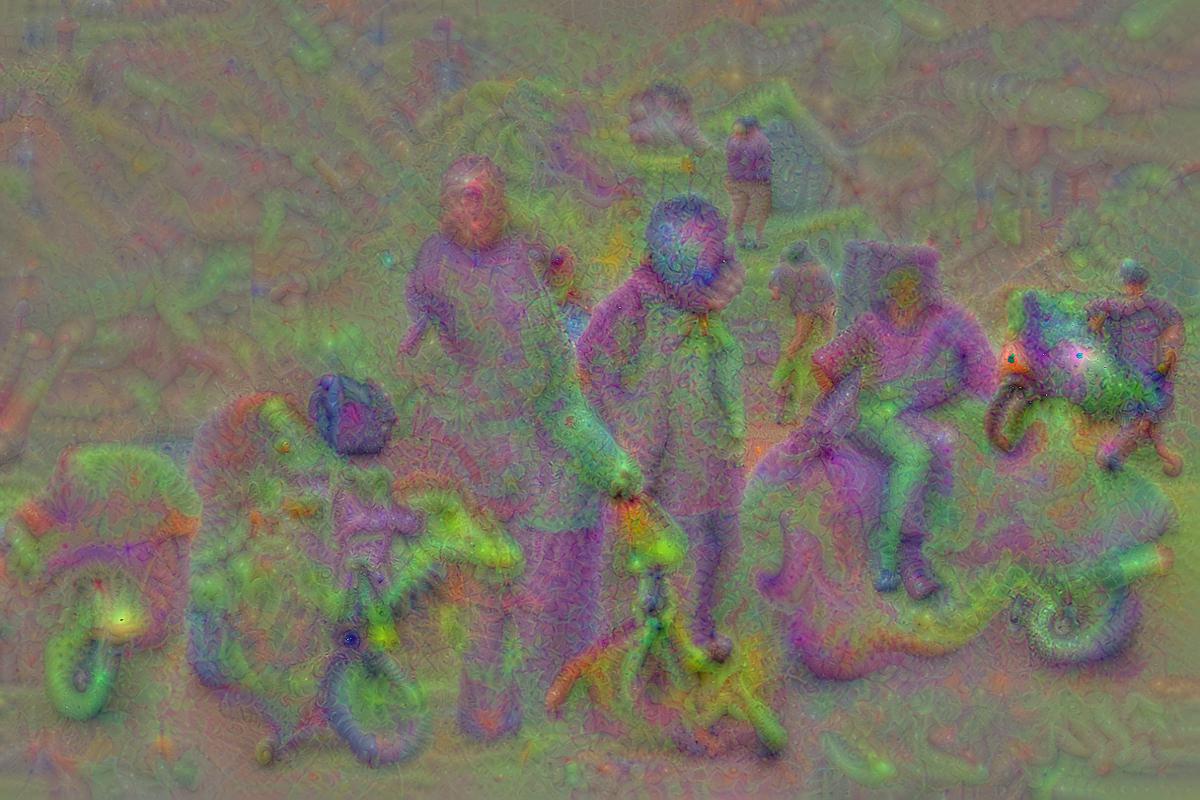}&
    \includegraphics[width=0.138\textwidth]{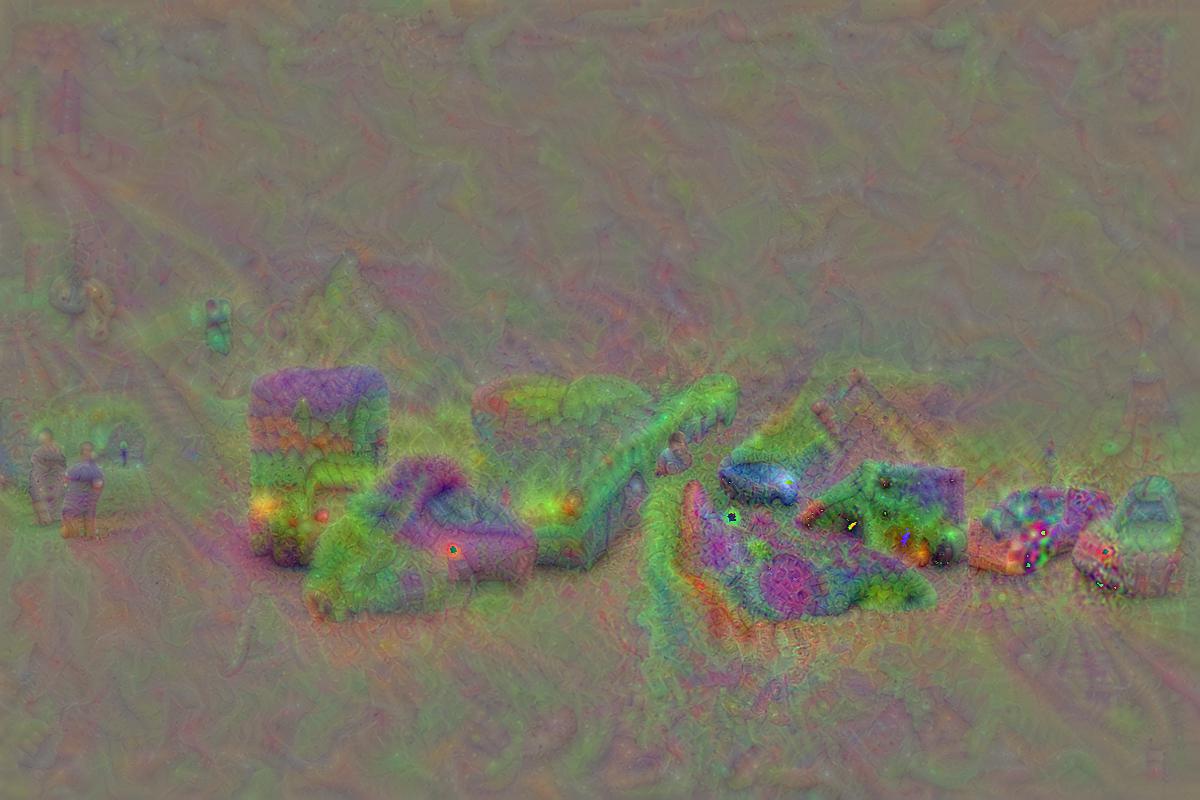}\\*[-0.5mm]
  %  \rotatebox{90}{\small  \hspace{2.5mm} Corner}
  %   \includegraphics[width=0.138\textwidth]{./image/figure1_visual_results/jpg/000000031322_Corner.jpg}&
  %   \includegraphics[width=0.138\textwidth]{./image/figure1_visual_results/jpg/000000439715_Corner.jpg}&
  %   \includegraphics[width=0.138\textwidth]{./image/figure1_visual_results/jpg/000000009891_Corner.jpg}&
  %   \includegraphics[width=0.138\textwidth]{./image/figure1_visual_results/jpg/000000018380_Corner.jpg}&
  %   \includegraphics[width=0.138\textwidth]{./image/figure1_visual_results/jpg/000000013923_Corner.jpg}&
  %   \includegraphics[width=0.138\textwidth]{./image/figure1_visual_results/jpg/000000038829_Corner.jpg}&
  %   \includegraphics[width=0.138\textwidth]{./image/figure1_visual_results/jpg/000000026204_Corner.jpg}
    \end{tabular}
    \vspace{-2mm}
    \caption{
      Visualization results with layout information input for Mask R-CNN, Faster R-CNN, RetinaNet, and FCOS,
      all using ResNet-50-FPN backbones.
      The layout information is the bounding box, category annotations, with additional instance masks only for Mask R-CNN.
      All generated images reproduce the desired layouts when passed to detectors.
      Zoom in for details.
    }
    \vspace{-6mm}
    \label{fig:visual_results}
    \end{figure*}

\textbf{Network We Visualize.}
We focus on a set of models spanning different meta-architectures~\cite{huang2017speed}, which we believe are a representative subset of modern detectors.
We use \textbf{RetinaNet}~\cite{lin2017focal} as a single-stage method.
It uses a fully-convolutional network to jointly classify and regress box predictions from a set of convolutional anchors.
We use \textbf{Faster R-CNN} ~\cite{ren2015faster} as a two-stage method.
It uses a fully-convolutional region proposal network (RPN) to predict regions of interest (RoIs), which are classified and refined with a second per-region network.
We use \textbf{Mask R-CNN}~\cite{he2017mask} as a instance segmentation method.
It augments Faster R-CNN with a fully-convolutional segmentation network applied per-RoI to predict per-object binary masks.
We use \textbf{FCOS}~\cite{tian2019fcos} as an anchor-free method. 
It uses a fully-convolutional network to classify and regress box predictions on a set of points on the feature map.
% We use \textbf{CornerNet}~\cite{law2018cornernet} as an anchor-free method.
% It predicts heatmap for top-left and right-bottom points and gets the box by grouping points with similar features.
We use reference models trained on COCO~\cite{lin2014microsoft}, provided by Detectron2~\cite{wu2019detectron2}.
All models use ResNet-50-FPN backbones~\cite{lin2017feature}.

\textbf{Optimization Details.}
We conduct 1000 alternating iterations, and in each iteration, we update $I$ using one-step SGD.  
We use periodically blurring~\cite{yosinski2015understanding}, TV-norm~\cite{mahendran2015understanding} and  $p$-norm~\cite{simonyan2013deep} as regularizer. 
See details in supplementary.

\subsection{Inverting Scene Layouts}
\label{sec:invert}

We invert scene layouts of images from the COCO validation set; results are shown in Figure~\ref{fig:visual_results}.
In all cases, the generated images reproduce the target layout when passed to the detector.
The inversions of different object detectors share some intriguing features.
Inversions show clear object structures and category-specific details:
shoes, pants, shirts, and faces for people~(col.~2); 
wheels, windows, car lights and traffic lights~(col.~7);
canonical shapes of furniture~(col.~5).
These features imply that detectors have learned these canonical traits as visual cues for detections. 

\begin{table}[ht!]
    \newcommand{\clsreg}{{\scriptsize (box)}}
    \newcommand{\boxmask}{{\scriptsize (box+mask)}}
    \newcommand{\hl}{\cellcolor[RGB]{255,242,204}}
    \newcommand{\M}{\cellcolor[RGB]{200,242,204}}
    \centering
    \resizebox{0.48\textwidth}{!}{%
    \begin{tabular}{ p{1.5cm} p{1.2cm} p{1.2cm}  p{1.2cm}  p{1.2cm} p{1.2cm}}
        \toprule [2 pt]        
        \small{invert\textbackslash test} &Retina &Faster &Mask$^{box}$ &Mask$^{mask}$ &FCOS\\
        \hline 
        Retina & \hl{91.6} &59.4 &      55.5& 17.5 & 61.0 \\
        Faster & 74.6 &   \hl{94.8} &      86.1& 25.7 &  75.1 \\
        Mask-b & 69.4 &   86.3 &      \hl{94.3}& 26.8 & 69.3\\
        Mask-m & 63.1 &   82.0 &      93.6& \hl{85.1} & 63.0 \\
        FCOS   & 56.2 &   54.6 &      52.7& 16.6 & \hl{95.3} \\
        % Corner & 6.1 &   13.34 &      11.8& 3.7  & 4.5\\
    \bottomrule [2 pt]
    \end{tabular}%
    }    
    \vspace{-3mm}
    \caption{
    \textbf{Inversion Transfer.}
    We invert 1k layouts from COCO'17 val using different object detectors;
    for Mask R-CNN we invert only boxes~(Mask-b) and boxes+masks~(Mask-m) (similar to Figure~\ref{fig:losses}).
    We run all generated images through each detector and compute AP to see whether detectors recognize objects similarly.
    Every row represents the detection results by different detected models with the images inverted from the same network.
    Mask$^{mask}$ is AP of instance segmentation using Mask R-CNN.
    }
    \vspace{-6mm}
    \label{tab:transfer}
\end{table}

Moreover, the object structure seems context-dependent, and objects of the same category have different modalities given different contexts.
People have different forms and actions under umbrellas~(col.~2),  on horses~(col.~2) or motorcycles~(col.~6), next to a table~(col.~4).

% We further analyze visualizations by comparing results from different detectors. 
We compare inversions from different detectors to analyze visual cues learned by differing models. 
Inverting layouts from Mask R-CNN more faithfully reproduce the original image since its layouts include mask information. 
However, inverting layouts with Faster R-CNN also results in clear object boundaries and internal parts, such as the horse's ear, eye, leg, and hoof~(col.~2), bicycle wheels and body~(col.~6).
It again implies that detectors have a good understanding of shapes and internal structures of objects. 

Inverting layouts with RetinaNet and FCOS gives qualitatively different results from two-stage methods.
Smaller objects show clear boundaries and structures, but larger objects may instead appear as an amalgamation of parts.
We hypothesize that single-stage methods, especially RetinaNet, struggle to model the global structures of large objects and resort to more local features for detections,
while RoI Align may help two-stage methods handle large objects, resulting in explicit boundaries and structures. 
Besides superior detection performance, inverting FCOS also shows finer results than RetinaNet, implying better features.
\begin{figure*}[ht!]
\centering
\begin{tabular}{@{}c@{\hspace{0.5mm}}c@{\hspace{0.5mm}}c@{\hspace{0.5mm}}c@{\hspace{0.5mm}}c@{\hspace{0.5mm}}c@{\hspace{0.7mm}}c@{}}
\small \text{Layout} &
\small \text{Cls+Reg+Mask} & 
\small \text{Mask} &
\small \text{Cls} &
\small \text{Reg} & 
\small \text{Cls+Reg} &
\small \text{Faster R-CNN}\\
\includegraphics[width=0.138\textwidth]{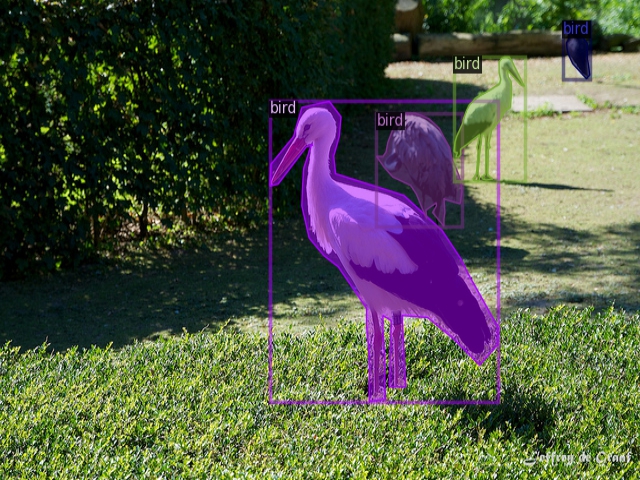}&
\includegraphics[width=0.138\textwidth]{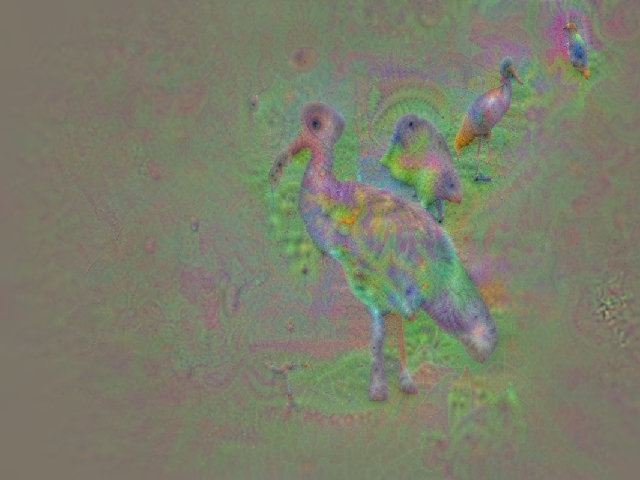}&
\includegraphics[width=0.138\textwidth]{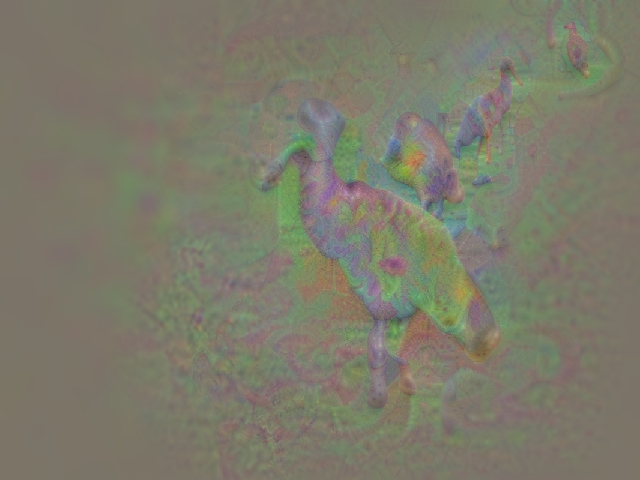}&
\includegraphics[width=0.138\textwidth]{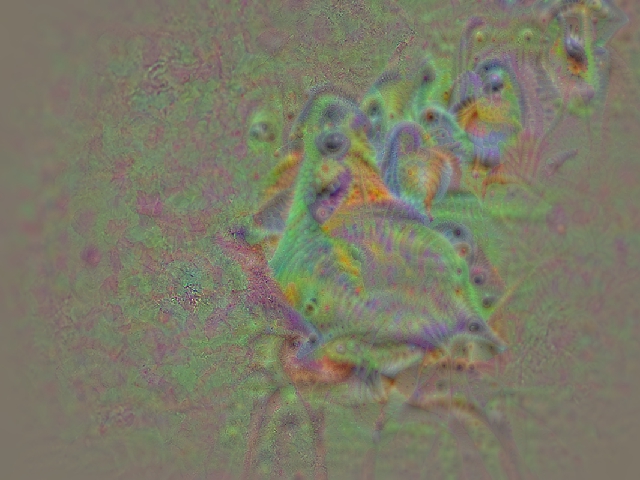}&
\includegraphics[width=0.138\textwidth]{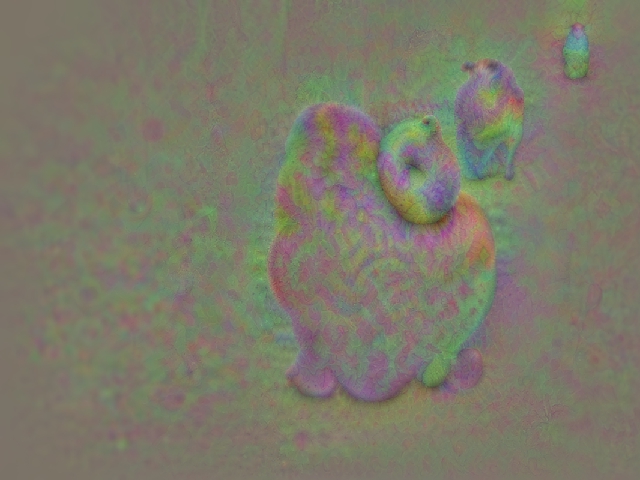}&
\includegraphics[width=0.138\textwidth]{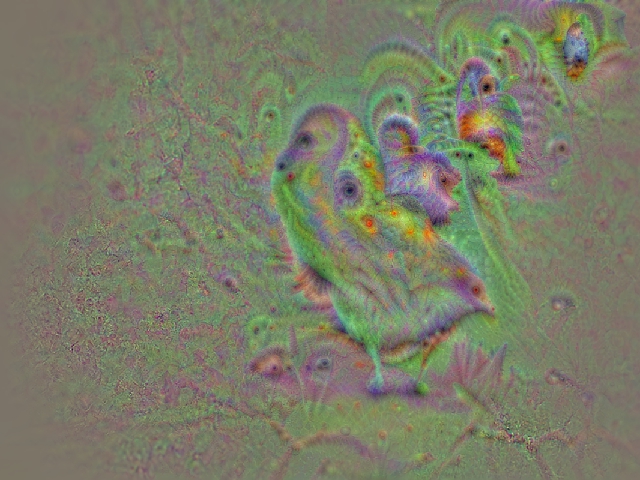}&
\includegraphics[width=0.138\textwidth]{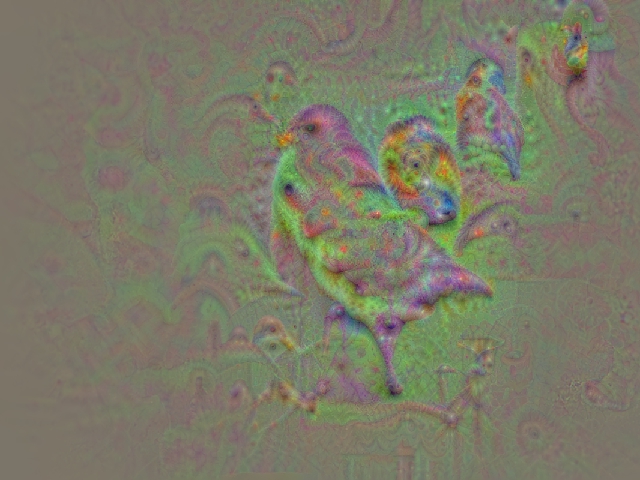}\\*[-0.5mm]
\includegraphics[width=0.138\textwidth]{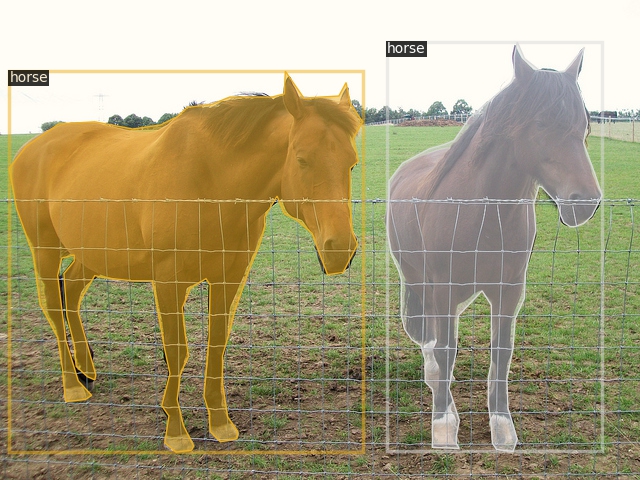}&
\includegraphics[width=0.138\textwidth]{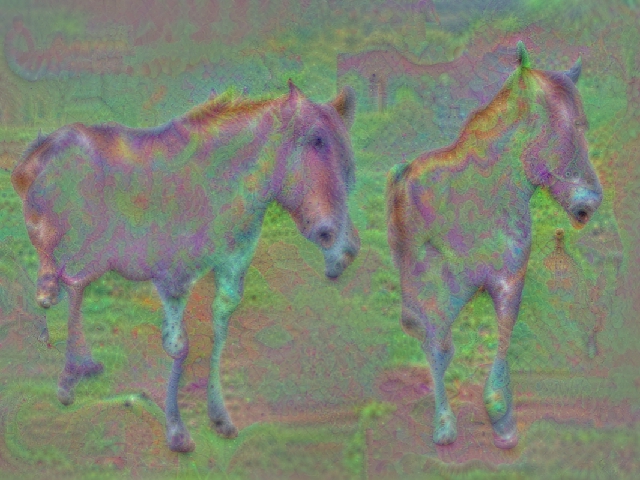}&
\includegraphics[width=0.138\textwidth]{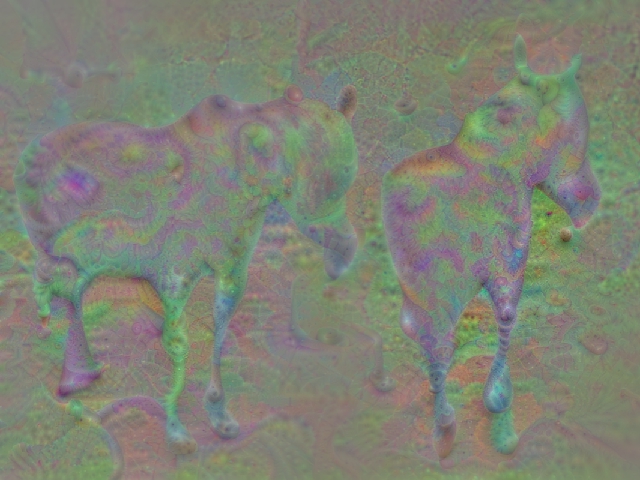}&
\includegraphics[width=0.138\textwidth]{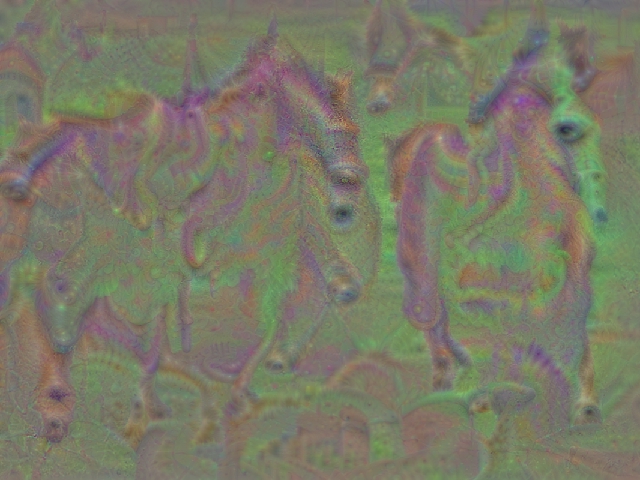}&
\includegraphics[width=0.138\textwidth]{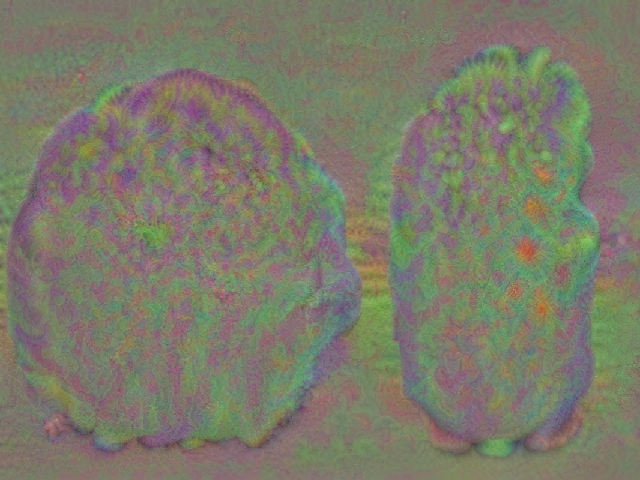}&
\includegraphics[width=0.138\textwidth]{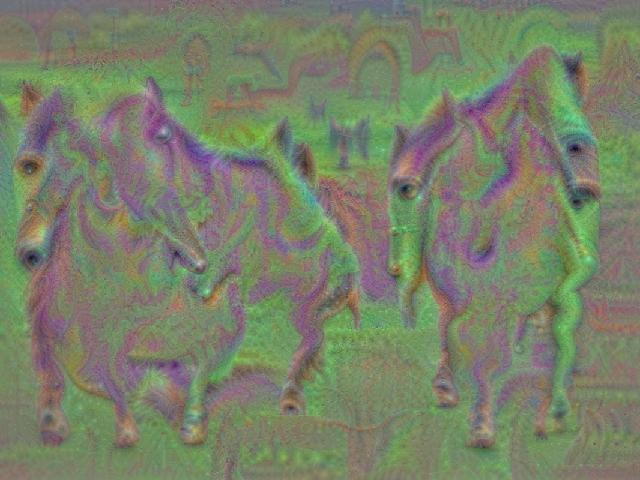}&
\includegraphics[width=0.138\textwidth]{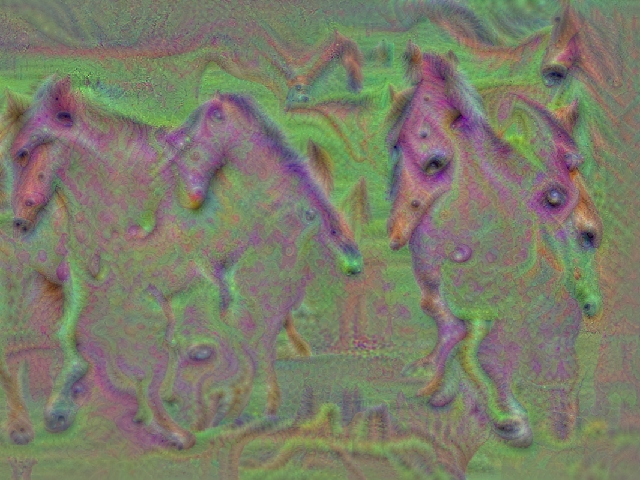}\\*[-0.5mm]
\end{tabular}
\vspace{-4mm}
\caption{
  Mask R-CNN produces three outputs per RoI: object classification (cls); regression from RoI to output box (reg); and segmentation mask.
  We perform layout inversion on Mask R-CNN where different subsets of the per-RoI outputs must match the layout while don't optimize the rest.
  We also show results using Faster R-CNN, which matches the (cls) and (reg) outputs only for comparison with cls+reg. 
}
\label{fig:losses}
\vspace{-5mm}
\end{figure*}

\begin{figure}[ht!]
    \centering
    \resizebox{0.48\textwidth}{!}{%
    \begin{tabular}{@{}c@{\hspace{0.5 mm}}c@{\hspace{0.5mm}}c@{\hspace{0.5mm}}c@{}}
    Sheep &  Horse & Banana & Elephant  \\
    % \rotatebox{90}{\tiny \hspace{4mm} \text{Image}}
    \rotatebox{90}{\scriptsize \hspace{5mm} \text{Image}}
    \includegraphics[width=0.14\textwidth]{./image/figure8_attributions_cls/jpg/000000046804.jpg}&
    \includegraphics[width=0.14\textwidth]{./image/figure8_attributions_cls/jpg/000000118209.jpg}&
    \includegraphics[width=0.14\textwidth]{./image/figure8_attributions_cls/jpg/000000560256.jpg}&
    \includegraphics[width=0.14\textwidth]{./image/figure8_attributions_cls/jpg/000000021903.jpg}
    \\*[-0.5mm]
    
    \rotatebox{90}{\scriptsize \hspace{0.5mm} Grad-CAM~\cite{selvaraju2017grad}}
    % \rotatebox{90}{\small \text{Grad-Cam}}
    \includegraphics[width=0.14\textwidth]{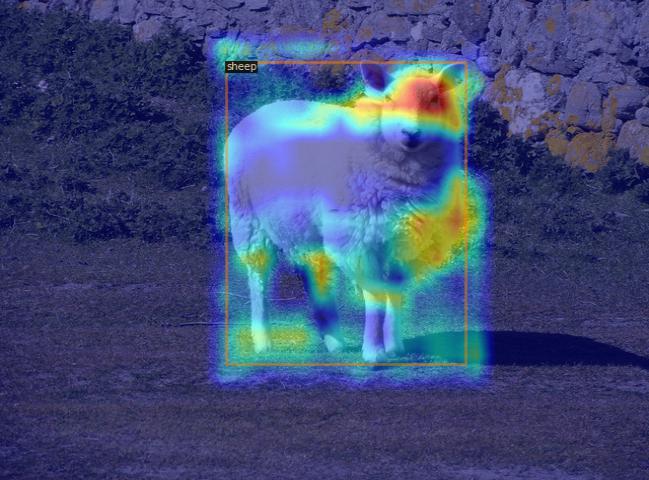}&
    \includegraphics[width=0.14\textwidth]{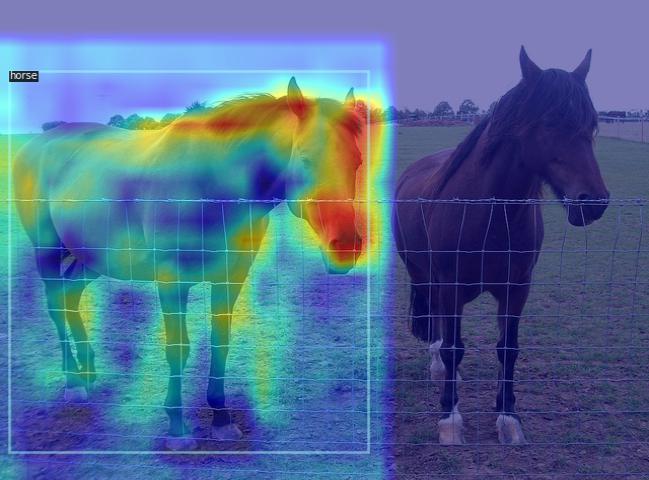}&
    \includegraphics[width=0.14\textwidth]{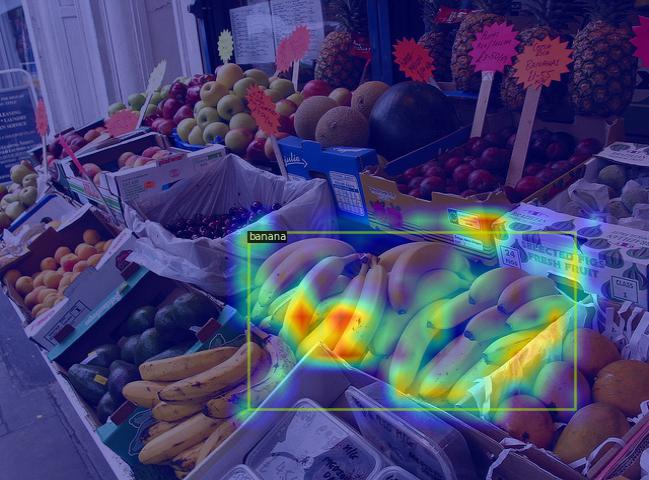}&
    \includegraphics[width=0.14\textwidth]{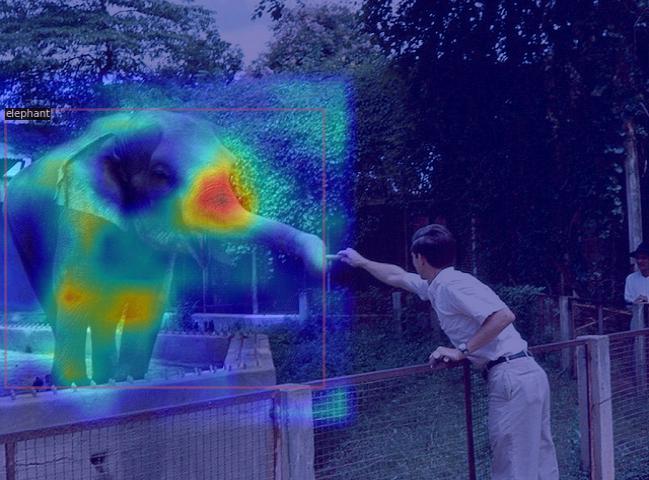}
    \\*[-0.5mm]
    % \rotatebox{90}{\tiny \hspace{1.5mm} Gradient~\cite{zeiler2014visualizing}}
    % % \rotatebox{90}{\small  \hspace{1mm} \text{Gradient}}
    % \includegraphics[width=0.14\textwidth]{./image/figure8_attributions_cls/jpg/000000046804_Instance_0_Gradient.jpg}&
    % \includegraphics[width=0.14\textwidth]{./image/figure8_attributions_cls/jpg/000000118209_Instance_0_Gradient.jpg}&
    % \includegraphics[width=0.14\textwidth]{./image/figure8_attributions_cls/jpg/000000560256_Instance_1_Gradient.jpg}&
    % \includegraphics[width=0.14\textwidth]{./image/figure8_attributions_cls/jpg/000000021903_Instance_1_Gradient.jpg}
    % \\*[-0.5mm]
    % \rotatebox{90}{\tiny \hspace{3mm} Mask~\cite{fong2019understanding}}
    \rotatebox{90}{\scriptsize \hspace{3mm} \text{Mask}~\cite{fong2019understanding}}
    \includegraphics[width=0.14\textwidth]{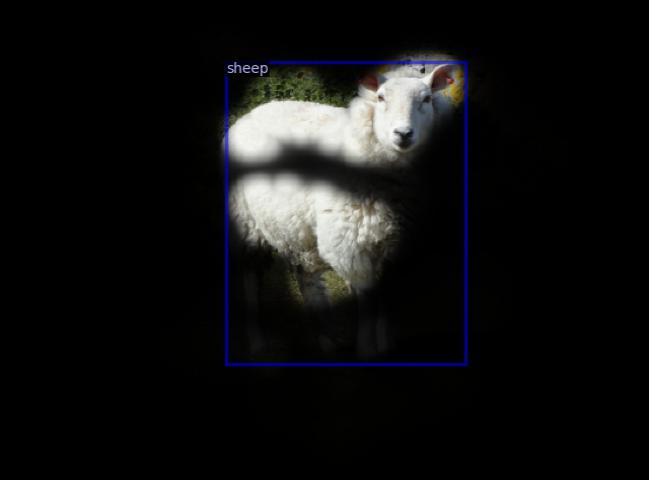}&
    \includegraphics[width=0.14\textwidth]{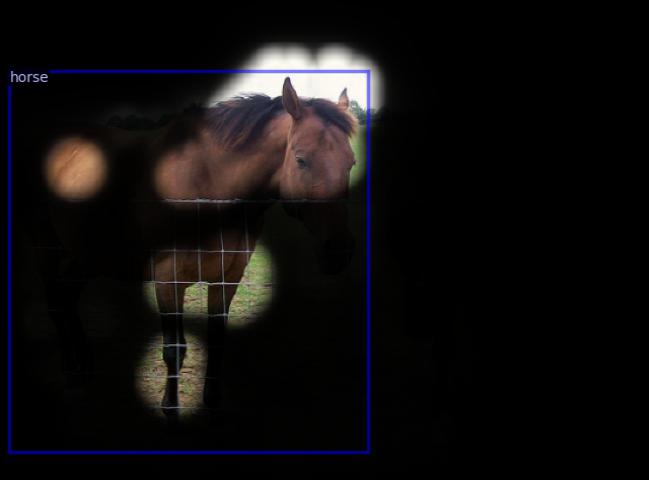}&
    \includegraphics[width=0.14\textwidth]{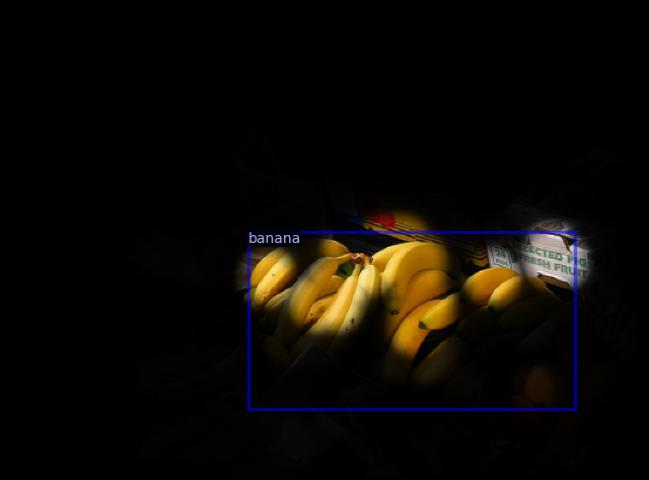}&
    \includegraphics[width=0.14\textwidth]{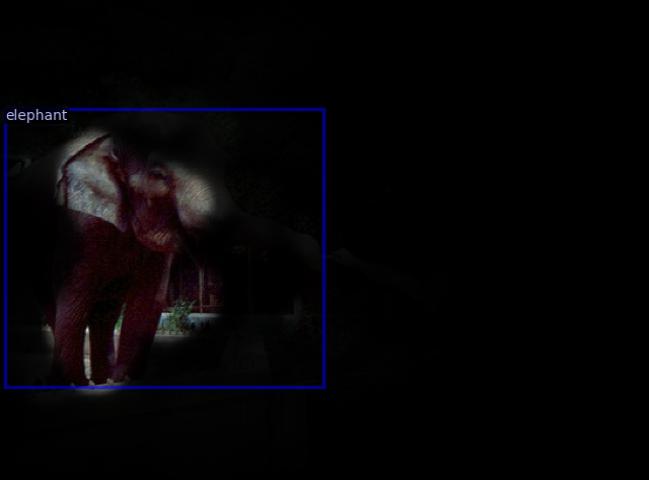}
  \\*[-0.5mm]
    \rotatebox{90}{\scriptsize \hspace{0.25mm} NormGrad~\cite{rebuffi2020there}}
    \includegraphics[width=0.14\textwidth]{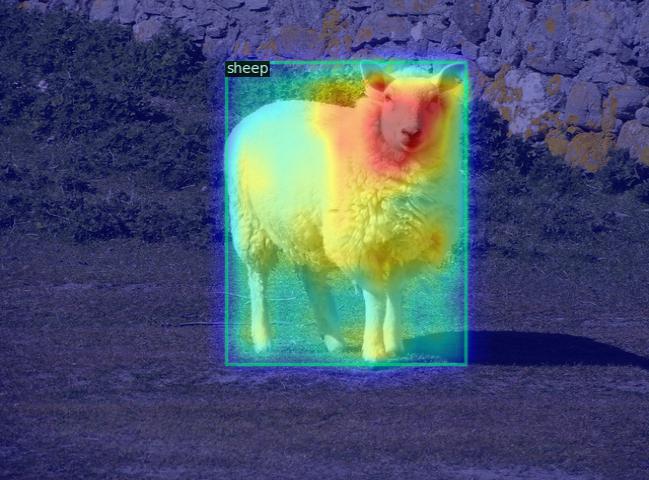}&
    \includegraphics[width=0.14\textwidth]{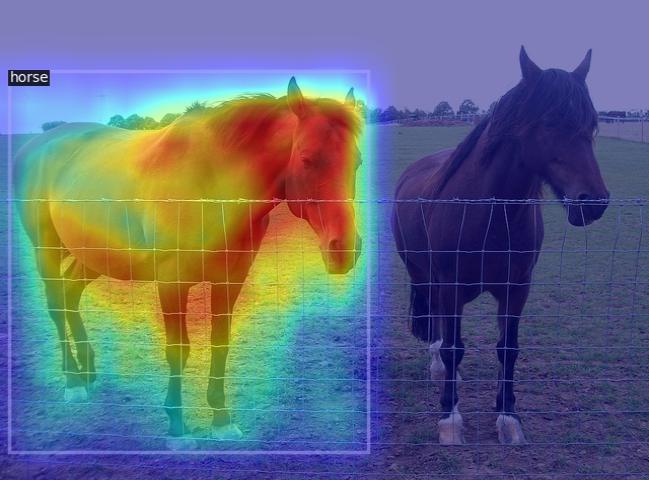}&
    \includegraphics[width=0.14\textwidth]{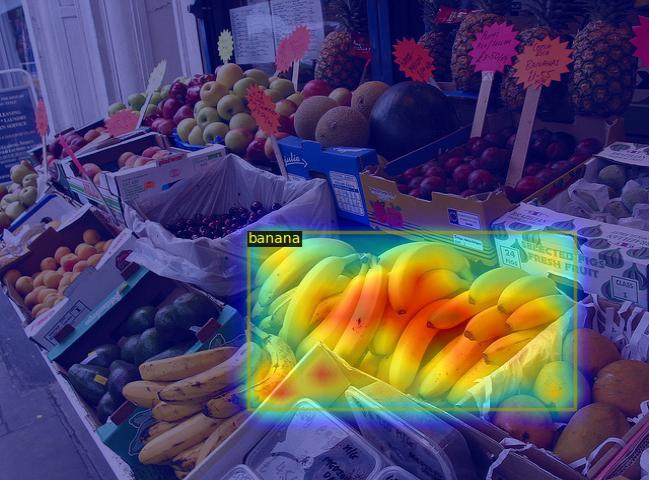}&
    \includegraphics[width=0.14\textwidth]{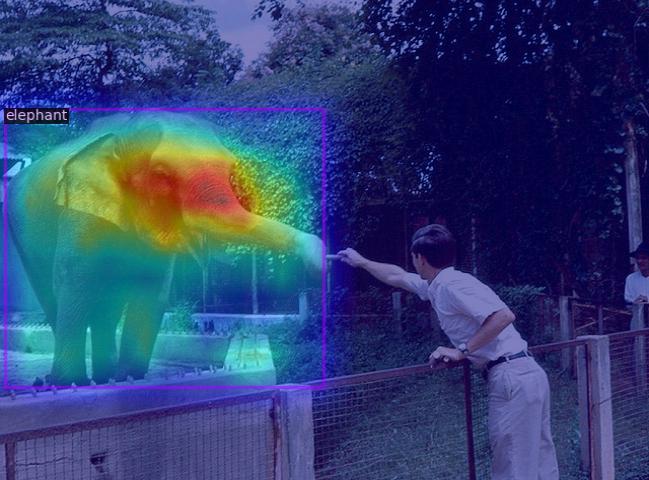}   
    \\*[-0.5mm]
    \end{tabular}}
    \vspace{-4mm}
    \caption{
      We use attribution methods to find image regions responsible for classification in Mask R-CNN's second stage.
      % We smooth Gradient~\cite{zeiler2014visualizing} results and
      We truncate Grad-CAM~\cite{selvaraju2017grad} results to the RoI;
      for Mask~\cite{fong2019understanding} we preserve area equal to half the ground-truth mask.
      See Supplementary for more examples on Mask R-CNN and other object detectors.
    %  Comparison between attribution methods on Mask R-CNN model for classification. Results of Gradient have been smoothed; we preserve the region of 0.5 target mask size in perturbation methods for better discrimination.
    }
    \vspace{-5mm}
    \label{fig:attribution_cls}
    \end{figure}

\subsection{Inversion Transfer.}
\label{sec:inver_trans}

We quantify our inversion and further measure the similarity between different object detectors via \emph{inversion transfer}.
We invert a layout using one detector, then pass the generated image to another detector and check whether it also reproduces the layout.
If two detectors similarly recognize objects or learn similar visual cues, we should expect high inversion transfer.
Table~\ref{tab:transfer} evaluates inversion transfer using 1k layouts from COCO'17 val, following the evaluation protocol of COCO.

Inverting and evaluating with the same model gives high but imperfect APs; inversions may not be perfect for small objects, crowded scenes, and bad annotated instances. 
In particular, imperfect APs come from box positions: achieving 100 AP under the 0.95 test IoU threshold requires the inverted bounding boxes to be exactly the same as annotated boxes for small objects at pixel level. 
Besides, image regularizers used in visualization~(such as blur, jitter) for better recognition may also affect the AP number.

Inversions are highly transferable between Faster and Mask R-CNN, showing that they use similar cues for recognition.
Inversions are less transferable between single~(RetinaNet, FCOS) and two-stage~(Faster and Mask R-CNN) detectors, partially supported by the qualitative difference of visualizations in Figure~\ref{fig:visual_results}.
Inversions are also less transferable between detection and instance segmentation models:
FCOS and RetinaNet have higher APs on  inversions from Faster R-CNN than from Mask R-CNN, and their inversions are  better detected by Faster R-CNN as well,
implying distinct features learned for instance segmentation task.
\begin{figure}[ht!]
    \centering
    \resizebox{0.48\textwidth}{!}{%
    \begin{tabular}{@{}c@{\hspace{0.5 mm}}c@{\hspace{0.5mm}}c@{\hspace{0.5mm}}c@{}}
    dx &dy &dx &dy  \\
    \rotatebox{90}{\scriptsize \hspace{5mm} \text{Image}}
    \includegraphics[width=0.14\textwidth]{./image/figure8_attributions_cls/jpg/000000018737.jpg}&
    \includegraphics[width=0.14\textwidth]{./image/figure8_attributions_cls/jpg/000000018737.jpg}&
    \includegraphics[width=0.14\textwidth]{./image/figure8_attributions_cls/jpg/000000040036.jpg}&
    \includegraphics[width=0.14\textwidth]{./image/figure8_attributions_cls/jpg/000000040036.jpg}\\*[-0.5mm]
    \rotatebox{90}{\scriptsize \hspace{0.5mm} Grad-CAM~\cite{selvaraju2017grad}}
    % \rotatebox{90}{\small \text{Grad-Cam}}
    \includegraphics[width=0.14\textwidth]{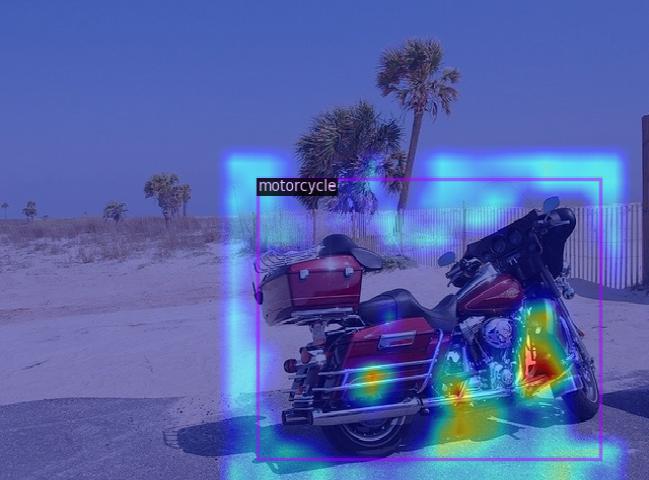}&
    \includegraphics[width=0.14\textwidth]{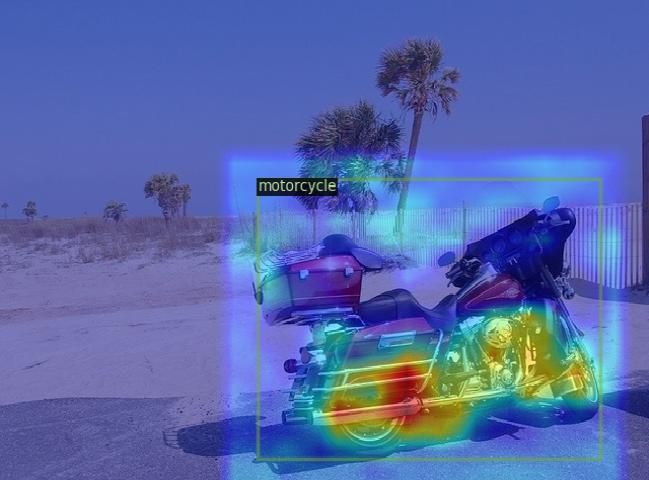}&
    \includegraphics[width=0.14\textwidth]{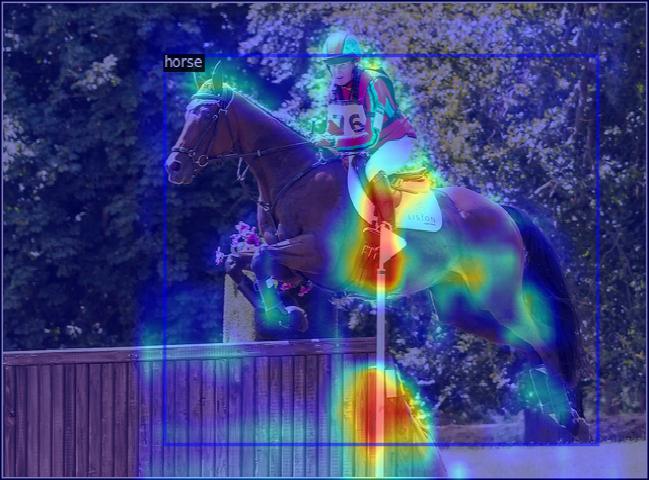}&
    \includegraphics[width=0.14\textwidth]{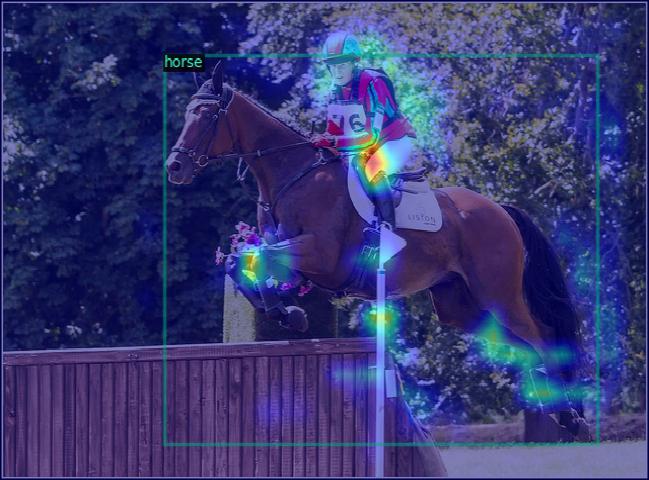}
    
    \\*[-0.5mm]
    % \rotatebox{90}{\tiny \hspace{1.5mm} Gradient~\cite{zeiler2014visualizing}}
    % % \rotatebox{90}{\small  \hspace{1mm} \text{Gradient}}
    % \includegraphics[width=0.14\textwidth]{./image/figure9_attributions_reg/jpg/000000018737_Instance_0_dx_Gradient.jpg}&
    % \includegraphics[width=0.14\textwidth]{./image/figure9_attributions_reg/jpg/000000018737_Instance_0_dy_Gradient.jpg}&
    % \includegraphics[width=0.14\textwidth]{./image/figure9_attributions_reg/jpg/000000040036_Instance_0_dx_gradient.jpg}&
    % \includegraphics[width=0.14\textwidth]{./image/figure9_attributions_reg/jpg/000000040036_Instance_0_dy_gradient.jpg}
    % \\*[-0.5mm]
    
    \rotatebox{90}{\scriptsize \hspace{3mm} \text{Mask}~\cite{fong2019understanding}}
    % \rotatebox{90}{\small \hspace{4mm} \text{ Mask }}
    \includegraphics[width=0.14\textwidth]{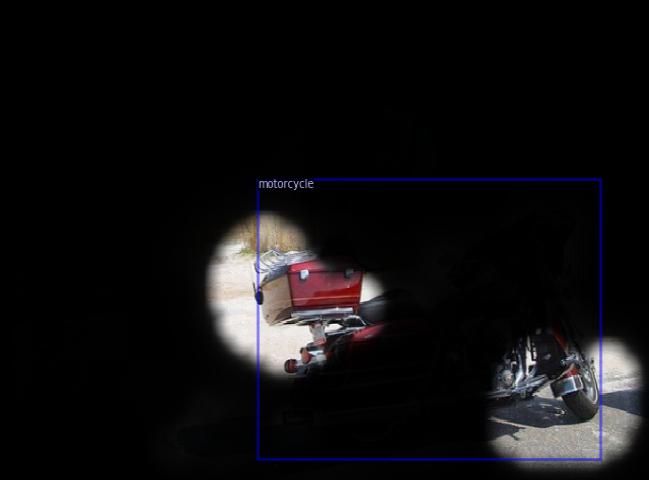}&
    \includegraphics[width=0.14\textwidth]{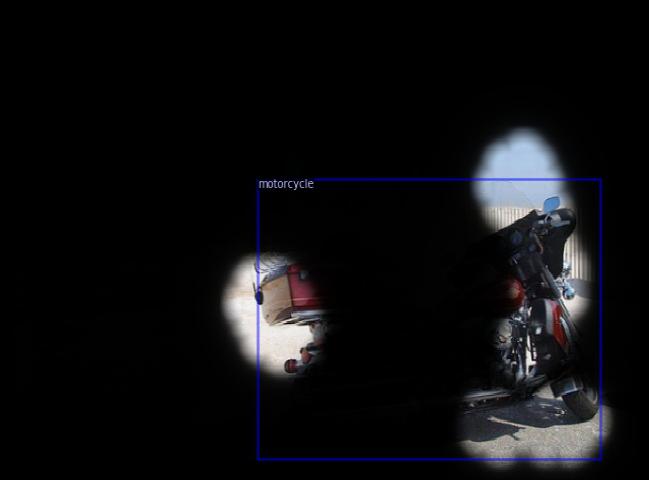}&
    \includegraphics[width=0.14\textwidth]{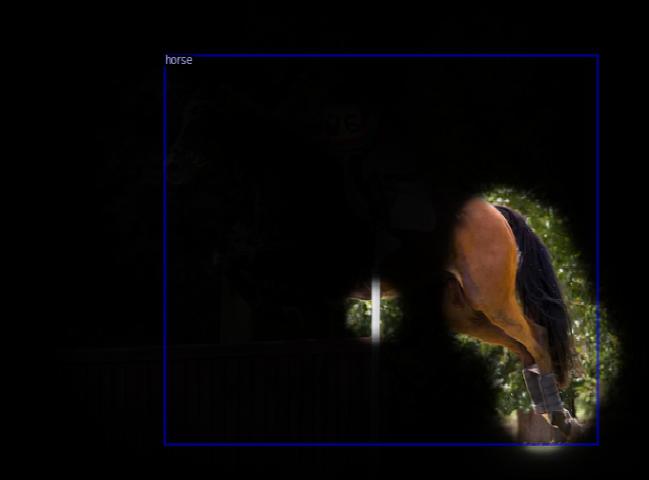}&
    \includegraphics[width=0.14\textwidth]{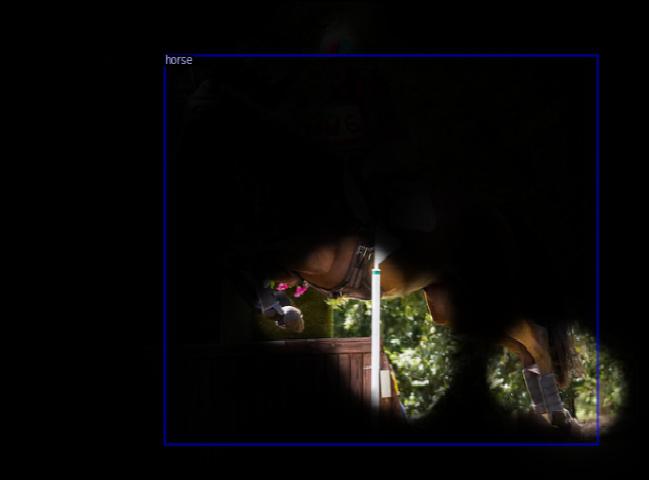}
    \\*[-0.5mm]
    \rotatebox{90}{\scriptsize \hspace{0.25mm} NormGrad~\cite{rebuffi2020there}}
    % \rotatebox{90}{\small   \text{NormGrad }}
    \includegraphics[width=0.14\textwidth]{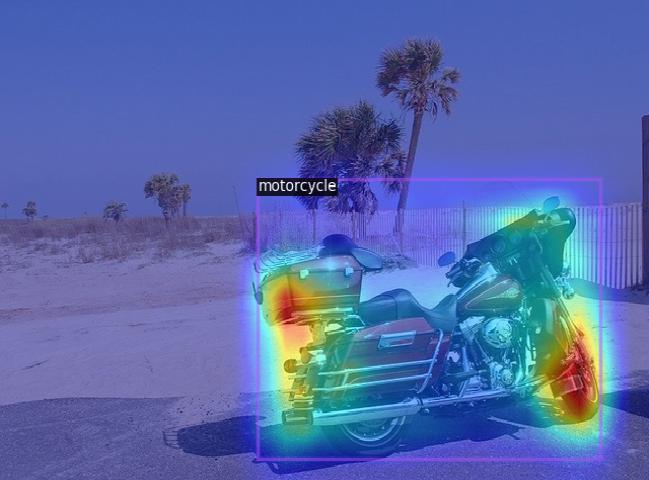}&
    \includegraphics[width=0.14\textwidth]{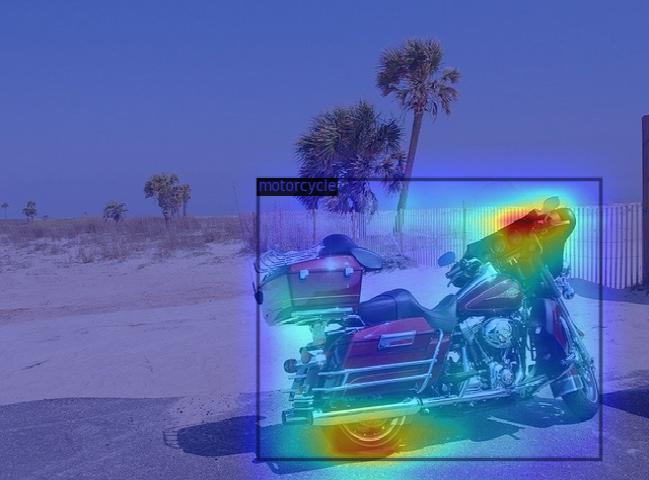}&
    \includegraphics[width=0.14\textwidth]{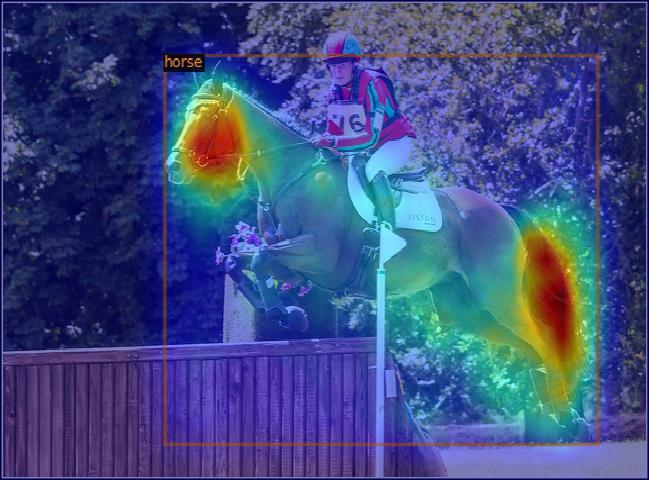}&
    \includegraphics[width=0.14\textwidth]{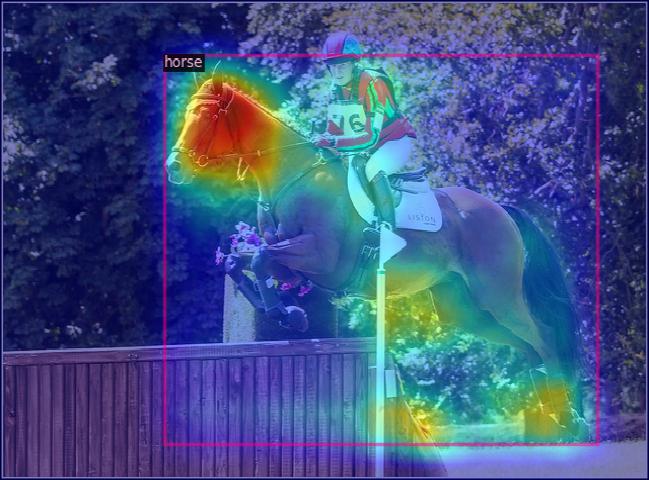}
    \\*[-0.5mm]
    
    \end{tabular}}
    \vspace{-4mm}
    \caption{
      We use attribution methods to find image regions responsible for box regressions in Mask R-CNN's second stage.
      % same  as Figure~\ref{fig:attribution_cls}.
      See Supplementary for more examples.
    }
      %   We smooth Gradient~\cite{zeiler2014visualizing} results and
    %   truncate Grad-CAM~\cite{selvaraju2017grad} results to the RoI;
    %   for Mask~\cite{fong2019understanding} we preserve area equal to half the ground-truth mask.
    %  Comparison between attribution methods on Mask R-CNN model for classification. Results of Gradient have been smoothed; we preserve the region of 0.5 target mask size in perturbation methods for better discrimination.
    
    \vspace{-5mm}
    \label{fig:attribution_reg}
    \end{figure}

\subsection{Disentangling Losses.}
\label{sec:dis_losses}
% To better understand how each element works and what distinct visual cues are learned for each task, it is crucial to disentangle the image features relevant to each loss.

Layout inversion involves a combination of classification, bounding box regression, and mask prediction~(for Mask R-CNN).
To understand distinct visual cues learned for each task, in Figure~\ref{fig:losses}, we disentangle losses by performing layout inversion on Mask R-CNN using different combinations of losses.
% In Figure~\ref{fig:losses}, we perform layout inversion on Mask R-CNN using different subsets of losses.

Inverting masks reproduces object boundaries but loses internal object structure (bird eyes, feather textures) present in full layout inversions.
Inverting category predictions only gives object parts without clear structures, similar to classification network visualization.
Inverting box regressions gives blobs in the correct locations that evoke coarse category-specific shape (bird body and beak).
Inverting class and box predictions with Mask and Faster R-CNN give similar results; the former's mask supervision seems not to affect its qualitative understanding of object appearance.

By observations, we hypothesize that regression head is trained to detect structures and boundaries of objects,
while classification head relies more on category-distinct visual cues like textures.
We employ \emph{attribution methods} to validate this idea by pointing to the image regions responsible for a model's decision on input images. 
We adapt several popular attribution methods developed for classification models to object detectors, showing attributions of classification and regression  in Figure~\ref{fig:attribution_cls} and Figure~\ref{fig:attribution_reg}.

\begin{figure}[ht!]
  \centering
  \resizebox{0.48\textwidth}{!}{%
  \begin{tabular}{@{}c@{\hspace{0.5mm}}c@{\hspace{0.5mm}}c@{\hspace{0.5mm}}c@{\hspace{0.5mm}}c@{}}
  \rotatebox{90}{\small \hspace{3mm} RetinaNet}
  \includegraphics[width=0.115\textwidth]{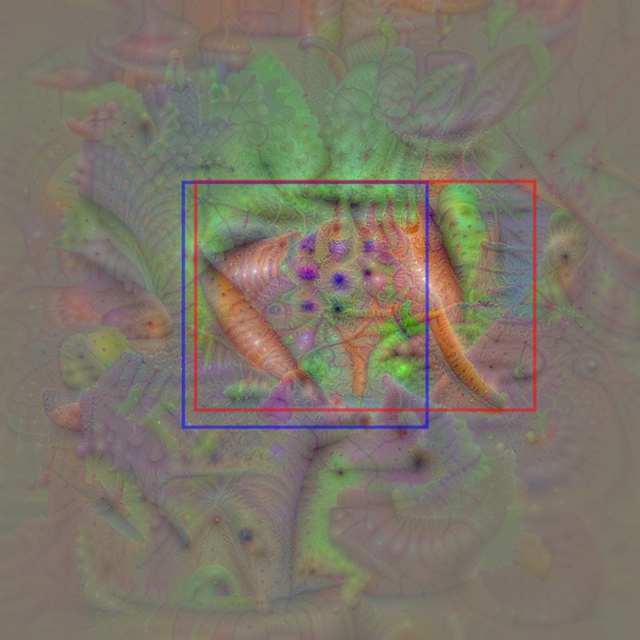}&
  \includegraphics[width=0.115\textwidth]{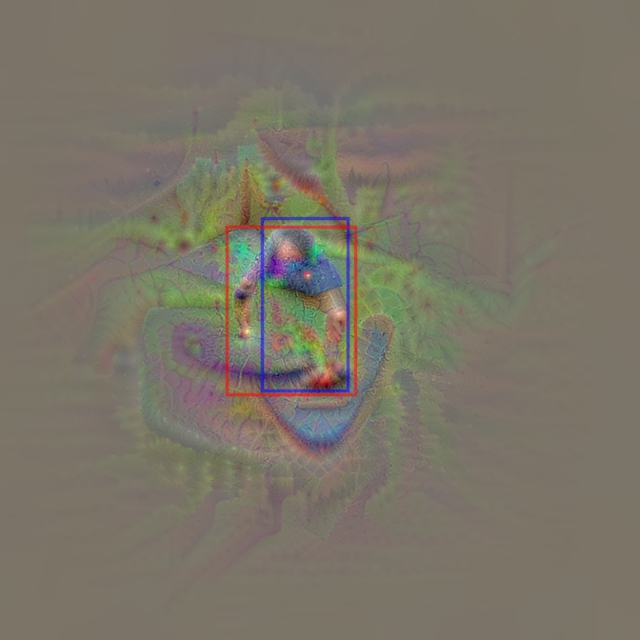}&
  \includegraphics[width=0.115\textwidth]{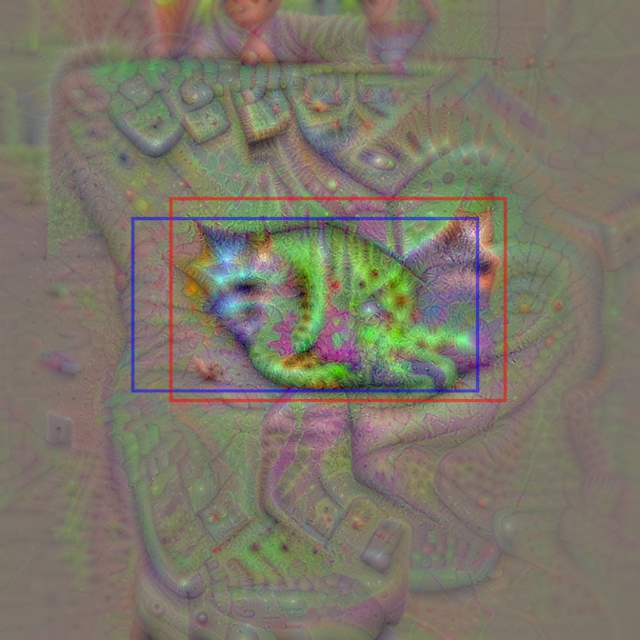}&
  \includegraphics[width=0.115\textwidth]{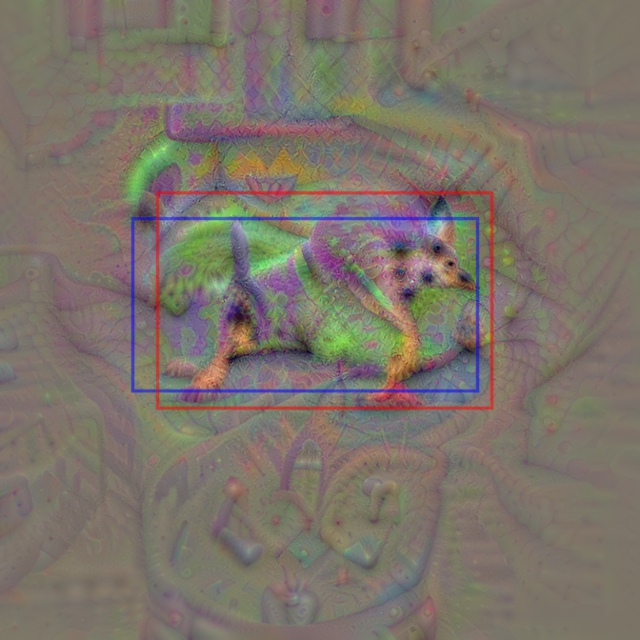}&
  \includegraphics[width=0.115\textwidth]{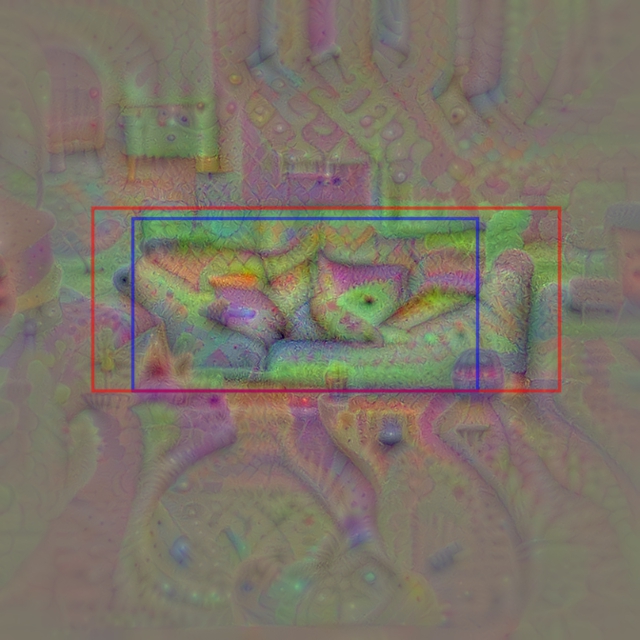}\\*[-0.5mm]
  \rotatebox{90}{\small  \hspace{4mm}Faster}
  \includegraphics[width=0.115\textwidth]{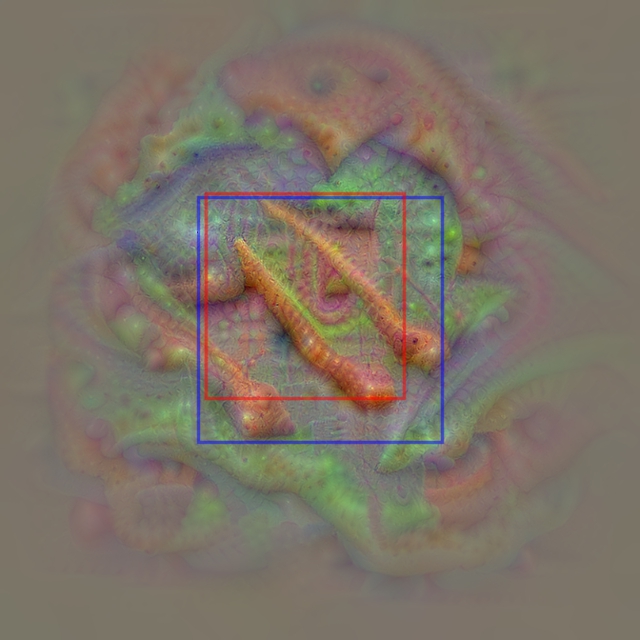}&
  \includegraphics[width=0.115\textwidth]{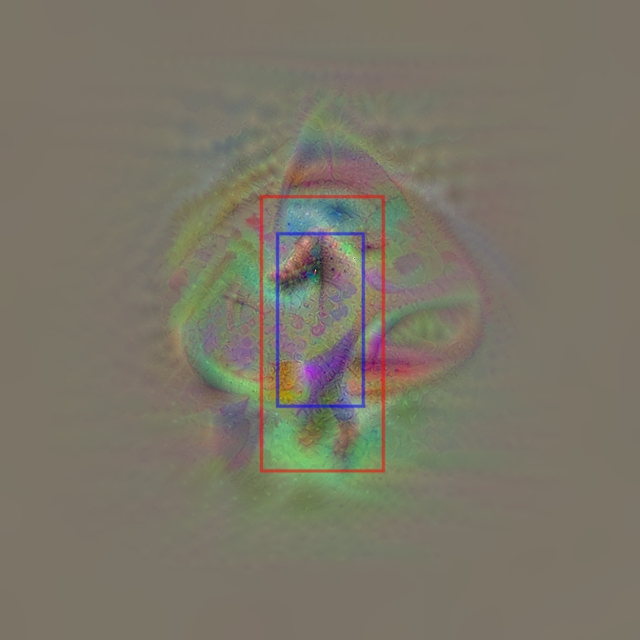}&
  \includegraphics[width=0.115\textwidth]{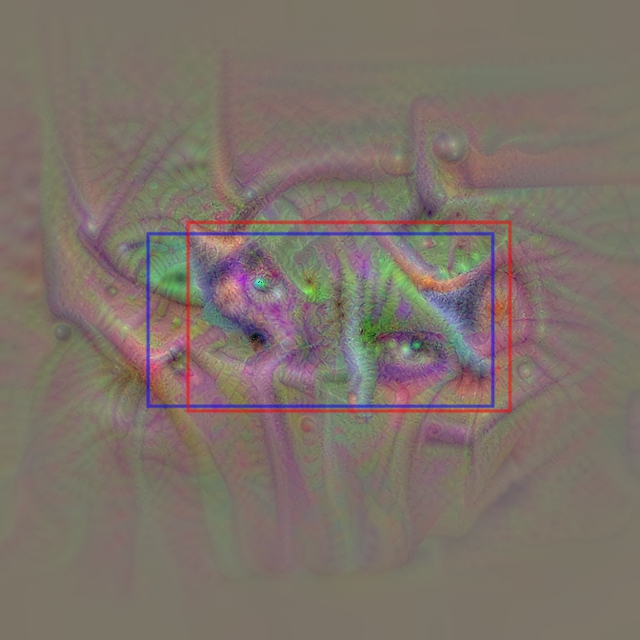}&
  \includegraphics[width=0.115\textwidth]{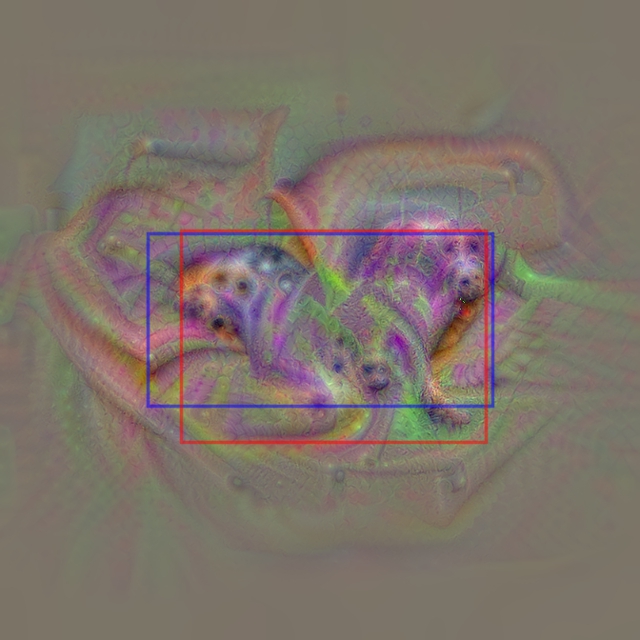}&
  \includegraphics[width=0.115\textwidth]{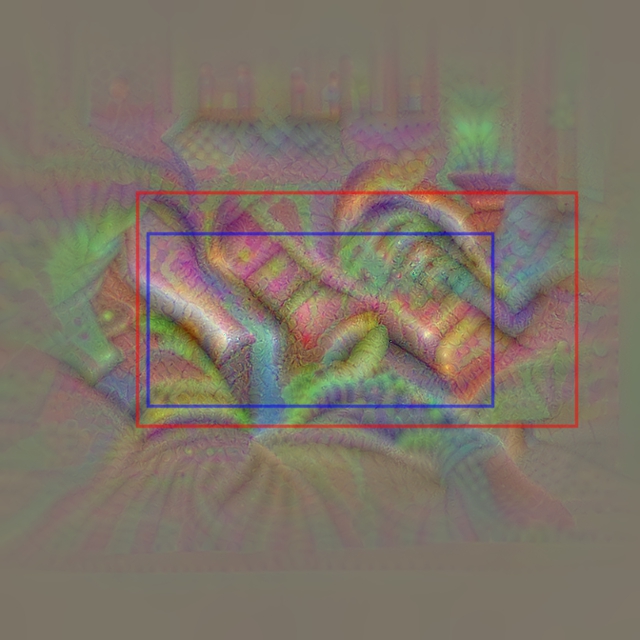}\\
  \small \text{Carrot}&
  \small \text{Person}&
  \small \text{Cat} &
  \small \text{Dog} &
  \small \text{Couch}
  \end{tabular}}
  \vspace{-3mm}
  \caption{
    We generate category-specific images by optimizing the predictions of RetinaNet and Faster R-CNN on a single anchor.
    Anchors are shown in blue, and predicted object boxes in red.
    % Both detectors rely heavily on context to make their predictions.
  }
  \label{fig:class_specific}
  \vspace{-3mm}
  \end{figure}

% \begin{figure}[ht!]
% \centering
% % \vspace{-3mm}
% \includegraphics[width=0.12\textwidth]{./image/figure6_neuron_category/jpg/Faster_51_2_detect.jpg}~
% \includegraphics[width=0.12\textwidth]{./image/figure6_neuron_category/jpg/Faster_16_3_2_detect.jpg}~
% \includegraphics[width=0.12\textwidth]{./image/figure6_neuron_category/jpg/Retina_0_2_2_detected.jpg}~
% \includegraphics[width=0.12\textwidth]{./image/figure6_neuron_category/jpg/Faster_57_2_detect.jpg}
% \vspace{-5mm}
% \caption{
%     Detection results of individual object visualization in Figure~\ref{fig:class_specific}.
%     While only maximizing prediction scores on an anchor, detectable contexts emerge with high probability.
% }
% \label{fig:detection}
% \vspace{-5mm}
% \end{figure}
\begin{figure}[ht!]
\centering
\resizebox{0.48\textwidth}{!}{%
\begin{tabular}{@{}c@{\hspace{0.5mm}}c@{\hspace{0.5mm}}c@{\hspace{0.5mm}}c@{}}
    \includegraphics[width=0.12\textwidth]{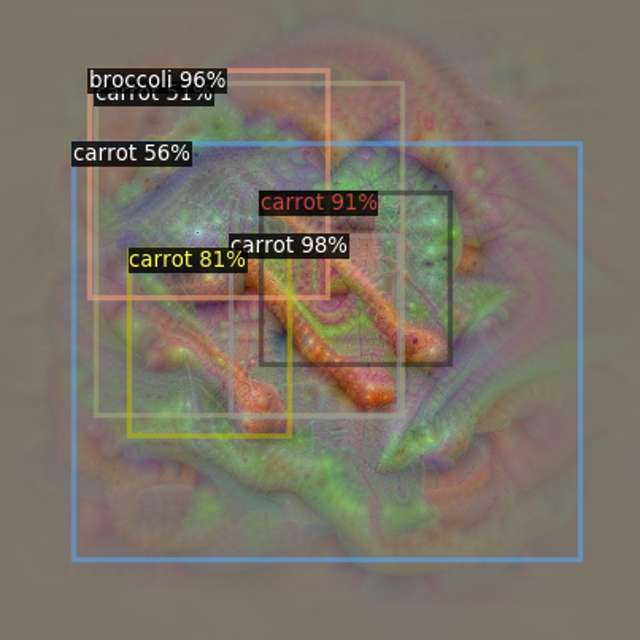}
    \includegraphics[width=0.12\textwidth]{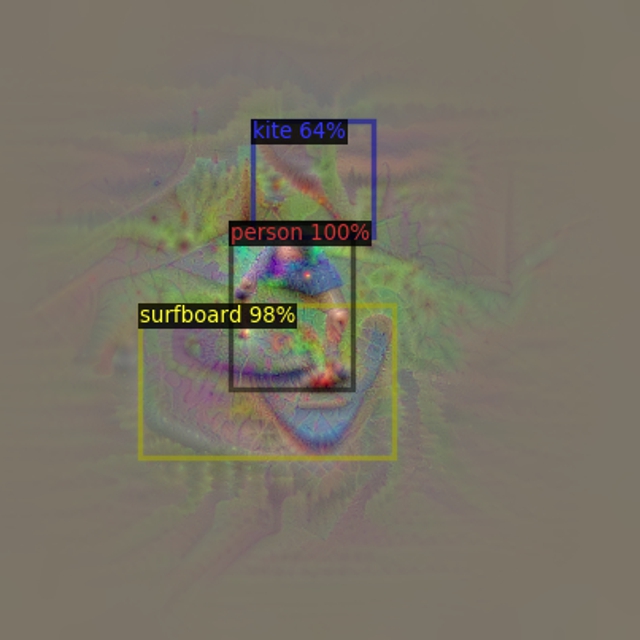}
    \includegraphics[width=0.12\textwidth]{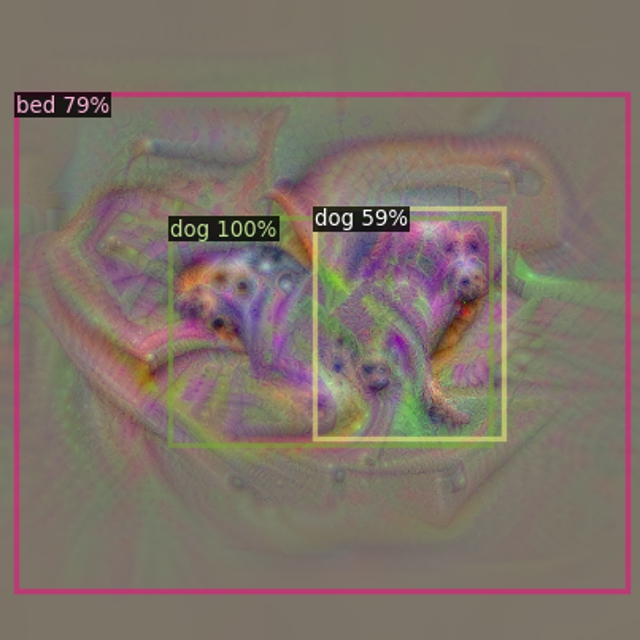}
    \includegraphics[width=0.12\textwidth]{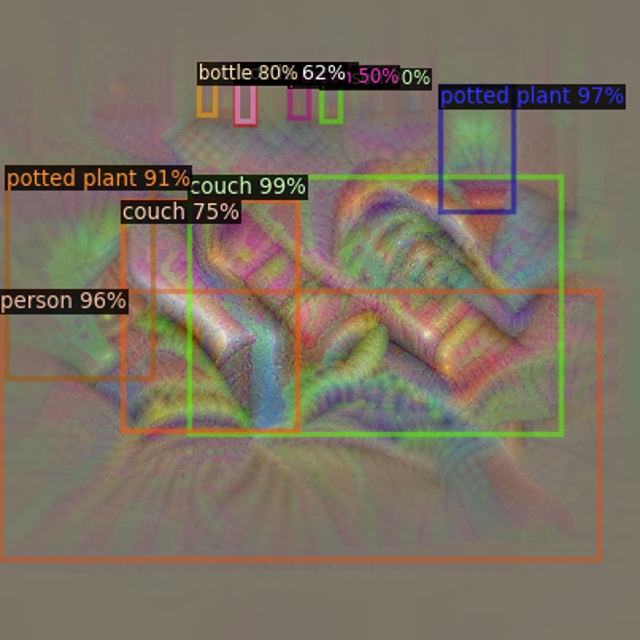}
\end{tabular}
}
\vspace{-3mm}
\caption{
    Detection results of individual object visualizations in Figure~\ref{fig:class_specific}.
    While only maximizing prediction scores on an anchor, detectable contexts emerge with high probability.
}
\label{fig:detection}
\vspace{-7mm}
\end{figure}
Attribution methods reveal that classification head and regression head pay attention to different regions, which supports our hypothesize.
Classification head focuses on local features like animal heads, hair textures.
The regression head has a better understanding of global structures and pays category-specific attention to boundaries and objects' corners.
From this perspective, we suggest the popular sibling head setting in object detector (classification and regression heads share almost the same weights)
may not be the best choice since these two heads focus on different regions. 
We find our hypothesis could be supported by some new object detector design of sibling heads~\cite{song2020revisiting,jiang2018acquisition,wu2020rethinking}.

\subsection{Visualizing Individual Objects.}
\label{sec:vis_obj}
% So far, we have inverted entire scene layouts, which requires both recognition of foreground objects and suppression of background regions.
% Alternatively, we could further explore the category-specific features by optimizing only a chosen anchor to be classified as the target category without constraining layouts and other anchors.
% We show our visualization results for several categories in Figure~\ref{fig:class_specific}.

Inverting entire scene layouts requires both recognition of foreground objects and suppression of background regions. 
To further explore the category-specific information learned by detectors, we visualize individual objects in Figure~\ref{fig:class_specific}.
We optimize the image to maximize only a single anchor's probability of being classified as the chosen category without constraining its regressed box position or any outputs from other anchors.

Visualizing individual objects in this way allows us to more precisely probe the priors learned by detectors.
In Section~\ref{sec:sur_con} we see that detectors learn consistent contexts for objects, and in Section~\ref{sec:var_scale} we find that detectors rely on different features to detect objects at different scales.

% To further explore the object detectors' category-specific features, in Figure~\ref{fig:class_specific},we visualize individual objects without layout constraints. 
% Specifically, we optimize the input only to maximize a chosen anchor's probability of being classified as the target category, without constraining layouts, other anchors' outputs, and regression of this anchor.   

\subsubsection{Object Context.}
\label{sec:sur_con}
% These surrounding contexts, emerging while visualizing an individual object, imply that object detectors have learnt these contexts as visual cues for detection, which may come from the bias of the dataset.
% We conduct design experiments on RetinaNet to explore the relationships between surrounding contexts and visualized objects quantitatively. 
In Figure~\ref{fig:class_specific} we only maximize the classification score of a single anchor.
As expected, this causes the specified object to appear.
Surprisingly other background objects also consistently emerge:
broccoli near the carrot;
a bed and couch below the dog and cat;
and a potted plant near the couch.
% In Figure~\ref{fig:class_specific}, while only maximizing the prediction result of target category on a selected anchor,
% we find that there consistently emerge category-specific surrounding contexts,
% including background and objects: 
% the green broccoli in carrot visualization; 
% the bed and couch below the dog and cat;
% the potted plant in couch visaulization.
\begin{figure}[ht!]
\centering
    % \begin{subfigure}
    %     \centering
        \includegraphics[width=0.235\textwidth]{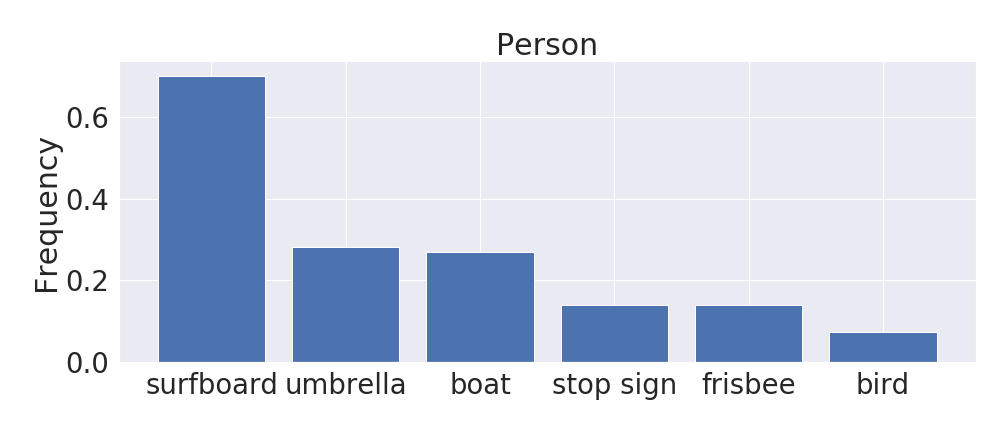}
        \includegraphics[width=0.235\textwidth]{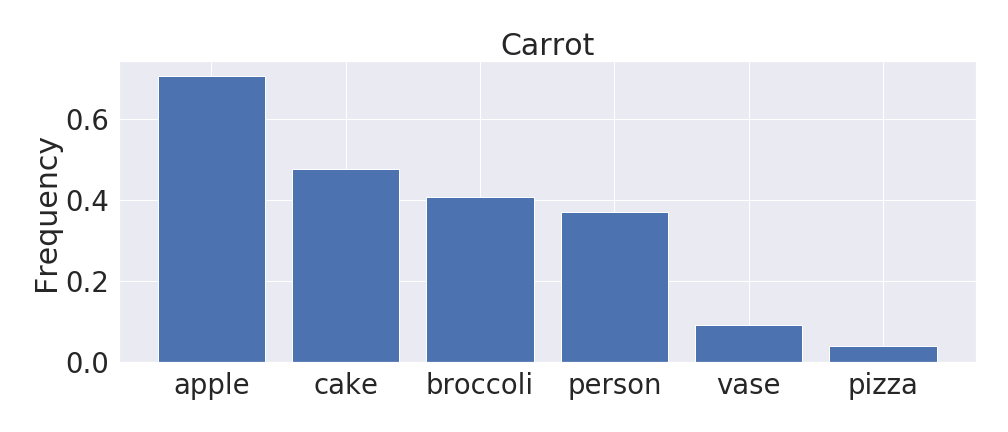}
    % \end{subfigure}
    % \begin{subfigure}
        % \centering
        \includegraphics[width=0.235\textwidth]{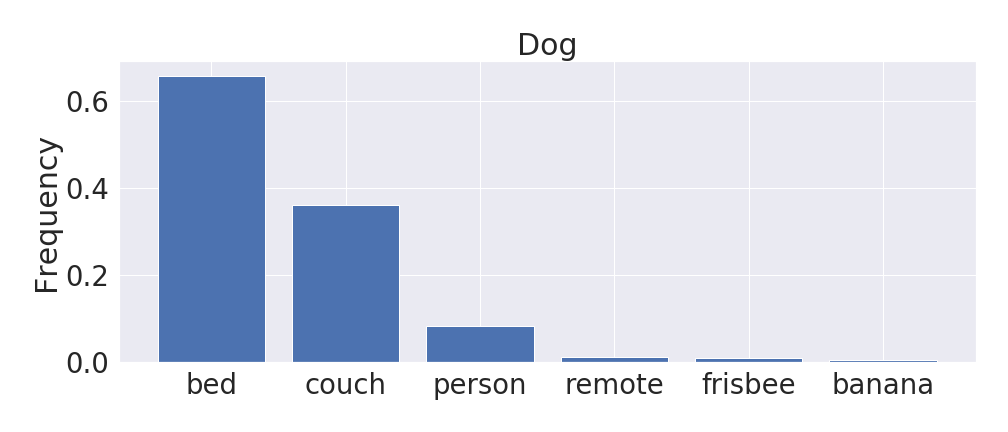}
        \includegraphics[width=0.235\textwidth]{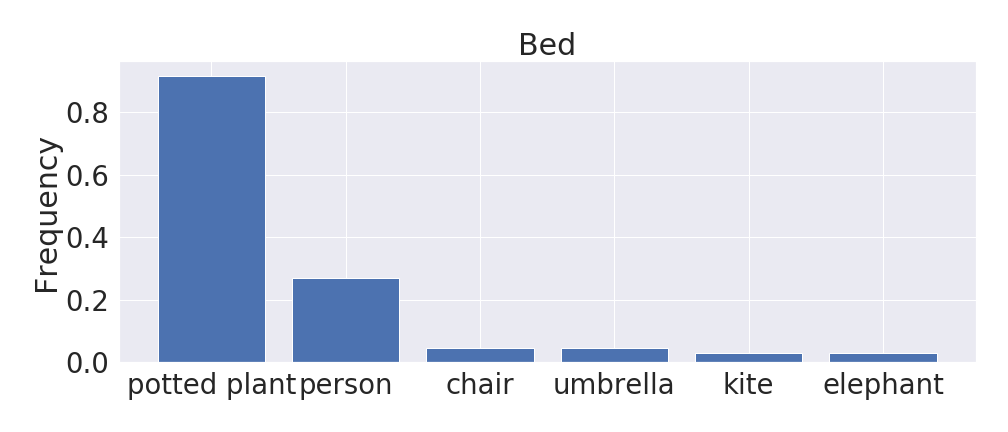}
    % \end{subfigure}
    \vspace{-5mm}
    \caption{We display the surrounding contexts with highest frequency while visualizing individual car, cat, horse and keyboard object for RetinaNet.
    Surrounding contexts show a consistent and category-specific pattern.
    See supplementary for more data.}
    \label{fig:sur_context_freq}
    \vspace{-3mm}
\end{figure}
\begin{figure}[ht!]
    \centering
    \includegraphics[width=0.49\textwidth]{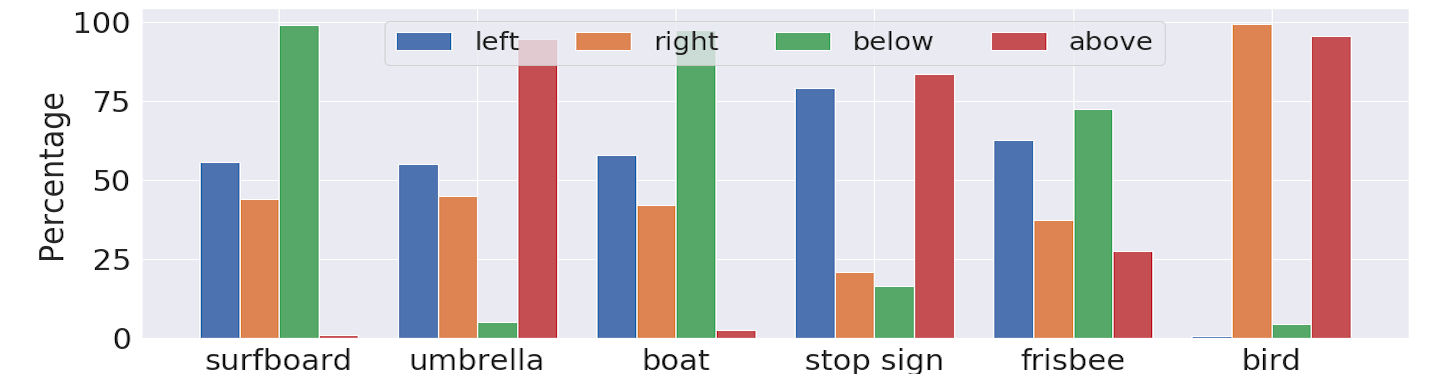}
    \vspace{-5mm}
    \caption{
      We show the frequency of relative positions of contextual objects when visualizing people on RetinaNet.
%       We show the relative position between surrounding contexts and visualized objects when visualizing a person object.
    }
    \label{fig:surrounding_position}
    \vspace{-7mm}
\end{figure}

These emergent \emph{contextual objects} are visible to detectors as well, as demonstrated in Figure~\ref{fig:detection}.
This suggests that detectors may learn \emph{motifs} of commonly occuring objects -- e.g. that carrots occur in groups near other vegetables -- which may be important signals for recognition.

% Furthermore, these surrounding contexts are detectable by object detectors with high confidence, which is shown in Figure~\ref{fig:detection}. 
% We hypothesize these contexts may be strong visual cues for detectors.
We quantify these motifs by scaling up the experiments of Figures~\ref{fig:class_specific} and \ref{fig:detection}.
For each category, we generate 1000 visualizations with a fixed anchor and different random seeds.
We feed these images back to the detector, and keep track of other confidently detected objects~($>0.5$).

% We quantitatively explore the relationships between the surrounding contexts and visualized objects by repeating the same procedure as Figure~\ref{fig:class_specific} and Figure~\ref{fig:detection}.
% We visualize individual objects at the center of images and detect them using the same detector.
% We only count the surrounding objects' detection results with relatively high confidence~($>0.5$). 
% For each category, we collect 1000 visualizations with a fixed anchor and different random seeds.

\textbf{Contextual Frequency.}
For each category, we count the frequency with which objects of other categories emerge.
Figure~\ref{fig:sur_context_freq} shows the six most frequently apearing objects when generating images containing a person, carrot, dog, or bed.%
\footnote{Instances of the same category are only counted once per image.}
We see detectors learn contexts which are consistent but not universal: surfboards appear near most but not all people.
Contexts are category-specific, and may be asymmetric: people emerge near $>20\%$ of beds, but beds rarely emerge near people.

%\textbf{Emerging Frequency.}
%We count the emerging frequency of surrounding contexts when visualizing individual objects;
%duplicate instances of the same category are only counted once in each visualization.
%In Figure~\ref{fig:sur_context_freq}, we show the top 6 categories with the highest frequency when visualizing person, carrot, dog, and bed.
%Surrounding contexts appear with high probability and possess category-specific pattern:
%visualizing objects of different categories leads to distinct surrounding contexts distributions.

\textbf{Relative Position.} 
We further analyze not only \emph{what} contextual objects emerge, but also \emph{where} they are located.
We classify each emergent contextual object as above/below and left/right of the central object based on object centers.
Figure~\ref{fig:surrounding_position} shows the frequency of contextual objects appearing at different positions when visualizing people.
The detector learns consistent spatial trends: surfboards and boats appear below people, while birds and umbrellas appear above people.
Left/right biases may reveal photography bias in the dataset (such as birds right of people).

% Detectors not only learn \emph{what} objects co-occur, but also common \emph{spatial patterns} among those objects.

%We further analyze the relative positions between contexts and visualized objects.
%We compute the center of surrounding contexts and categorize left or right, above or below, based on visualized objects centers.
%In Figure~\ref{fig:surrounding_position}, we show the relative position statistics when visualizing a person:
%surfboards and boats are biased to appear below the person, while birds and umbrellas are biased to be above.
%Some strange left-right biases are hard to explain intuitively and may come from datasets.
% : birds always appear on the person's right side.

One might expect detectors to be less context-dependent than classifiers since they are trained for localization.
However our experiments show that detectors learn canonical patterns of co-occurring objects, including common spatial positions.
It seems likely that detectors also use these common contexts to help recognize objects, which accords with the research of objects and their context in early vision literature \cite{oliva2007role, rabinovich2007objects}. 
We show context statistic in supplementary.

\begin{figure}
\centering
\resizebox{0.48\textwidth}{!}{%
\begin{tabular}{@{}c@{\hspace{0.5mm}}c@{\hspace{0.5mm}}c@{\hspace{0.5mm}}c@{\hspace{0.5mm}}c@{}}

\rotatebox{90}{ \small \hspace{2mm} Scale 1}
\includegraphics[width=0.096\textwidth]{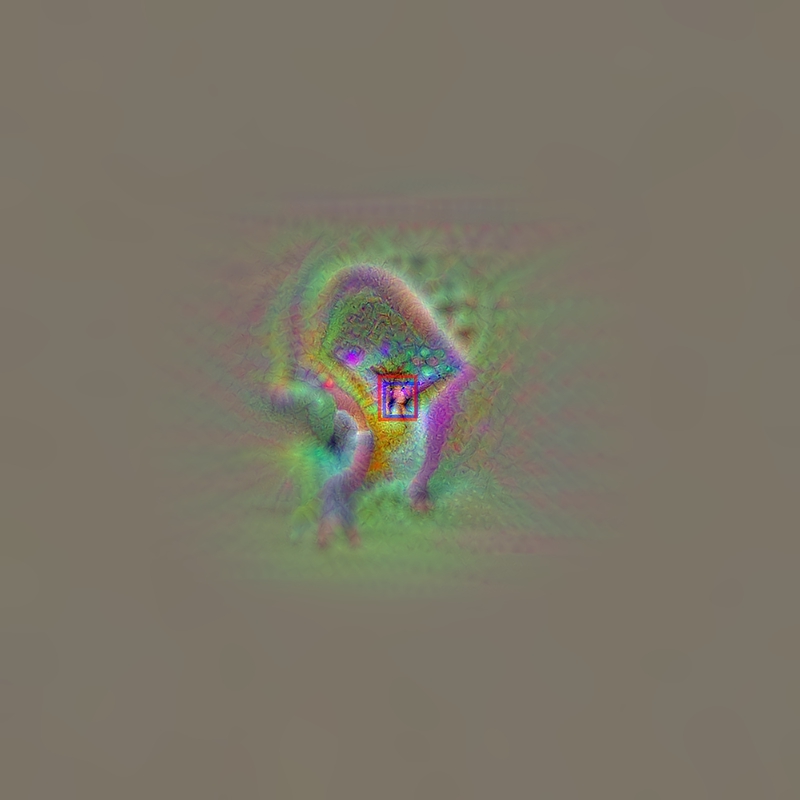}&
\includegraphics[width=0.096\textwidth]{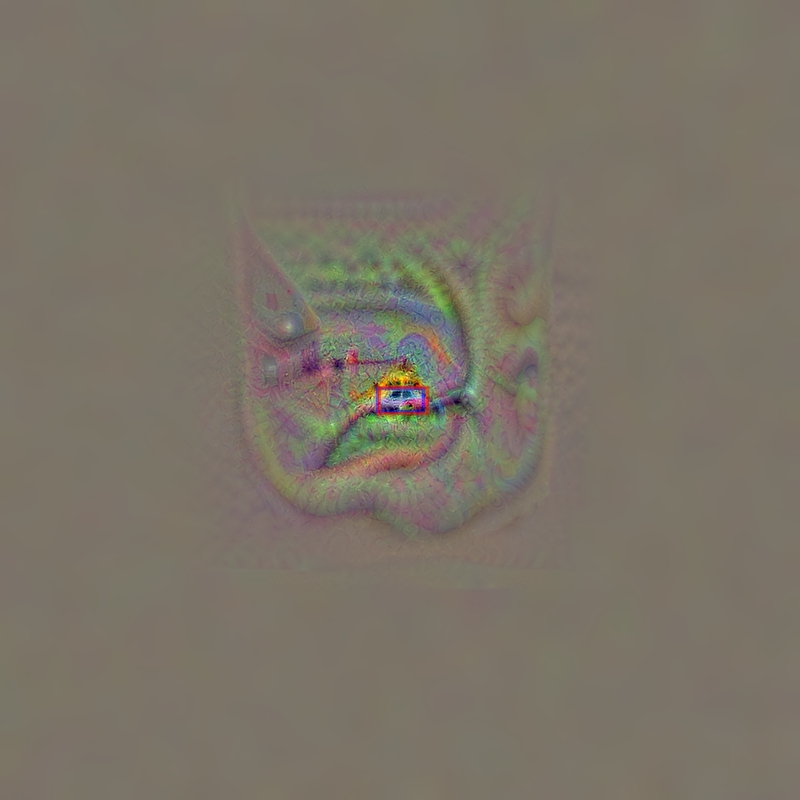}&
\includegraphics[width=0.096\textwidth]{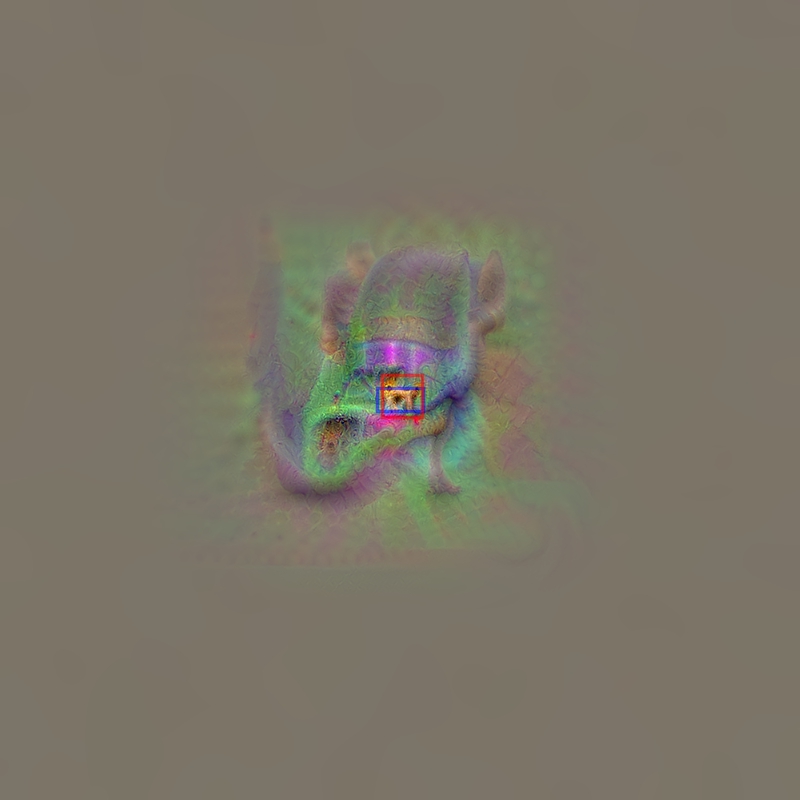}&
\includegraphics[width=0.096\textwidth]{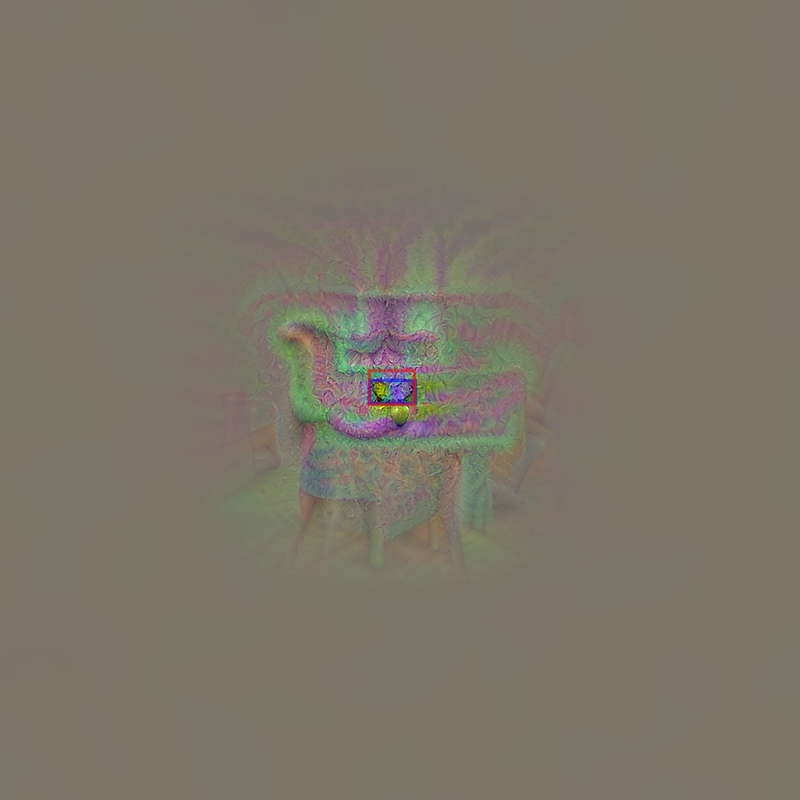}&
\includegraphics[width=0.096\textwidth]{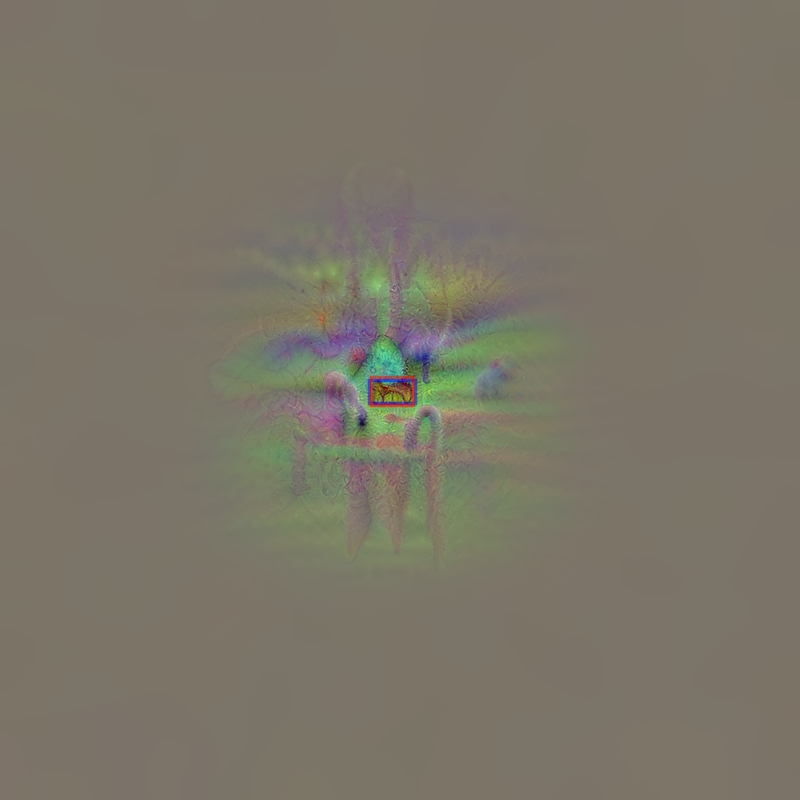}
\\*[-0.5mm]

\rotatebox{90}{\small \hspace{2mm} Scale 2}
\includegraphics[width=0.096\textwidth]{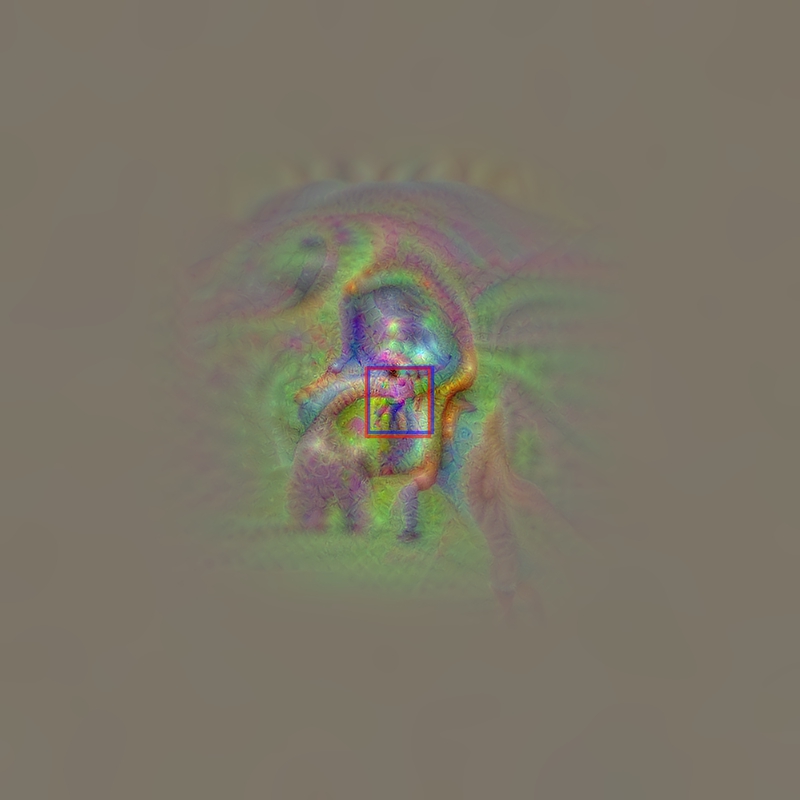}&
\includegraphics[width=0.096\textwidth]{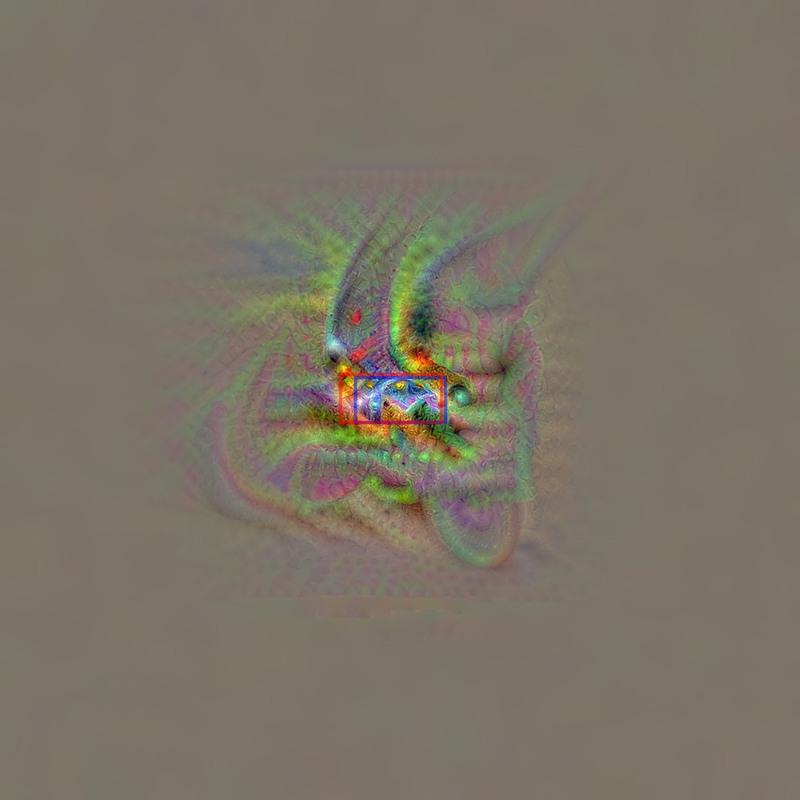}&
\includegraphics[width=0.096\textwidth]{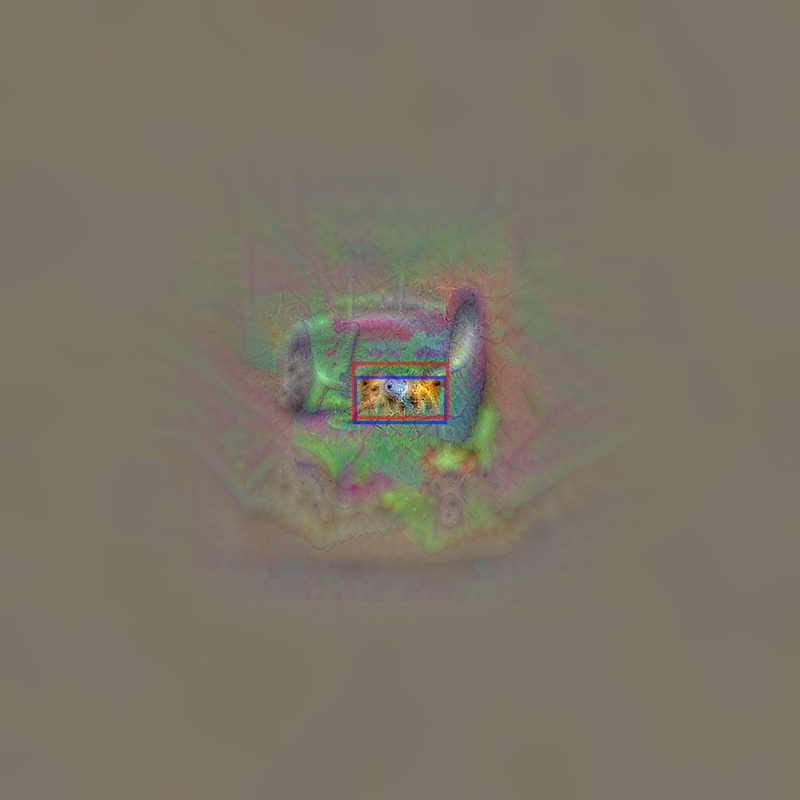}&
\includegraphics[width=0.096\textwidth]{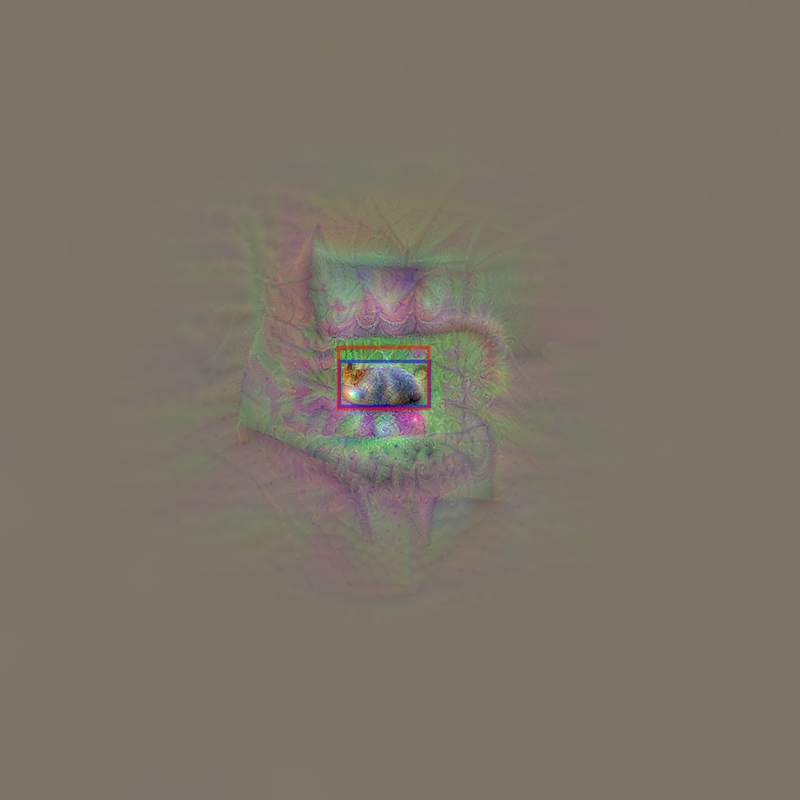}&
\includegraphics[width=0.096\textwidth]{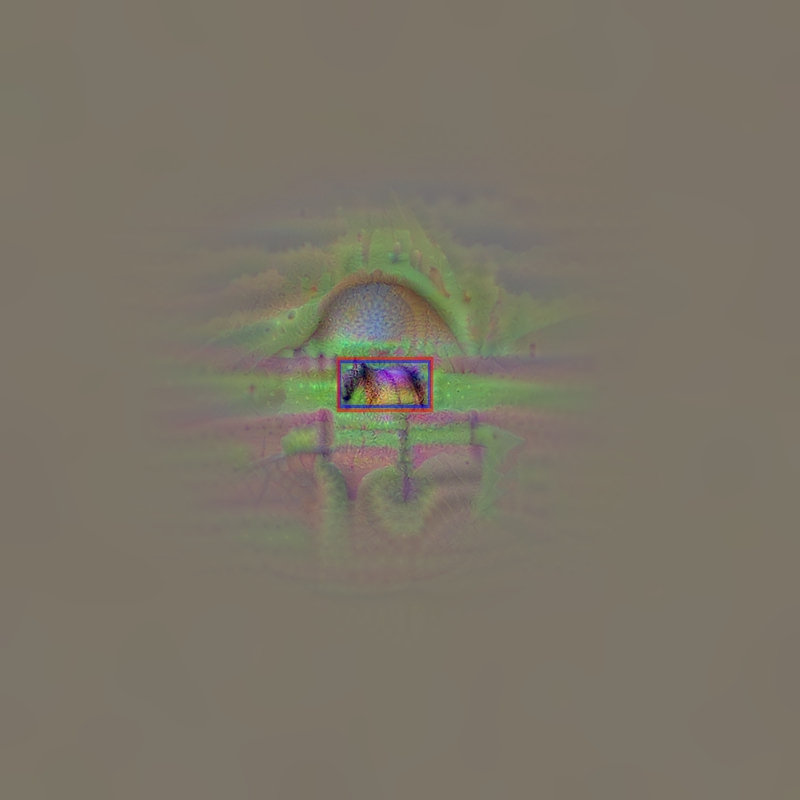}
\\*[-0.5mm]

\rotatebox{90}{\small\hspace{2mm} Scale 3}
\includegraphics[width=0.096\textwidth]{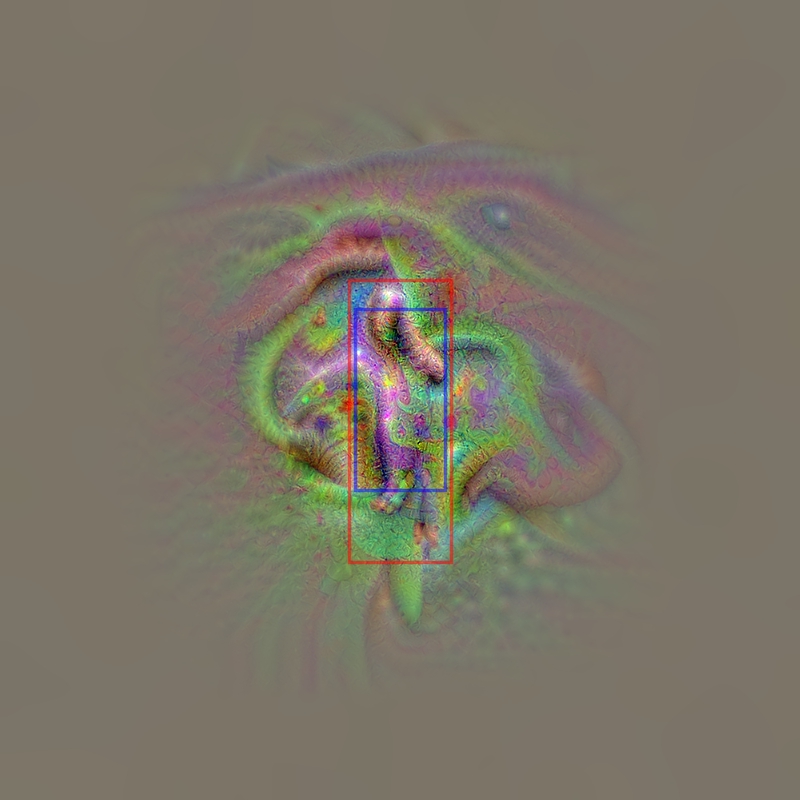}&
\includegraphics[width=0.096\textwidth]{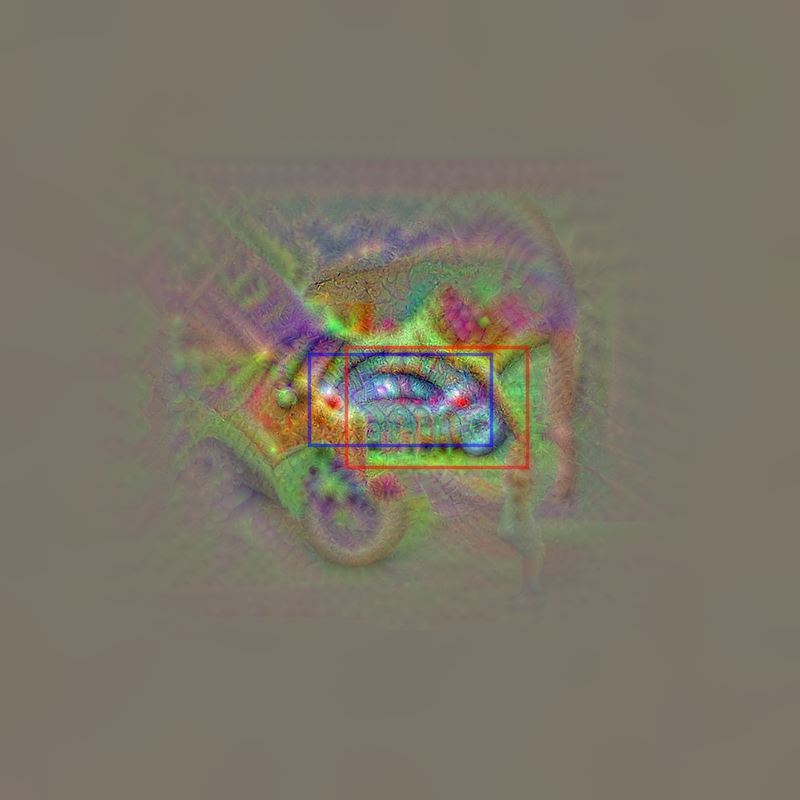}&
\includegraphics[width=0.096\textwidth]{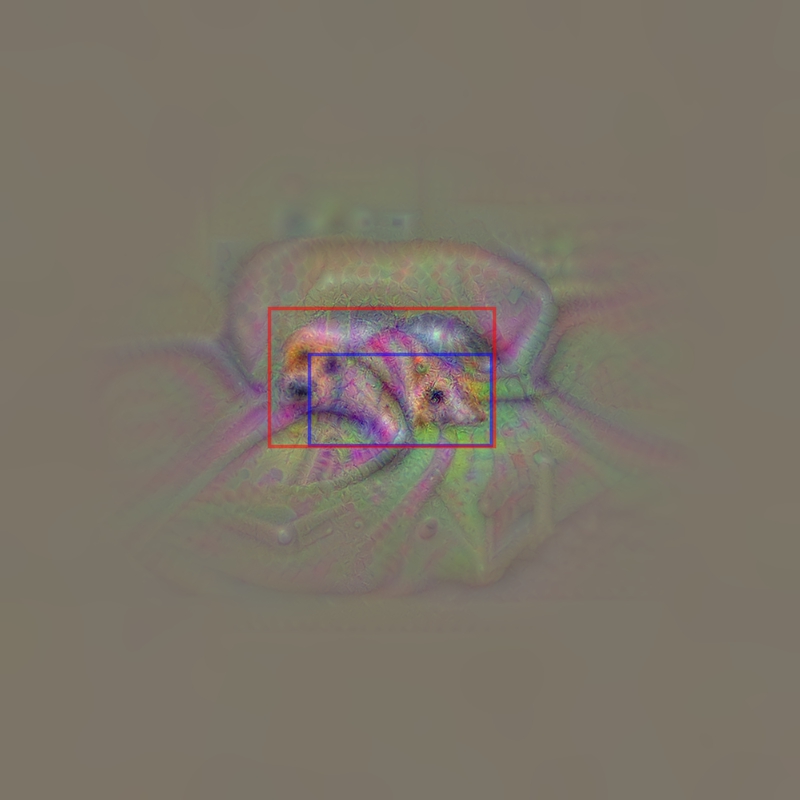}&
\includegraphics[width=0.096\textwidth]{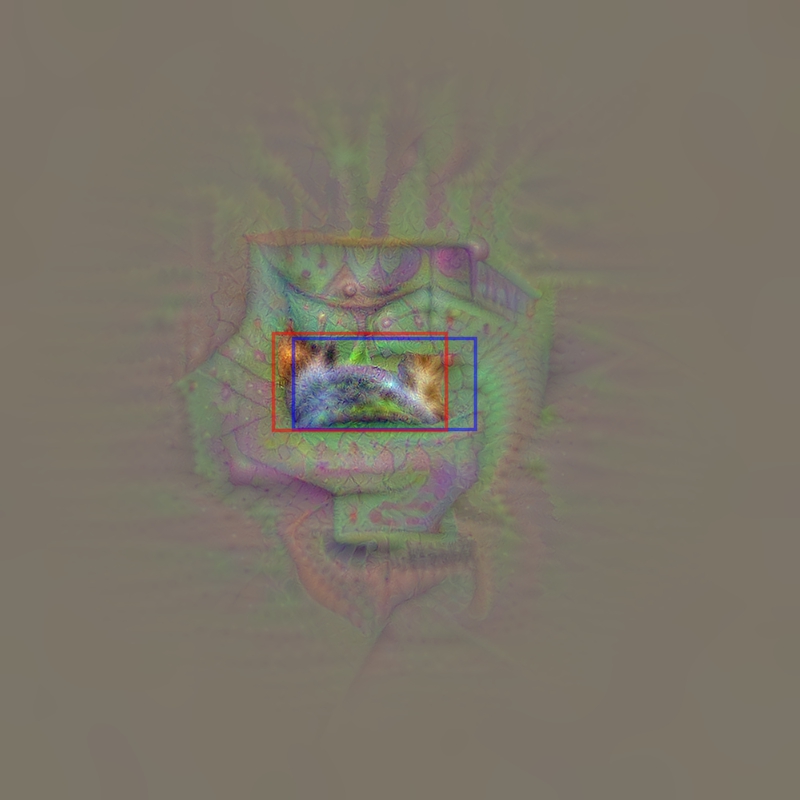}&
\includegraphics[width=0.096\textwidth]{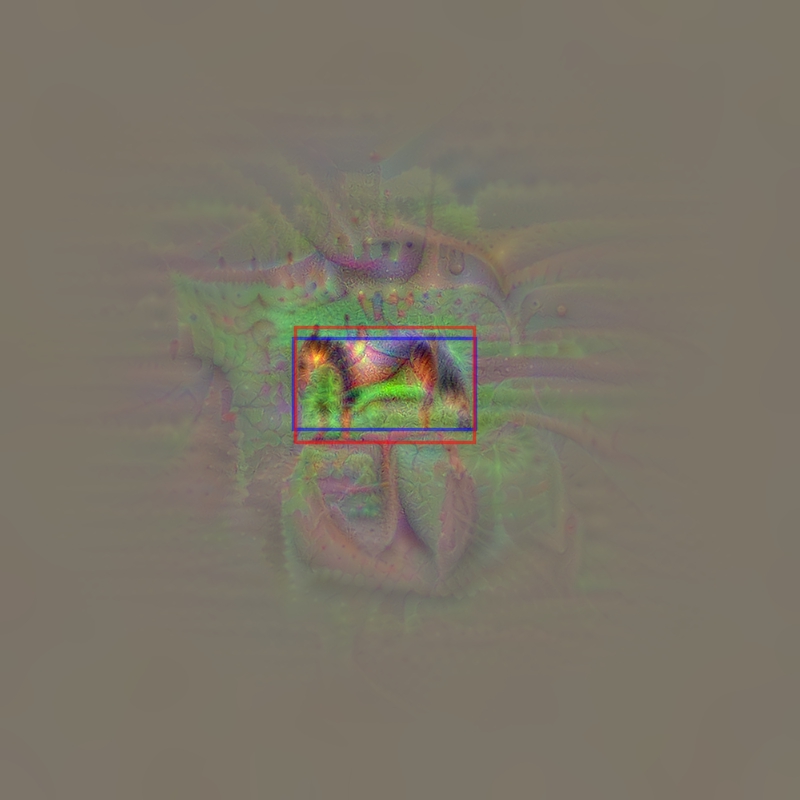}
\\*[-0.5mm]

\rotatebox{90}{\small \hspace{2mm}Scale 4}
\includegraphics[width=0.096\textwidth]{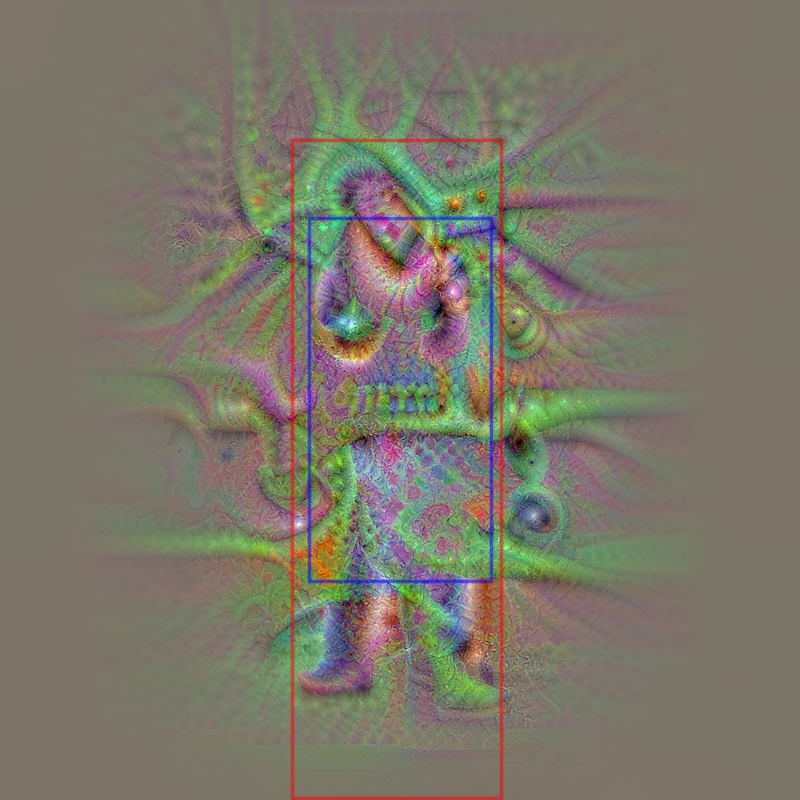}&
\includegraphics[width=0.096\textwidth]{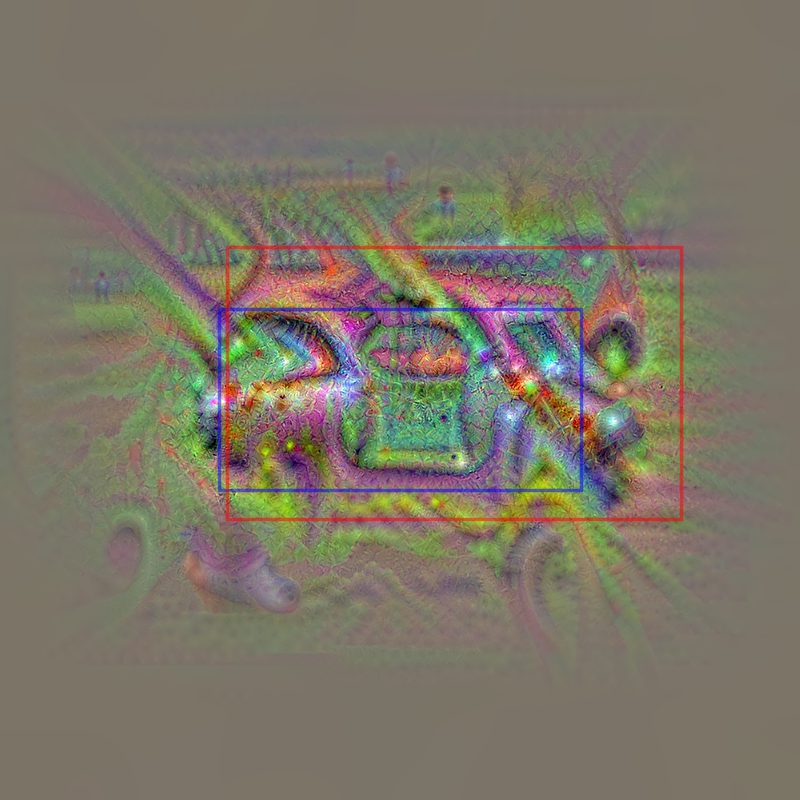}&
\includegraphics[width=0.096\textwidth]{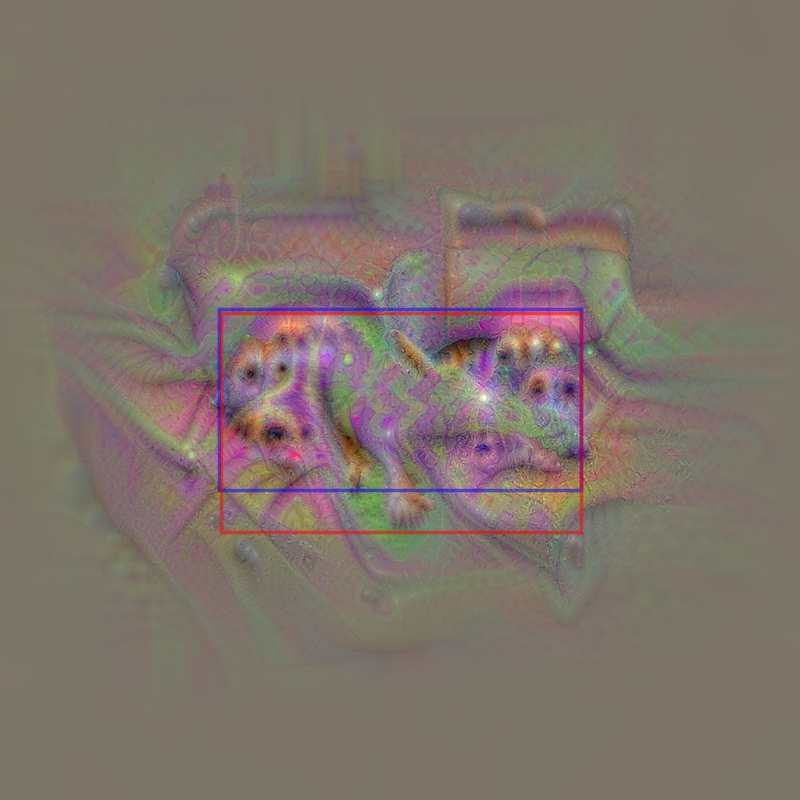}&
\includegraphics[width=0.096\textwidth]{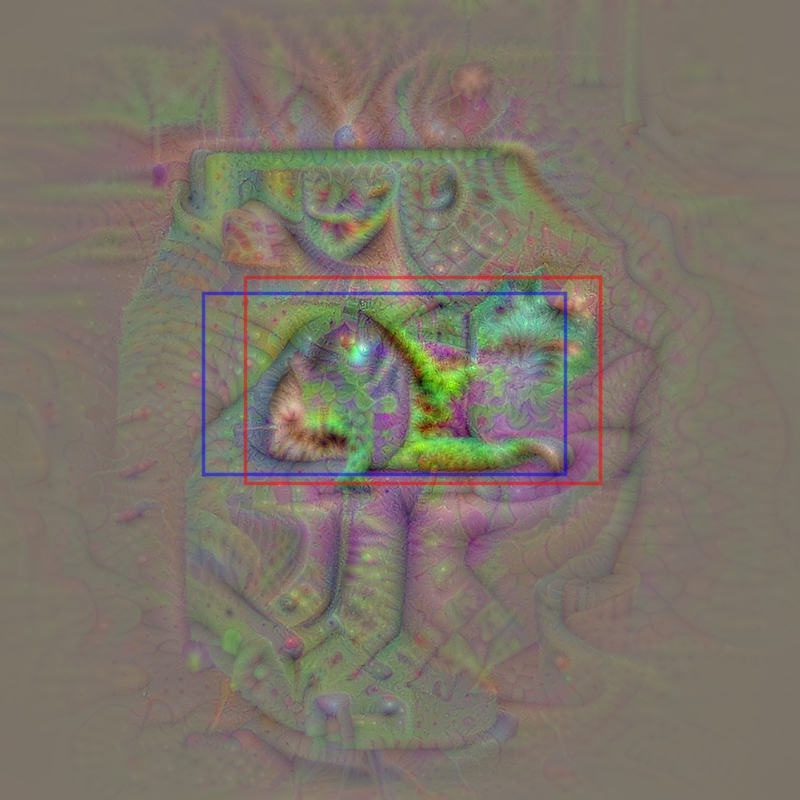}&
\includegraphics[width=0.096\textwidth]{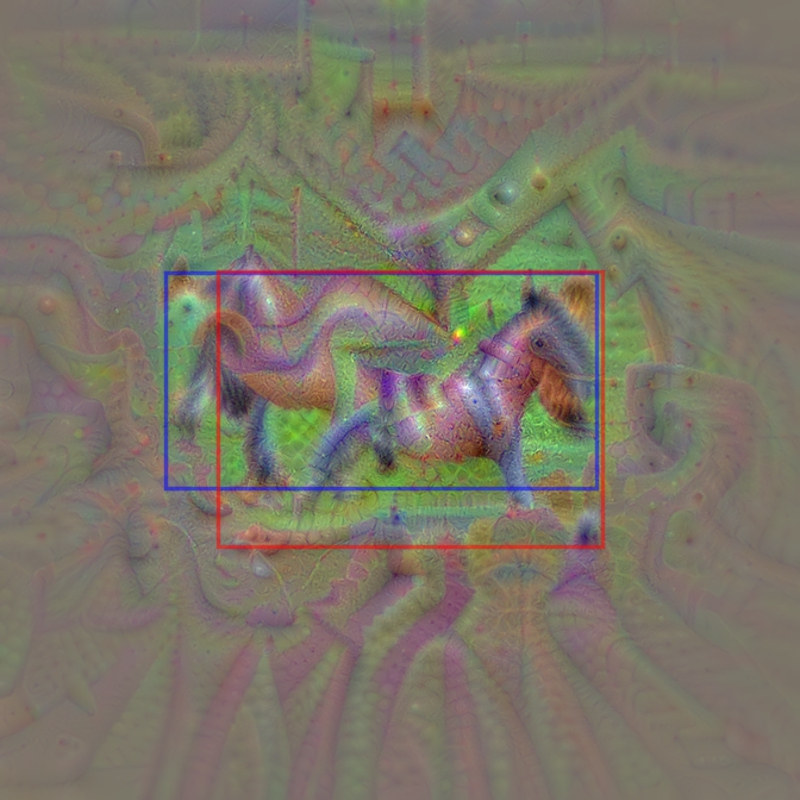}
\\*[-0.5mm]

\rotatebox{90}{\small \hspace{2mm}Scale 5}
\includegraphics[width=0.096\textwidth]{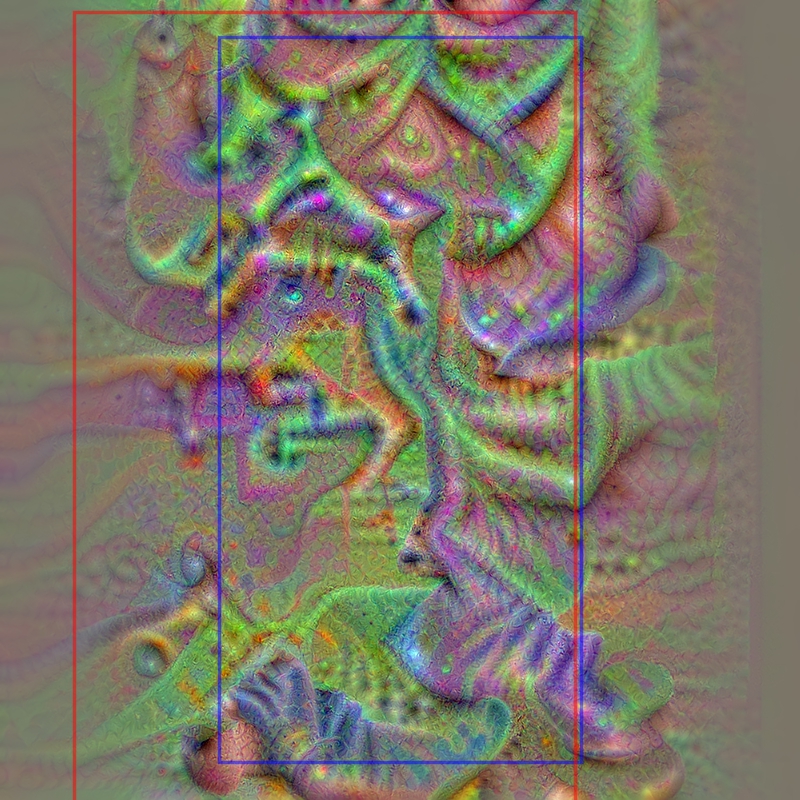}&
\includegraphics[width=0.096\textwidth]{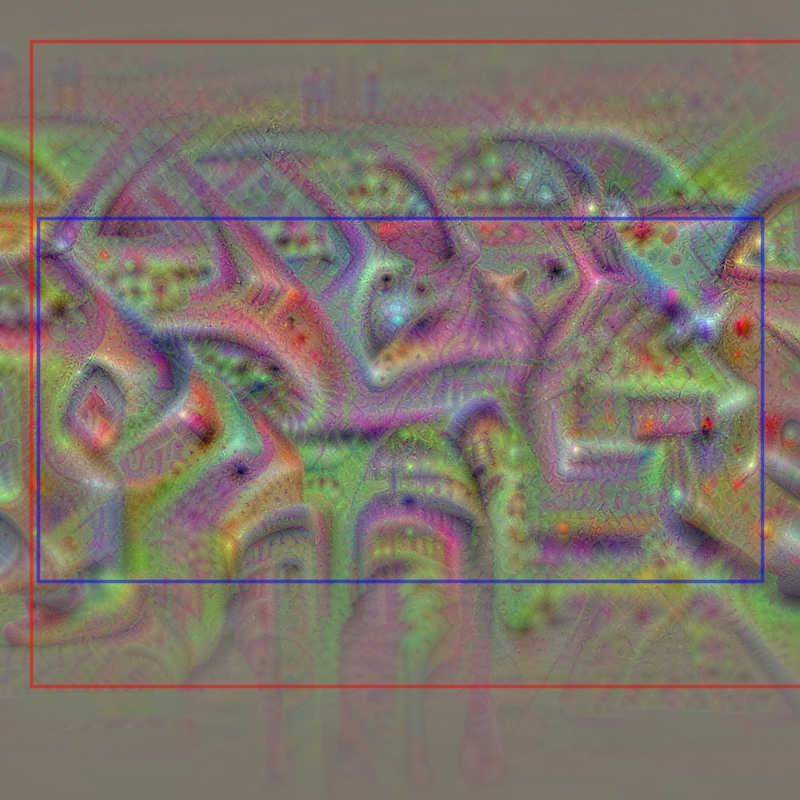}&
\includegraphics[width=0.096\textwidth]{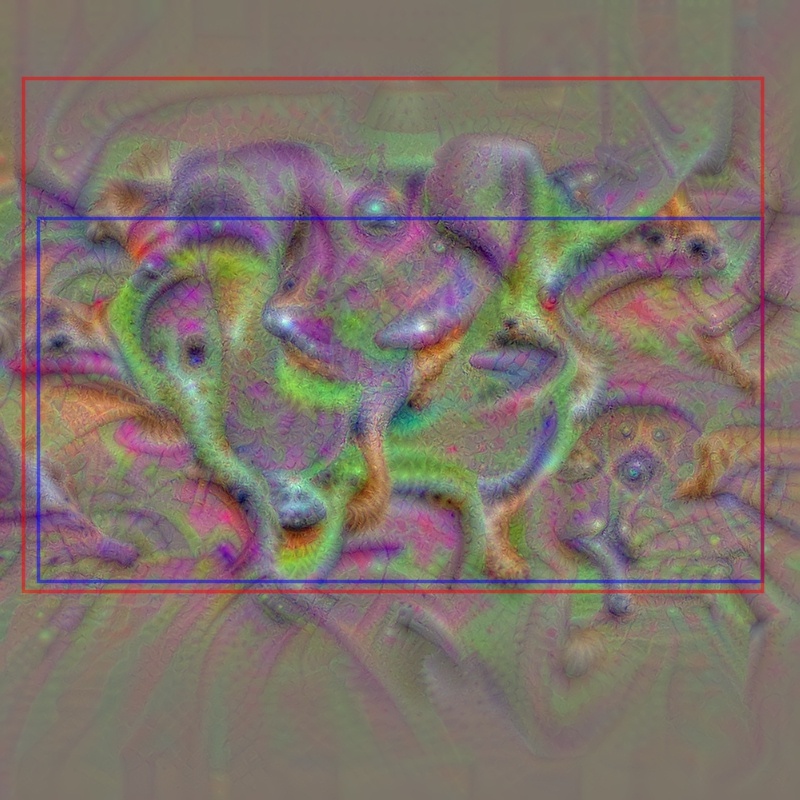}&
\includegraphics[width=0.096\textwidth]{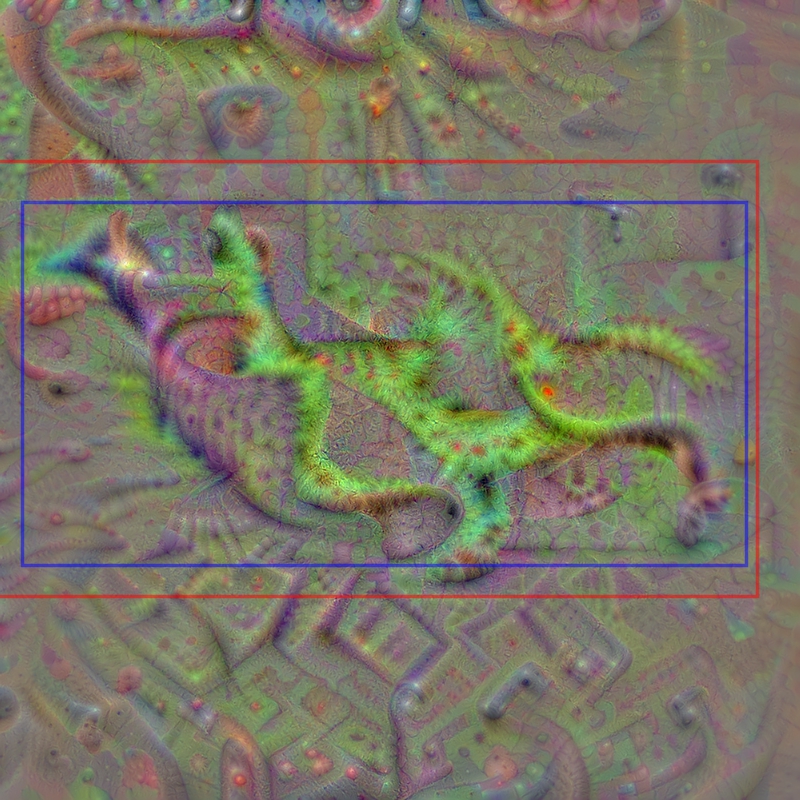}&
\includegraphics[width=0.096\textwidth]{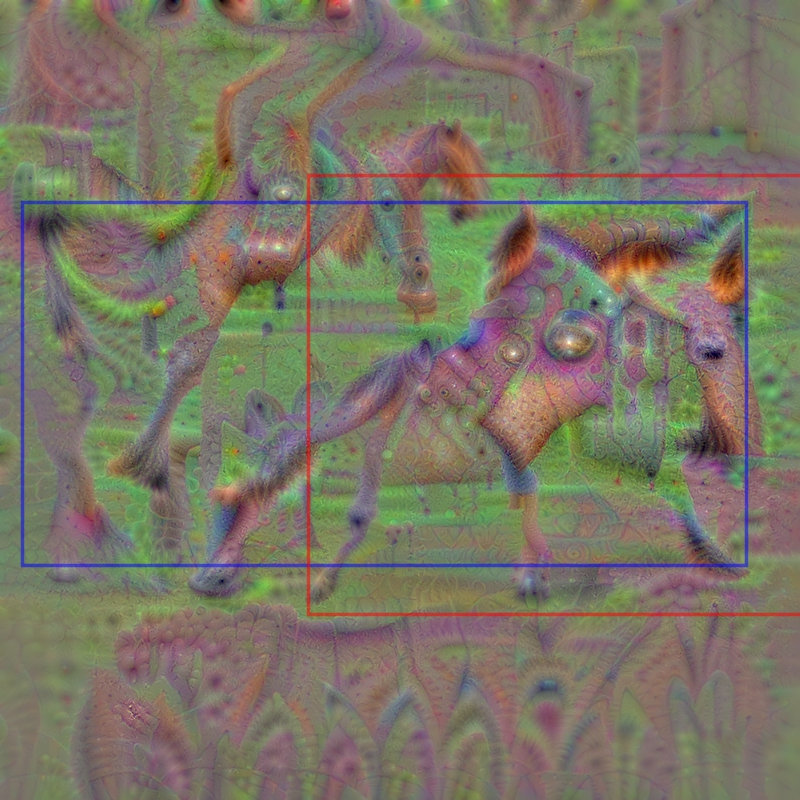}

\\
\small \text{Person}&
\small \text{Car}&
\small \text{Dog}&
\small \text{Cat}&
\small \text{Horse}\\
\end{tabular}}
\vspace{-3mm}
\caption{
  We optimize the predictions of Faster R-CNN~(col 1-3) and RetinaNet~(col 4-5) on a single anchor (like Figure~\ref{fig:class_specific}) at different anchor scales.
  Anchors are shown in blue and predicted boxes in red. 
  For both detectors, along with scale increasing, objects gradually lose global structures and show disconnected parts.
}
\vspace{-6mm}
\label{fig:scale}
\end{figure}

\subsubsection{Varying Object Scale.}
\label{sec:var_scale}
Unlike classification models, detectors are expected to recognize objects of different scales, often with Feature Pyramid Network (FPN).
In Detectron2 settings, objects are categorized into scale 1-5 with object areas of $\{32^2, 64^2, 128^2, 256^2, 512^2\}$.
To this end, it is of great interest to visualize individual objects of different scales.
We show such visualization in Figure~\ref{fig:scale}.  
% To better understand this setting, in Figure~\ref{fig:scale} we visualize individual objects on anchors in different FPN scales on Faster R-CNN.

We see that detectors rely on qualitatively different cues when recognizing the same class at different scales.
Small objects (scales 1 and 2) rely heavily on context and possess good global structures: dogs are near people, and cars are on roads or next to houses.
Medium-sized objects (scale 3) rely on context (e.g. car behind motorcycle) and also have clear structures: dogs have faces and legs; cars have wheels and windows.
Large objects (scale 4 and especially 5) have little coherent context or structure; detectors seem to recognize them disconnected parts or textures.
\begin{figure}[ht!]
    \centering
    \begin{subfigure}{0.20\textwidth}
        \centering
        \includegraphics[width=1\textwidth]{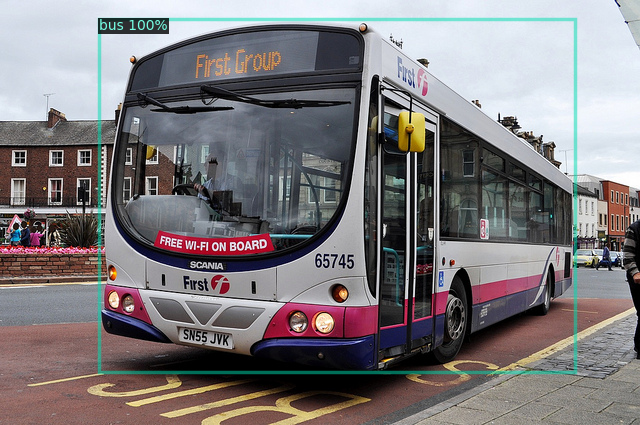}
        \caption{Original Image}
    \end{subfigure}%
    \begin{subfigure}{0.285\textwidth}
        \centering
        \includegraphics[width=1\textwidth]{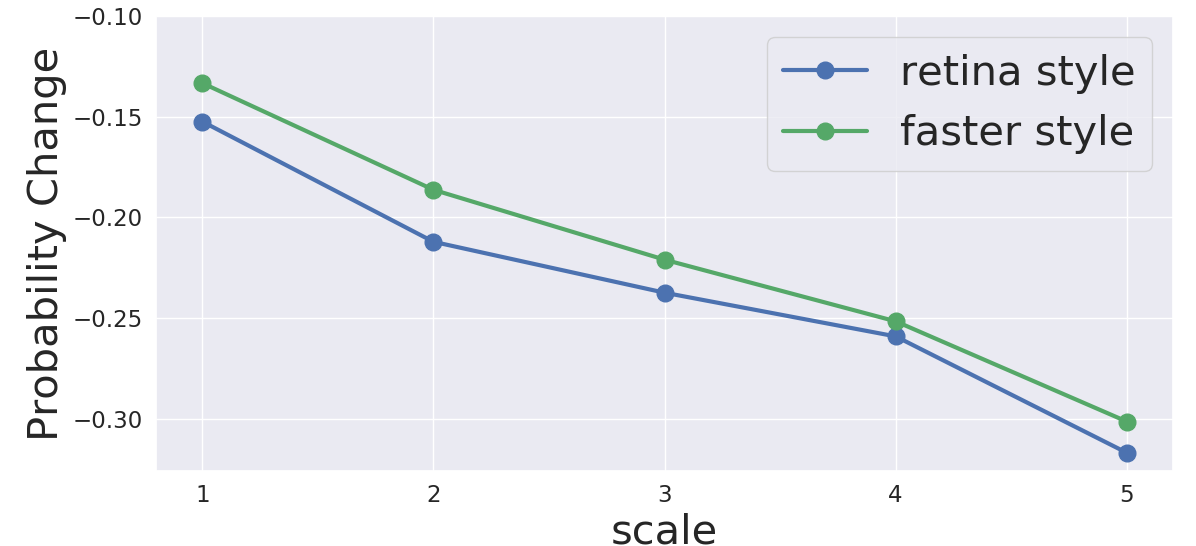}
        \caption{Probability change vs Scale}
    \end{subfigure}
    \\
    \begin{subfigure}{0.20\textwidth}
        \centering
        \includegraphics[width=1\textwidth]{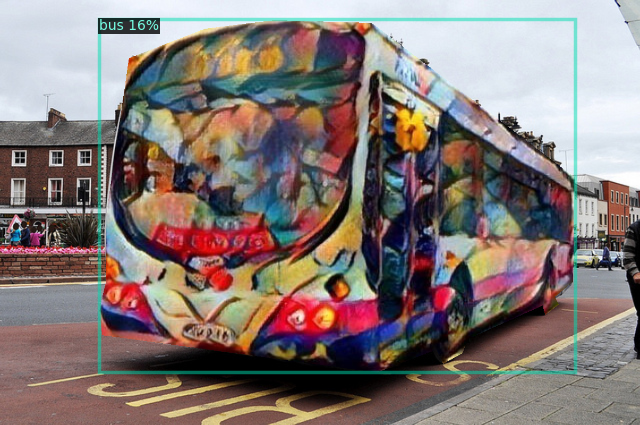}
        \caption{Masked stylized image}
    \end{subfigure}%
    \begin{subfigure}{0.285\textwidth}
        \centering
        \includegraphics[width=1\textwidth]{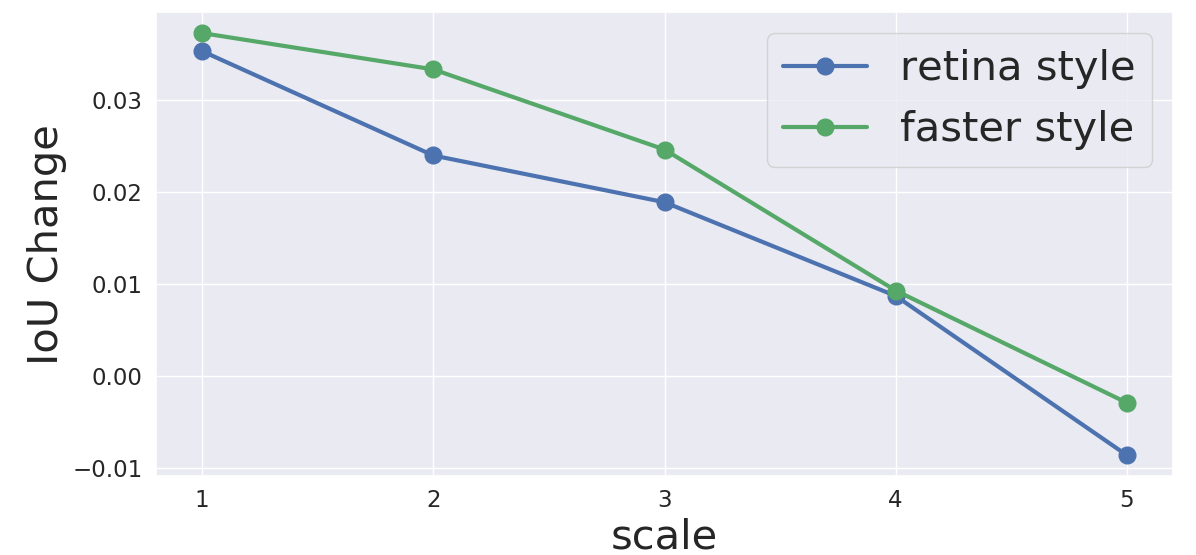}
        \caption{IoU change vs Scale}
    \end{subfigure}
    \vspace{-3mm}
    \caption{
      We eliminate object texture by performing masked style transfer.
      We measure how this changes detectors' predicted probability and IoU with ground-truth as a function of object scale.
%      We then calculate the difference between detectors' predicted probability and box  as a function of object scale.
      % We replace the original pixels of objects by style transfer
    % to eliminate texture information, and calculate prediction probability and IoU changes for different scale objects.
    }
    \vspace{-5mm}
    \label{fig:padding}
\end{figure}

The qualitative difference between different scales indicates the shifting of visual cues for object detectors: 
from relying on global shape and context at small scales, to large structure and texture at large scales.
% from relying on global shapes and contexts to local structures/textures with the growing of object size.
This may help explain texture-bias in ImageNet-trained classification models~\cite{geirhos2018imagenet}, since most objects in ImageNet are large.
% This shifting also relates to the texture-bias in ImageNet-trained classification model \cite{geirhos2018imagenet},
% since objects have big sizes in ImageNet.

% To validate this idea, for each object, we fill the appearance of the object by silhouette~(the mean color of the dataset) or stylized padding and compute performance changes after padding.
% Inspired by~\cite{geirhos2018imagenet}, we quantify the feature difference used to detect different-scale objects by masked style transfer.
Inspired by~\cite{geirhos2018imagenet}, we use masked style transfer to quantify the reliance of object detectors on texture.
As shown in Figure~\ref{fig:padding}, we use an object's ground-truth mask to perform style transfer keeping background pixels unchanged, obfuscating the object's texture.
We pass the original and stylized images to RetinaNet and Faster R-CNN and record the change in predicted probability and box location.
We repeat for all objects in the \verb|val2017| split and plot the results as a function of object scale.

%As shown in Figure~\ref{fig:padding}, for each object,
%we pad the mask-annotated object with style-transferred pixels while keeping the other pixels unchanged.
% We calculate prediction and IoU changes for each object after paddings and experiment on the COCO-val dataset with RetinaNet and Faster R-CNN.  Padding with stylized pixels eliminates objects' texture while keeping or even slightly enhancing global shapes since padding may cause shaper boundaries. 

As object scale increases, prediction probability for stylized objects drops sharply.
This indicates that detectors rely more on texture for large objects than for small objects.

% Large-scale objects have massive prediction probability drops for both detectors, indicating that object detectors rely more on textures to predict big objects than small objects.

Comparing IoU and probability changes after stylization, probability is decreased significantly with stylization for objects of all scales,
while IoU is less affected, and even slightly improved. 
This supports our claim in Section~\ref{sec:dis_losses} that classification and regression heads rely on different features; specifically texture is less important for the regression head. 
The marginal improvement in IoU may be due to sharper boundaries caused by stylization.

Comparing the detectors, Faster R-CNN has smaller drops in probability and larger increases in IoU,
suggesting that Faster R-CNN relies less on texture and more on shape than RetinaNet, though the effect is small.
This helps explain the layout inversions in Figure~\ref{fig:visual_results}, where RetinaNet struggles with boundaries of large objects.

% Compared with RetinaNet, Faster R-CNN has smaller drops in probability and bigger increases in IoU, 
% which implies that Faster R-CNN relies more on global shapes, compared with RetinaNet. 
% This trend is also shown in layout inversion in Figure~\ref{fig:visual_results}, where RetianNet struggles to show the boundaries of big objects clearly.   

\section{Conclusion}
In this paper, we extend visualization to modern object detector and propose an optimization-based approach for layout inversion.
Our visualization reveals intriguing properties of object detectors.
We hope these insights can help practitioners diagnose and improve object detectors.

{\small
\bibliographystyle{ieee_fullname}
\bibliography{egbib}
}

\end{document}

\input{supp_sections/layout_inversion.tex}
\input{supp_sections/inverse_transfer.tex}
\input{supp_sections/attributions.tex}
\input{supp_sections/individual.tex}
\input{supp_sections/shift_scale_variance.tex}